\documentclass{article}
\usepackage{preprint,times}

\usepackage{amsmath,amsfonts,bm}

\def\eqref#1{equation~\ref{#1}}

\def\1{\bm{1}}

\DeclareMathAlphabet{\mathsfit}{\encodingdefault}{\sfdefault}{m}{sl}
\SetMathAlphabet{\mathsfit}{bold}{\encodingdefault}{\sfdefault}{bx}{n}

\usepackage{hyperref}
\usepackage{url}
\usepackage{graphicx}
\usepackage{booktabs}
\usepackage{amsmath,amssymb}
\usepackage{multirow}
\usepackage{subcaption}
\usepackage{float}
\usepackage{xcolor}
\usepackage{colortbl} %
\definecolor{ourrow}{HTML}{FFE5E5} %
\usepackage{pgfplots}
\usepackage{pgfplotstable}
\usetikzlibrary{positioning,arrows.meta,patterns,calc,fit,backgrounds}
\pgfplotsset{compat=1.18}

\usepackage{longtable} %
\usepackage{array} %
\usepackage[section]{placeins} %

\newcommand{\dstar}{\ensuremath{\Delta^{*}}}
\newcommand{\present}{\textsc{present}}
\newcommand{\absent}{\textsc{absent}}

\title{Have I Seen Enough? Frozen Video-Language Models Encode Evidence Readiness}

\author{
  Dan Ben-Ami$^{1}$ \quad Kobi Cohen$^{2}$ \quad Chaim Baskin$^{1}$ \\[0.5em]
  \normalsize $^{1}$INSIGHT Lab, Ben-Gurion University of the Negev, Israel \\
  \normalsize $^{2}$Ben-Gurion University of the Negev, Israel
}

\begin{document}

\maketitle

\begin{abstract}
Streaming video-language models must decide not only what to answer, but whether the evidence needed
for the current question has arrived. Existing systems learn that decision as a separate trigger; we
ask whether an unmodified model already computes it. We show that frozen VideoLLMs carry a linearly
readable \emph{evidence-readiness} signal, labelled from timestamped evidence rather than from model
output. It decodes in all seven models of a shared byte-identical evaluation (AUROC $0.733$--$0.905$
under the strictest not-ready sampling, where a fitted clock is near chance), and a probe fitted
without any of a benchmark family's footage still reads that family. It is question-conditioned: on
byte-identical windows, changing only the question reverses the readout on $66.1\%$ of pairs, while
every question-blind control is at chance by construction. The model can answer incorrectly and still
encode readiness: AUROC remains $0.722$ among wrong answers. Readiness also beats uncertainty
estimators and their supervised combination on latency-matched answer selection, and tracks independent
human judgments more closely than confidence. Released streaming triggers are also linear readouts,
yet a trained trigger read on its own base model's activations is approximately orthogonal to readiness
and decodes it far less accurately than a probe. We turn the readout into \emph{Readiness Gating}, an answer-timing
policy that improves accuracy by up to $+9.75$ pp at matched video duration with negligible computational
overhead. How much it gains varies with the accuracy headroom the task makes available: across $26$
configurations the gain tracks that headroom, and an intervention that moves it over identical pixels
moves the gain with it. \href{https://danbenami.github.io/have-i-seen-enough/}{\includegraphics[height=1em]{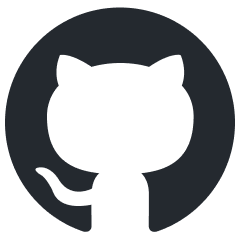}}

\end{abstract}

\section{Introduction}
\label{sec:intro}

\begin{figure}[t]
\centering
\includegraphics[width=0.97\textwidth]{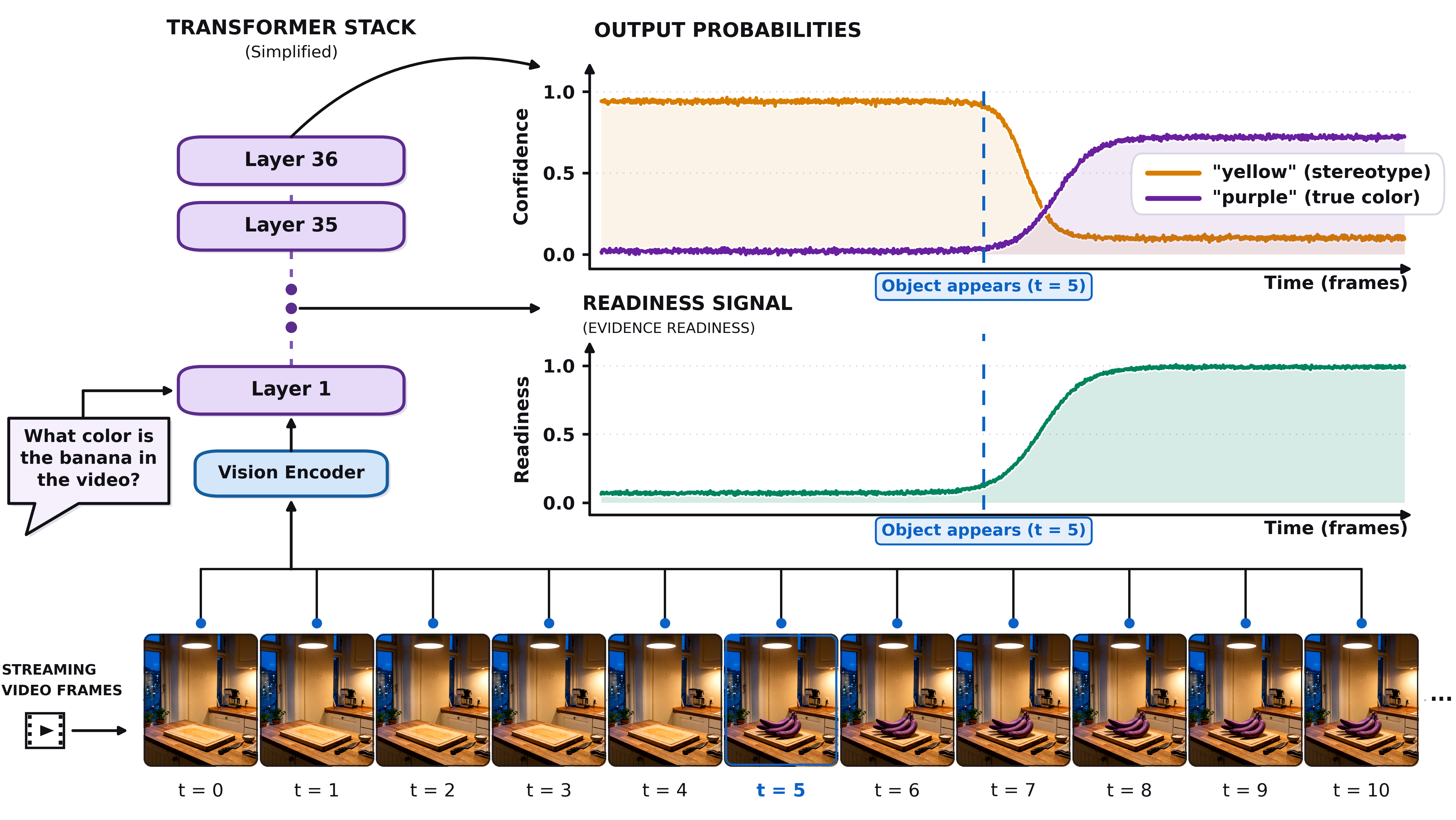}
\caption{\textbf{A model can be confident and unready at the same time.} Asked an object's colour
before it enters the stream, the model answers confidently from prior alone; a linear readout of
its frozen residual stream reports that the evidence has not arrived. \emph{(Schematic; measured in
\autoref{sec:signal}.)}}
\label{fig:teaser}
\end{figure}

Video-language models are moving from offline clips to continuous, interactive streams
\citep{lin2024streamingbench,chen2024videollm,wang2025streambridge,azad2026streamready,xie2026streamov}.
This changes the inference problem: in an offline benchmark the complete clip is available before the
question is answered, whereas in a stream the question may arrive before, during, or after the event
that resolves it, so a system must decide both \emph{what} to answer and \emph{whether it has seen
enough yet}. Existing streaming systems learn a readiness token, response trigger, auxiliary head, or
activation model; we ask whether the state they require is already computed inside a frozen VideoLLM.

Consider a model asked what object a person will pick up. Before the hand reaches it, scene and
language priors already give a plausible answer high probability; after contact the relevant pixels are
available, yet perception or reasoning can still fail. Confidence cannot distinguish these
cases and can point the wrong way (\autoref{fig:teaser}). We therefore separate three variables video QA often conflates: \emph{evidence availability}, a property of the delivered video relative to
the question; \emph{confidence}, a property of the output distribution; and \emph{correctness}, which
also depends on perception and reasoning. Our labels use question-specific evidence timestamps, fixed
before any model output is seen.

Nothing in how these models are trained asks them to represent this: next-token prediction supervises
what to say, not whether the frames delivered so far settle the question, so a model could reach high
accuracy on complete clips without ever forming it. Frozen VideoLLMs form it anyway. Readiness becomes
linearly readable in the residual stream after the first transformer block and strengthens with depth.
Changing only the question reverses the readout on byte-identical pixels, tying the signal to
vision--language interaction.

Readiness survives depth, shuffle, frame-count and identical-pixel controls, remains readable within
confidence strata and wrong answers, and appears across seven models on identical inputs and footage
excluded from probe training. Eleven uncertainty estimators and released streaming triggers test
competing readouts. \emph{Readiness Gating} uses the signal for answer timing, with gains depending on
task headroom.

\paragraph{Contributions.}
\begin{itemize}\itemsep2pt\parskip0pt
\item \textbf{We identify evidence readiness in frozen VideoLLMs:} a linear readout recovers whether
the current question's evidence is available. On identical video windows, changing only the question
reverses the readout on $66.1\%$ of pairs (\autoref{sec:signal}).

\item \textbf{We distinguish readiness from confidence and correctness:} it remains decodable among
wrong answers, outperforms eleven uncertainty estimators on the tested selection tasks, and tracks
independent human judgments more closely than output confidence (\autoref{sec:notconf}).

\item \textbf{We establish generality across models and footage:} readiness decodes in seven models,
and probes retain informative rankings on held-out benchmark families (\autoref{sec:general}).

\item \textbf{We introduce \textit{Readiness Gating}:} an answer-timing policy that improves accuracy
by up to $9.75$ pp at matched video duration in offline replay, without an additional forward pass
(\autoref{sec:utility}).
\end{itemize}

\section{Related work}
\label{sec:related}

\paragraph{Streaming video and answer timing.}
Streaming benchmarks expose the gap between complete-clip models and systems that operate as frames
arrive \citep{lin2024streamingbench,li2025ovobench}, and online VideoLLMs address it with interleaved
training, memory, proactive-response modules, or explicitly learned readiness mechanisms
\citep{chen2024videollm,wang2025streambridge,azad2026streamready,xie2026streamov}. These optimize when
a model should respond; we ask instead whether a standard frozen model already contains a reusable
internal variable from which that decision can be made. Where the learned trigger is released it is
itself a linear readout of a residual stream \citep{wang2024mmduet,qian2025dispider,chen2024videollm},
which lets us compare it with our probe directly (\autoref{sec:rivals}).

\paragraph{Temporal grounding and evidence.}
Moment retrieval localizes a language-described interval, while grounded and evidence-backed VideoQA
associate an answer with temporal or spatio-temporal support
\citep{hendricks2017localizing,xiao2023trust,wang2026evidence}. Adaptive frame selection chooses which
frames to present to a VideoLLM \citep{yu2023selfchained,yu2025framevoyager,benami2026himu}; we use
evidence annotations to assess whether the delivered frames already support the question. An annotated span
need not be the earliest or only sufficient evidence, a measured boundary on utility.

\paragraph{Uncertainty, abstention, and answerability.}
Calibration and self-evaluation estimate whether an output is correct
\citep{guo2017calibration,kadavath2022language}; sampling-based consistency and semantic entropy
measure stability across generations or meanings
\citep{wang2023selfconsistency,farquhar2024semantic}; and work on abstention separates answer
correctness from question answerability \citep{wagner2026two}. Our target is question-specific visual
evidence in the frames observed so far. Because predicting a correct answer does not necessarily reveal
missing evidence, we evaluate readiness separately at the representation and decision levels.

\paragraph{Probing and activation intervention.}
Probes can expose information available in a representation even when generation does not express it
\citep{belinkov2021probing,orgad2025llms}. To distinguish represented information from probe learning
\citep{hewitt2019designing}, we accompany decoding with layer-$0$ controls, grouped
nulls, identical-pixel question interventions, and analyses conditioned on confidence and correctness.
Activation steering tests whether a decoded direction affects behavior
\citep{li2023inference,zou2023representation}, but conclusions depend on the intervention and control
\citep{zhang2023towards,tan2024analysing}, so we use per-example magnitude matching and treat
sufficiency and necessity as separate hypotheses.

\section{Measuring evidence readiness}
\label{sec:method}

\subsection{Video inputs and readiness labels}

For question $q$, we label readiness from the benchmark's evidence timestamps and the frames shown to
the model, independently of its answer. We use HD-EPIC \citep{perrett2025hdepic}, NExT-GQA
\citep{xiao2023trust}, ST-Evidence \citep{wang2026evidence}, ActivityNet-Captions
\citep{krishna2017dense}, Oops! \citep{epstein2020oops}, HERBench \citep{benami2025herbench} and
OVO-Bench \citep{li2025ovobench}. Most evaluations use natural evidence; controlled stimuli add colour
cues that contradict prior expectations to egocentric video. We construct questions from source
annotations for HD-EPIC action pairs, HERBench shot identification, ActivityNet-Captions and Oops!
(Appendix~\ref{app:constructed}). Benchmarks supplying questions and evidence timestamps test whether
readiness is encoded in model representations without relying on our question construction; policy
evaluations require settings where evidence arrival can improve answers.

\paragraph{Fixed-duration windows.}
Our reference input, used wherever possible, is a $16$\,s video window advanced every $2$\,s, with $32$
sampled frames. Across fixed-duration formats, a question is labelled ready only when the delivered
frames cover the interval from the earliest evidence start to the latest evidence end. This interval
is the \emph{evidence hull}; covering it completely is \emph{full containment}. Windows covering none
of the evidence are not ready. Partial-coverage examples are excluded from binary probe fitting and
scoring, and retained as not ready in policy evaluation, because an answering policy must handle
them; against fractional coverage they are ordered rather than dropped
(Appendix~\ref{app:geom-contain}). Fixed
duration removes window length as a cue to readiness; window times are chosen independently of
evidence timestamps.

\paragraph{Growing prefixes.}
A growing prefix $[0,t]$ shows sampled frames from the video's beginning to the current time. Its
binary label marks \emph{onset}: at least one delivered frame lies within an annotated evidence
segment, rather than requiring the full interval to have passed. Switching to containment on the
same cached representations, with partial cases excluded, changes probe AUROC by at most $0.011$ on
the two natural prefix datasets (Appendix~\ref{app:geom-contain}). For growing prefixes, we also
measure how much of the question's evidence has been shown. The target $\mathrm{seen}/K$ is the
number of annotated evidence segments observed divided by the total number for that question,
giving questions with different numbers of segments a common scale. We evaluate prefixes at times
selected from the annotated evidence starts and ends. Because this schedule depends on evidence
annotations, these results support comparisons of which policy answers earlier on that schedule,
but not claims about absolute latency in deployment.

Agreement between windows and prefixes supports the signal's presence across different ways of
presenting video. Seven additional formats serve specific designs. Frame count is constant within
each format, and AUROCs should not be compared across formats (\autoref{app:geometries}). Each result
identifies its label.

\paragraph{Checking against people.}
Evidence timestamps mark where an answer is shown, not necessarily when it first becomes answerable.
Three raters at our institution, unaffiliated with this work, judged $\mathbf{900}$ blinded screens
showing the frozen model's exact input frames. Before adjudication, the two primary raters'
three-category judgments on all screens (yes, no, uncertain) yield Cohen's $\kappa = 0.717$
(\autoref{app:raters}).
Section~\ref{sec:signal} evaluates the probe against their judgments.

\paragraph{Sampling not-ready windows.}
We compare not-ready windows sampled before the evidence, just before it, after it, and on both
sides. Moving from before-only to both-sided sampling reduces the time-only baseline's AUROC from
$0.950$ to $0.527$, while readiness remains decodable across all seven models. We therefore use
both-sided sampling as the primary regime and the full sweep to assess sensitivity to timing
shortcuts (\autoref{sec:general}).

\subsection{Frozen representations, probe, and controls}

\paragraph{The probe tests linear availability; it does not teach the model readiness.}
We freeze the VideoLLM, Qwen3-VL-8B unless stated otherwise, and extract residual-stream states after
decoder layers addressed by fraction of model depth, so one procedure applies across architectures; the primary state is the final prompt token
immediately before answer generation, with option-token and pooled visual states as complementary
controls. Layer $0$, the input embeddings before any transformer block, differs in kind between the two
positions: at the language position that state is identical across checkpoints for a fixed question, so
chance there is an entailment of the architecture, whereas visual layer-$0$ states can already separate
ready from not-ready inputs on raw image statistics and so supply a measured baseline the language-position
readout must exceed. An L2-regularized logistic regression is standardized on training-fold statistics
only, with readout layer and regularization selected inside nested cross-validation; whole videos are
held out, and both members of a matched pair or all four members of a crossed quad stay in one fold. We
report held-out AUROC for binary readiness and partial Spearman correlation for graded progress, with
cluster-bootstrap intervals over videos or controlled cue groups. Where one probe is fitted across
benchmarks, it is fitted on the union of all-but-one benchmark family and applied frozen to the family
held out (Appendix~\ref{app:probe-cv}).

\paragraph{Elapsed time means a different thing on each of three pools.}
The basic null is a within-group label shuffle; further baselines include elapsed time or fractional
video progress, confidence, pixel statistics, motion and frame self-similarity, evidence-segment count,
answer letter and correctness. Where not-ready rows are drawn only from before the evidence, elapsed
time measures the sampling shortcut above, and both-sided sampling weakens it. On a growing prefix it is informative: $0.628$--$0.722$ on
the prefix evaluations of \autoref{tab:core}, and higher where evidence onset is itself strongly
time-correlated, as an entailment of the geometry that no sampling choice removes, since the prefix
lengthens while the evidence arrives once. On a \emph{policy} ladder, where checkpoints are placed without
reading any timestamp and each delivers a fixed span, it carries no clock at all and sits at
$0.500$--$0.506$. Every clock number below names its pool. The decision comparison includes eleven
estimators adapted from five method families covering self-consistency, semantic and predictive entropy, direct
self-assessment, option-order consistency and SigLIP retrieval \citep{zhai2023siglip}. We give each
estimator its most favourable tie-break and include model confidence and three no-forward-pass window
ensembles as references (Appendix~\ref{app:signal-baselines}).

\paragraph{The crossed-quad DiD cancels everything but question--evidence alignment.}
Recent-window matched pairs hold the video, question, option order and frame count fixed while changing
evidence containment. Crossed quads are stronger: two questions from one video with disjoint evidence
hulls are crossed with two windows, each containing exactly one hull, so every window is ready for one
question and not ready for the other. Their \emph{difference-in-differences} (DiD), the $2\times2$
interaction of readiness score in question and window, takes both inner differences on byte-identical
frames, so any fixed question or window effect cancels and what survives is the alignment between which
evidence is present and which question is asked. The DiD is $0$ for any readout that ignores the
question, and any statistic depending only on the pixels receives the same positive and negative
multisets and so is necessarily at AUROC $0.5$ (Appendix~\ref{app:probe-quad}).

\subsection{Readiness Gating}
\label{sec:gating-def}

Probing establishes that the state is present; \emph{Readiness Gating} is the policy that acts on it:
inference-time rules that decide when to answer from the probe's score rather than from output
confidence, in two modes. \textbf{Gated stopping} stops at the first checkpoint where readiness and
confidence both clear one training-selected threshold on their train-fold quantile ranks;
\textbf{gated selection} returns, at a fixed deadline, the answer cached at the highest-readiness
window so far, so all selection methods use identical video duration. Both add a $1.06\,\mu$s dot
product to a $251$ ms forward pass, reusing an existing residual vector, and neither decodes at a
checkpoint: the answer logits come from that same pass, so a checkpoint retains a letter and a
scalar rather than a key-value cache (\autoref{app:rivals}).
Throughout, the \emph{probe} is the fitted readout that AUROC measures, while \emph{Readiness
Gating} is the policy built on it that the timing and selection numbers measure. We report the
accuracy difference $\dstar$, in percentage points (pp), against the same rule on confidence alone
at matched video seconds, so $\dstar$ is what readiness adds to confidence, not what it beats.

\section{Evidence readiness: presence, separation, generality, utility}
\label{sec:results}
\label{sec:signal}

\subsection{A computed, question-conditioned readiness signal}

\paragraph{Readiness decodes even when prior knowledge cannot answer.}
We composite $119$ recoloured cue objects into clean HD-EPIC cooking windows, making their usual
colours unreliable guides to the question ``what colour is the object?'' Each composited window is asked about the colour of
the object in it and about the colour of an object that never appears, so the same frames carry one
ready and one not-ready case. Across $260$ instances, $520$ prefixes and $31$ videos, under nested
cross-validation with the cue object held out, readiness decodes at AUROC $\mathbf{0.985}$ $[0.965,0.999]$ against a grouped shuffle of $0.524$ at constant frame
count, and separates the queried cue from an equally salient irrelevant one
(Appendix~\ref{app:signal-decode}; Appendix~\ref{app:signal-question}).

\begin{table}[!t]
\centering
\scriptsize
\setlength{\tabcolsep}{2pt}
\begin{tabular}{lllc@{\hspace{4pt}}ccc@{\hspace{5pt}}>{\columncolor{ourrow}[3pt][2pt]}c}
\toprule
source & geometry & label & units & shuffle control & confidence & visual layer-$0$ & \textbf{readiness AUROC} \\
\midrule
HD-EPIC$^\dagger$ & composite & cue present & 520 prefixes & $0.524$ & $0.845$ & $0.500$ & $\mathbf{0.985}$ $[.965,.999]$ \\
HD-EPIC$^\dagger$ & matched pair & availability & 4,800 windows & $0.495$ & $0.636$ & $0.495$ & $\mathbf{0.827}$ $[.812,.840]$ \\
NExT-GQA & quad & containment & 1,208 quads & $0.498$ & $0.588$ & $0.500$ & $\mathbf{0.727}$ $[.710,.745]$ \\
NExT-GQA & prefix & onset & 5,580 windows & $0.502$ & $0.580$ & $0.544$ & $\mathbf{0.733}$ $[.715,.749]$ \\
ST-Evidence & prefix & onset & 6,546 windows & $0.484$ & $0.589$ & $0.552$ & $\mathbf{0.722}$ $[.706,.739]$ \\
ST-Evidence-Gen & prefix & onset & 4,346 prefixes & $0.578$ & $0.622$ & $0.651$ & $\mathbf{0.827}$ $[.813,.840]$ \\
Oops!$^\dagger$ & short window & containment & 1,204 windows & $0.496$ & $0.642$ & $0.654$ & $\mathbf{0.874}$ $[.855,.892]$ \\
ActivityNet$^\dagger$ & containment window & containment & 2,390 windows & $0.493$ & $0.579$ & $0.517$ & $\mathbf{0.764}$ $[.742,.787]$ \\
HERBench$^\dagger$ & burst & availability & 6,209 bursts & $0.498$ & $0.738$ & $0.534$ & $\mathbf{0.918}$ $[.906,.930]$ \\
\bottomrule
\end{tabular}
\caption{\textbf{Core representation results.} Labels come from evidence timestamps, not model
correctness. Rows marked $^\dagger$ pose questions we construct rather than questions the source ships
(Appendix~\ref{app:constructed}). The column is nine separate tests, not one
scale: rows differ in geometry and strictness, so each AUROC is read against the three columns beside
it rather than against the rows above it (Appendix~\ref{app:geometries}). Read that way the verdict is
uniform, with readiness above all three columns on every row by $0.133$ to $0.220$. Only the shuffle is a control, and it is
measured on every row rather than $0.500$ by construction; the other two are competing readouts that may
legitimately carry signal. Visual layer-$0$ reads raw patch embeddings before any decoder layer, against
a language-position layer-$0$ readout forced to $0.500$ everywhere (\autoref{sec:method}); it is forced
by the design on the crossed quads, and on the controlled composites is quoted under a pair-preserving
grouping (\autoref{app:probe-quad}).}
\label{tab:core}
\end{table}

\paragraph{On identical pixels, changing the question flips the readout on most pairs.}
The decisive intervention changes the question on fixed video bytes. In the controlled set, swapping
``what colour is A?'' for ``what colour is B?'' reverses the score on $98.1\%$ of $520$ pairs. On
natural NExT-GQA footage each crossed quad pairs two disjoint-evidence questions
with two windows, one containing each hull. Across $1{,}208$ quads, changing the question on identical
pixels shifts model accuracy by $\mathbf{+14.0}$ pp in the predicted direction ($77.9\%$ versus $63.9\%$)
and reverses the readout on $\mathbf{66.1\%}$ of swaps. Readiness reaches AUROC $\mathbf{0.727}$
$[0.710,0.745]$. Every window-only statistic, including elapsed time, motion, brightness, segment count
and pooled vision, has AUROC $0.500$ by construction (\autoref{sec:method}). The
difference-in-differences, which cancels any fixed difference between the two question embeddings and
so is stronger than a single swap (Appendix~\ref{app:signal-question}), is $0.429$ $[0.378,0.482]$ and
replicates at $0.487$ $[0.393,0.594]$ on ST-Evidence-MCQ. Each is on its own
probe's fitted scale, so the sign and separation from zero carry the claim rather than the magnitude.

\paragraph{On natural footage, the signal rises with how much evidence has arrived.}
The readiness advantage extends to natural footage, exceeding both time and confidence on all three
prefix evaluations: AUROC $0.733$ against a clock of $0.633$ on NExT-GQA, $0.722$ against
$0.628$ on ST-Evidence-MCQ, and $0.827$ against $0.722$ on ST-Evidence-Gen. The growing-prefix geometry
itself makes elapsed time informative (\autoref{sec:method}). On
ST-Evidence-Gen ($K=1$--$11$ evidence segments) readiness rises with $\mathrm{seen}/K$ even
after controlling fractional video progress (partial $\rho=\mathbf{0.378}$ $[0.350,0.402]$); raw
segment count gives a false null, since ``one of one'' and ``one of five'' differ
(Appendix~\ref{app:signal-graded}). A donor question on byte-identical frames collapses within-arm AUROC from
$0.832$ to $0.546$. Readiness also exceeds the clock on the settings used below: $0.874$ versus $0.527$
for Oops!\ short windows and $0.764$ versus $0.505$ for constructed ActivityNet-Captions windows
(\autoref{tab:core}).

\paragraph{Human judgments support the readiness interpretation.}
On the same windows, raters recover the timestamp labels at AUROC $0.681$--$0.778$ and the probe
those same labels at $0.671$--$0.791$ (Appendix~\ref{app:raters-parity}). Their disagreement with the annotations highlights
the distinction between timestamp coverage and perceived sufficiency. The raters' judgments also
predict answerability: model accuracy is $16.5$, $36.9$ and $28.0$ pp higher on windows they call ready
(Appendix~\ref{app:raters-target}).
Together, the depth (\autoref{fig:depth}), crossed-question, graded and rater analyses support a
question-conditioned evidence-readiness signal beyond question-blind visual cues and output confidence.

\subsection{Separation from confidence, correctness, and overt behaviour}
\label{sec:notconf}

\paragraph{Confidence does not explain the readout.}
Readiness is not a relabelling of answer probability, and three prediction tests say so. Within six
fixed confidence bins on the HD-EPIC matched \present/\absent\ windows, readiness AUROC stays at
$0.688$--$0.757$; residualizing the readiness score on confidence moves AUROC by at most $0.0002$ on
every pool in Appendix~\ref{app:signal-conf}; and adding the activation score to confidence improves
AUROC by $+0.088$ to $+0.397$; readiness outranks confidence in every one of the seven models of
\autoref{fig:depth}b. The
separation survives replacing the target. Scored against the $900$
screens' human readiness verdict instead of a timestamp (\autoref{sec:method}), the probe reads
$0.871$--$0.944$ on the three evaluations where the model's own confidence reads $0.643$--$0.691$. An
external multimodal LLM (\texttt{gemini-3-flash-preview}), shown the identical frames and asked
directly whether they suffice, ranks the human verdict at only $0.576$--$0.629$. AUROC is
threshold-free, so that is a failure to order the screens as people do rather than a badly set
yes-rate, and it is not noise: the judge answers consistently across repeats ($\kappa=0.698$)
(Appendix~\ref{app:raters-judge}).

\paragraph{The uncertainty estimators we tested predict correctness, not evidence availability.}
The sharper test is downstream answer selection (\autoref{sec:utility}) on Oops!\ evidence windows and an
evidence-sensitive ActivityNet-Captions condition, the two natural-video settings where readiness selection converts most. All
eleven estimators of \autoref{sec:method}, each given its own most favourable tie-break, lose on both,
every interval excluding zero. They are
not simply weak: confidence, self-consistency and option-order agreement predict the model's own
correctness above chance here ($0.572$--$0.687$) while ranking evidence availability worse and picking
the wrong window more often, and semantic entropy sits near chance because descriptive video answers
fragment into almost one cluster per sample, a limit of this application rather than of the method
(Appendix~\ref{app:signal-baselines}).

\begin{table}[t]
\centering
\scriptsize
\setlength{\tabcolsep}{4pt}
\begin{tabular}{lcc}
\toprule
measurement & Oops! & ActivityNet-Captions \\
\midrule
readiness AUROC on evaluated windows & $0.831$ & $0.753$ \\
elapsed-time AUROC on the same windows & $0.519$ & $0.436$ \\
\midrule
selection accuracy, answering at end of video & $48.7\%$ & $35.8\%$ \\
selection accuracy, confidence & $60.6\%$ & $54.0\%$ \\
\rowcolor{ourrow} selection accuracy, Readiness Gating (ours) & $\mathbf{67.6\%}$ & $\mathbf{64.7\%}$ \\
\midrule
gating advantage over supervised combination & $+4.5$ pp $[+0.3,+8.7]$ & $+8.2$ pp $[+4.6,+11.7]$ \\
combination with gating, relative to gating alone & $+0.3$ pp $[-1.6,+2.2]$ & $-0.2$ pp $[-1.8,+1.5]$ \\
\bottomrule
\end{tabular}
\caption{\textbf{Uncertainty baselines do not recover readiness on the two tested high-conversion
settings.} All arms answer at an identical deadline; the middle block gives the accuracies the
advantages come from (paired margins $+7.1$ and $+10.7$ pp over confidence). The two AUROCs are read on
the windows the policy chooses among, not the \autoref{tab:core} pools of the same names. Every
estimator loses on both, and adding a supervised combination back to gating gives no detectable gain.}
\label{tab:uncertainty}
\end{table}

\paragraph{Readiness reads evidence the model fails to use.}
Availability and correctness change the readout differently: readiness stays decodable when
the model is wrong. Across $4{,}800$ HD-EPIC examples, AUROC is $\mathbf{0.722}$ among wrong answers,
including $681$ with evidence present, and $0.753$ among correct answers. Model accuracy is $53.5\%$,
and the same-example clock reaches $0.506$. HERBench closes the other regime: on a
growing prefix whose evidence onset is strongly time-correlated (clock $0.956$ against readiness
$0.980$), so that it contributes the within-wrong comparison only, availability still decodes at
$0.970$ among $2{,}169$ present-but-wrong examples at $22.7\%$ accuracy against a $20.0\%$ chance
level. The readout tracks evidence availability whether or not the model can use it.

\paragraph{The probe exposes readiness more reliably than overt responses.}
The natural-data prompts carry an explicit ``there is not enough evidence yet'' option, so a model can
report what the probe reads. It largely does not: on NExT-GQA, Qwen3-VL-8B selects it on $9.7\%$ of
not-ready against $3.8\%$ of ready rows ($5{,}695$ rows), InternVL3.5-8B on $7.5\%$ against $2.1\%$,
LLaVA-OneVision-7B on $0.3\%$ against $0.0\%$. The latter two run on the same $5{,}580$ byte-identical
rows with indistinguishable readouts, both at AUROC $0.720$ ($[0.704,0.738]$ and $[0.704,0.737]$), yet
differ $\mathbf{26\times}$ in how often they act on it. The option is usable in principle (composited cues drive
deferral to $59.2\%$ and $52.5\%$ on not-ready items against $0.2\%$ and $0.8\%$ when ready) and deferral rates
depend on the option set and instruction, so this describes these models under this prompt; within that
scope it is a third dissociation: the probe exposes readiness that these prompts leave largely unspoken.

\paragraph{Injected at one site, the direction turns deferral into an answer.}
\label{sec:causal}
Forced in, the state does move that behaviour. We estimate a rank-$32$ readiness subspace from matched
\present/\absent\ differences and add it at a preselected mid-depth residual site in Qwen3-VL-8B,
before the answer letter is decodable, against a wrong-target direction of equal residual magnitude on
the same example: $\mathbf{101}$ deferral-to-answer flips against three for that control, a commitment
log-odds increase of $+0.392$ $[+0.331,+0.454]$ over it, monotone across four doses ($\rho=1.0$),
exceeding all $35$ alternative directions ($p=0.0278$) and leaving answer content inside a pre-specified
equivalence margin. With the site and analysis unchanged, InternVL3.5-8B produces $91$ flips against
the control's seven and a commitment log-odds increase of $+0.350$ $[+0.306,+0.395]$. The direction and
magnitude replicate; the dose curve is not fully monotone. This
establishes sufficiency at the tested site, not necessity: subtraction changes the model no more than a
magnitude-matched wrong-target edit, and the strictest component knockout amplifies the addition effect
rather than removing it (\autoref{app:causal}).

\subsection{Generality across models, footage, and tasks}
\label{sec:general}

\begin{figure}[t]
\centering
\includegraphics[width=\textwidth]{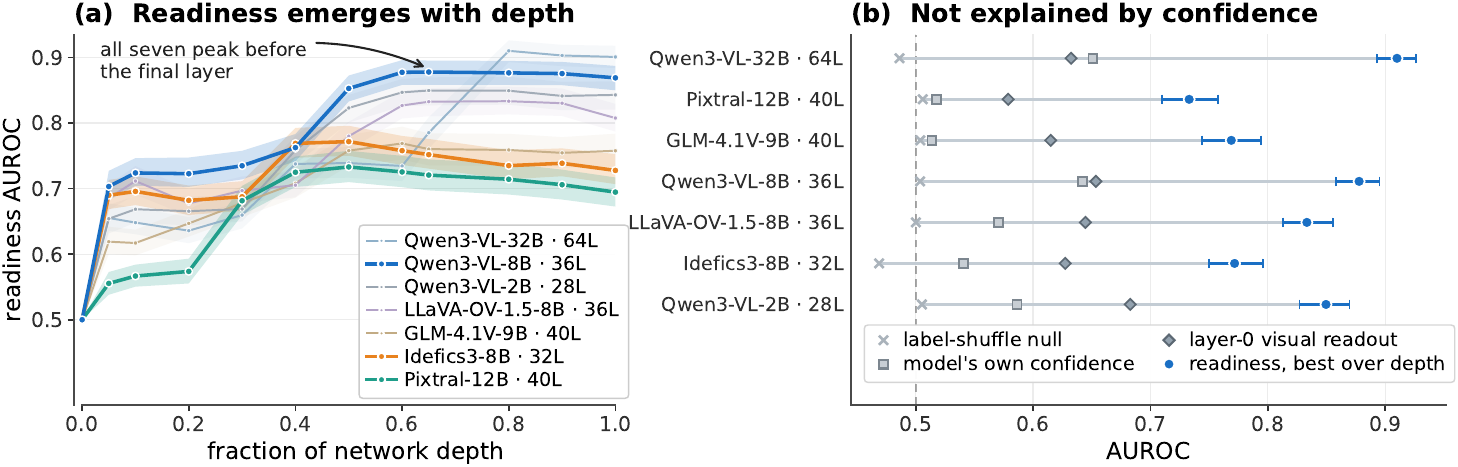}
\caption{\textbf{Readiness accrues with depth, and it is not the model's own confidence.} Seven models,
four language-backbone families, on one frozen Oops!\ manifest of $1{,}204$ windows over $301$ videos
with not-ready rows drawn from both sides of the evidence. \textbf{(a)} Held-out AUROC at the final
language position against fraction of decoder depth, cluster-bootstrap bands over videos; layer $0$ is the raw input embedding, exactly $0.500$ by
construction, and every model peaks between $0.5$ and $0.8$ of depth rather than at the output.
\textbf{(b)} Readiness at each model's best depth, rows ordered by parameter count, against the model's
own answer probability ($0.514$--$0.651$), the measured layer-$0$ visual readout ($0.579$--$0.683$) and
the label-shuffle null, all on the same rows. Parameter count does not order readiness: the $2$B model
reads above four larger ones.}
\label{fig:depth}
\end{figure}

\paragraph{Every model we tested carries the signal.}
The seven models read one frozen manifest of Oops!\ short windows, the substrate of the Oops!\ row of
\autoref{tab:core}, with byte-identical frames and option order, and readiness decodes in all of them
under every negative-sampling scheme. In the primary regime, with not-ready windows drawn from both
sides of the evidence, AUROC runs $\mathbf{0.733}$--$\mathbf{0.905}$ and none of the $28$
model-by-sampling measurements falls lower; grouped shuffles run $0.467$--$0.506$, and
every language-position readout clears its measured visual layer-$0$ baseline of $0.579$--$0.683$
(Appendix~\ref{app:gen-modelaxis}). The three Qwen3-VL sizes rise from $0.849$ to $0.905$ alongside task
accuracy, so the claim is architectural breadth rather than a scaling law; the seven span four
language-backbone families, a fifth entering on the grounded pools. A depth sweep on the same rows (\autoref{fig:depth}) reads
$0.500$ at the raw input embedding and peaks between $0.5$ and $0.8$ of depth in all seven, so
question-specific availability appears only after vision and language interact rather than being
readable off the input, and the models' own answer probabilities rank those rows at only
$0.514$--$0.651$.

\paragraph{Ranking transfers to unseen footage; the threshold does not.}
OVO-Bench is a stronger shift. The probe is fitted on one family only, $16$\,s containment windows over
ST-Evidence-Gen footage ($3{,}059$ rows over $693$ videos, ready when the window holds released
evidence), and is then read on OVO's $42$ unseen videos against OVO's own labels and its own task
definition. Carrying the scaler, direction and layer over \emph{without refitting} yields $0.748$
$[0.689,0.799]$ on the $3{,}913$ scored windows, about $90\%$ of the $0.832$ a target-fitted probe
reaches, while the obvious proxies point the wrong way rather than sitting at chance (shuffle $0.509$,
elapsed time $0.451$, the benchmark's own checkpoint index $0.479$, confidence $0.350$). The operating
point does not come with it: the frozen probe calls $99.3\%$ of target rows ready and reaches $13.8\%$
accuracy at the carried-over threshold against $86.8\%$ at a target-chosen one, though refitting the
threshold alone on eight target videos recovers $86.5\%$. At these base rates even an oracle threshold
beats a constant predictor by at most $0.028$, so what crosses the shift is the ranking and not the
operating point (\autoref{app:transfer}).

\paragraph{One direction reads five families it was not fitted on.}
The transfer above moves one direction across a single shift. We test transfer to an entirely unseen
family by training on the union of the remaining families and applying the probe without refitting
(\autoref{sec:method}), holding out whole footage groups rather than single benchmarks because
NExT-GQA and ST-Evidence-MCQ share $221$ NExT-QA videos.
The off-diagonal probe then reads each of the five held-out families at
$\mathbf{0.653}$--$\mathbf{0.761}$ against that family's own fitted ceiling of $0.745$--$0.874$,
$0.922$ of it on average with ActivityNet weakest, controls holding in every cell (shuffles
$0.498$--$0.521$, visual layer-$0$ $0.497$--$0.551$, clock $0.455$--$0.545$) and dropping the one
off-geometry family moving every remaining cell by less than $0.02$ (Appendix~\ref{app:gen-lobo}). Evidence
readiness is thus observed in seven shared-input models over four language-backbone families and eight
benchmark families; ranking is portable, while the operating point needs a little target calibration.

\subsection{Readiness Gating and the attainable-band relationship}
\label{sec:utility}

\paragraph{Gating buys accuracy at matched latency.}
Readiness Gating (\autoref{sec:gating-def}) is cheap enough that what limits it is the quality of the
score, not the cost of reading it, but the gain depends strongly on the task. Gated stopping improves
accuracy over confidence-gated stopping observing the same amount of video at matched latency:
$\mathbf{+9.75}$ pp $[+8.19,+11.54]$ on HERBench shot identification, $\mathbf{+6.03}$ pp
$[+3.62,+8.61]$ on Oops!, and $\mathbf{+4.66}$ pp $[+2.84,+6.54]$ on the evidence-sensitive
ActivityNet-Captions condition.
Every $\dstar$ we report is matched on video seconds rather than on
checkpoint index (\autoref{app:utility}). Gated selection is a threshold-free alternative: at a fixed
deadline return the cached answer from the highest-readiness window, so every method observes the same
duration and latency is matched by construction. Its statistic is not $\dstar$ but the selection edge
over the same policy driven by confidence, positive in 24 of 26 configurations with intervals excluding
zero in 15: a median $+3.85$ pp on fixed-span windows, $+3.26$ pp on growing prefixes, and $+40.5$ pp on
sparse four-option bursts, aided there by their one-in-eight readiness rate.

\paragraph{Policy gains track task headroom.}
The \emph{attainable band} is the accuracy gap between the evidence oracle and answering at the end
(Appendix~\ref{app:utility}), and it is worth measuring before optimizing a response trigger. A
controlled distractor manipulation tests how the reference gap and policy gain move together: holding
videos, windows, frames, evidence labels and answer positions fixed on the ActivityNet arm and
replacing only the distractor options moves end-of-video accuracy from $35.8\%$ to $79.5\%$ and the
band from $34.5$ to $17.5$ pp, and the gain falls with it from $+4.66$ to $+1.10$ pp, a
$\mathbf{4.2\times}$ contrast at an implied $+0.209$ pp of gain per pp of headroom. Across all $26$
matched-latency configurations over ten substrates, each entering once on its full item pool, $\dstar$
tracks the band observationally at roughly $\mathbf{15\%}$ conversion and rank correlation
$\rho = \mathbf{0.79}$, which holds at $0.70$ over the $23$ that remain when the three sparse-burst
configurations are dropped (Appendix~\ref{app:utility}). What the band identifies is where evidence-timed answering can help, not
how much: representation quality, calibration and policy error determine what is realized, and the
Oops!\ gate converts $0.48$ of its band against the pooled $0.15$. This is why utility is conditional
while the representation claim is not.
\subsection{Released streaming triggers versus the readiness probe}
\label{sec:rivals}

\paragraph{Zero-shot, the trained triggers do not transfer.}
Streaming systems use learned heads to decide when to respond. In the released systems examined here,
these heads are linear residual-stream readouts, like our probe: MMDuet uses $2\times3584$ maps on
LLaVA-OneVision (Qwen2-7B), Dispider a $1\times1536$ head, and VideoLLM-online an unembedding row
\citep{wang2024mmduet,qian2025dispider,chen2024videollm}. We drive the two whose released weights we could run end-to-end through their own inference
code, scoring them against our labels by ranking rather than at a threshold (\autoref{tab:rivals}).
The readiness probe substantially outperforms MMDuet's \texttt{relevance} head on identical rows:
AUROC $0.753$ versus $0.484$ $[0.470,0.498]$ on ActivityNet-Captions, and $0.926$ versus $0.513$
$[0.496,0.529]$ on HERBench shot-ID. On ActivityNet-Captions, the trigger's interval lies entirely below
chance. It is not being driven incorrectly: through the
same code on MMDuet's own published benchmark, Charades-STA temporal grounding \citep{gao2017tall},
it reaches $\mathrm{R@IoU}{=}0.5$ of $53.2$ against the $42.4$ originally reported
(\autoref{app:rivals}).

\begin{table}[t]
\centering
\scriptsize
\setlength{\tabcolsep}{3pt}
\begin{tabular}{lcccc}
\toprule
& \multicolumn{2}{c}{HERBench shot-ID} & \multicolumn{2}{c}{ActivityNet, same-action} \\
\cmidrule(lr){2-3}\cmidrule(lr){4-5}
monitor & readiness AUROC & selection acc. & readiness AUROC & selection acc. \\
\midrule
answer at end (no selection)   & ---      & $0.251$ & ---      & $0.358$ \\
evidence oracle                & ---      & $0.897$ & ---      & $0.703$ \\
\midrule
\rowcolor{ourrow} Readiness Gating (ours) & $\mathbf{0.926}$ & $\mathbf{0.763}$ & $\mathbf{0.753}$ & $\mathbf{0.647}$ \\
confidence                     & $0.749$ & $0.495$ & $0.594$ & $0.540$ \\
elapsed time                   & $0.507$ & ---      & $0.436$ & ---      \\
\midrule
MMDuet \texttt{relevance} (zero-shot)   & $0.513$ & $0.276$ & $0.484$ & $0.413$ \\
MMDuet \texttt{informative} (zero-shot) & $0.532$ & $0.325$ & $0.496$ & $0.433$ \\
Dispider \texttt{silent} (zero-shot)    & $0.477$ & $0.289$ & ---      & ---      \\
\bottomrule
\end{tabular}
\caption{\textbf{Released streaming triggers, scored against evidence-readiness labels on two
substrates.} These are the policy ladders, $6{,}368$ and $16{,}109$ windows, not the
\autoref{tab:core} probing pools of the same names (\autoref{app:rivals}). At a common deadline every arm
emits the answer cached from its own highest-scoring window, so selection is latency-matched by
construction. Our probe is fitted on each dataset's training folds and evaluated on held-out folds;
the released triggers are applied without fitting. Removing that asymmetry does not
change the ordering: on MMDuet's own activations a probe reads $0.843$ where its trained head reads
$0.607$ (\autoref{sec:rivals}). The comparison tests transfer to readiness labels, not which complete
streaming system performs best. Dispider is a secondary baseline and decides once per $16$\,s, too
coarse for the ActivityNet ladder, so those cells are empty; every rival selection arm is
significantly worse than plain confidence ($-0.107$ to $-0.219$).}
\label{tab:rivals}
\end{table}

\paragraph{On its own base model's activations, the trained direction points elsewhere.}
MMDuet's base model is by value the same language model in the same basis as a checkpoint our
extraction path can load, so the substrate difference can be removed. A forward pre-hook captures
exactly what MMDuet's \texttt{relevance} head receives, using its runtime, adapter, replaced visual
projector and query-first prompt order. We evaluate $1{,}204$ balanced Oops!\ windows over $301$
videos, drawn from both sides of the evidence so that a fitted clock on them reads $0.528$. The
trained direction reads $0.607$ $[0.579,0.634]$ there, clearing by $+0.059$ the $0.548$ a layer-$0$
readout at the same frame position reaches, where a probe fitted on the identical rows and
activations reads $\mathbf{0.843}$ $[0.822,0.864]$ and clears the same floor by $+0.295$. The two
directions are statistically indistinguishable from orthogonal, at absolute cosine $0.016$ against a
random-direction null whose $99$th percentile is $0.043$. MMDuet's question-blind \texttt{informative}
head detects our readiness labels better than its \texttt{relevance} head, achieving AUROC $0.666$ at
that same cell. The two head directions are nearly orthogonal
($\cos=0.029$). Thus, the head trained for relevance is not the better readiness readout.
The transplant rests on this one substrate, the only one whose base our path can load
(\autoref{app:rivals}).

\section{Discussion}
\label{sec:conclusion}

Frozen VideoLLMs contain a linearly readable state for whether the current question's evidence
has arrived. Identical-pixel crossed questions and depth controls, uncertainty
baselines and wrong-answer analyses, the seven-model evaluation with held-out footage, and external
transfer make this a representation result rather than a confidence or timing artifact, and the
probe exposes readiness these prompts leave largely unspoken. The released streaming triggers
we tested transfer poorly to our readiness labels, and on its own base model's activations the one
trigger we could check is approximately orthogonal to readiness and decodes it far less well than a
probe on those same activations (Appendix~\ref{app:rivals}). Readiness Gating turns it into an answer-timing policy whose
gain varies with task headroom (Appendix~\ref{app:utility}). Evidence readiness is a promising
primitive for streaming systems: already present in the model, separable from confidence, and
portable in ranking.

\paragraph{Limitations.}
Each bound below is expanded with its numbers in \autoref{app:limitations}. A live-decoding audit
complements our offline policy replay; long-stream latency and memory growth remain outside scope.
Outside the explicit transfer evaluations probes are fitted within each benchmark on video-disjoint
folds; ranking transfers externally, so gated selection carries over, but gated stopping needs an
operating point that can fail under label shift.
Evidence annotations identify where the answer is shown, although it may be inferable earlier.
Independent human ratings measure this gap, but the labels remain timestamp-based. Two natural-data
evaluations also share footage.
The trigger comparison establishes non-transfer of a foreign
trigger, not superiority as a system. The causal evidence covers one primary intervention site and
two video-language models, and establishes sufficiency there rather than necessity.

\section*{AI use statement}

Generative AI tools were used for writing assistance, literature discovery, and the experimental
procedures described below.

The first is aiding and polishing writing. Tools were used to draft and revise passages of this
paper from author-supplied content and notes, to improve the readability and concision of text
written by the authors, to tighten the argument, to help organise the manuscript into
sections, and to format references, tables, and figure captions in \LaTeX. Every such passage was
read, checked, and edited by the authors, who wrote or verified each claim it makes.

The second is retrieval and discovery. For literature discovery, AI tools helped find candidate papers
and relevant passages. Searches covered streaming video and answer timing, temporal grounding,
uncertainty and abstention, and probing and activation intervention. The authors read the papers used.
No citation entered this paper on a tool's assertion alone: each reference was
retrieved and read at its primary source, and its attributed claim checked there, by the authors.

Generative AI was also used to construct and screen benchmark items and as an evaluation baseline,
as documented in the relevant appendices. The authors take responsibility for all text, data,
analyses, and reported results.

\section*{Ethics statement}

This work analyses frozen, publicly released video-language models on publicly released video
question-answering benchmarks (HD-EPIC, NExT-GQA, ST-Evidence-MCQ and ST-Evidence-Gen,
ActivityNet-Captions, Oops!, HERBench, and OVO-Bench); no new video was recorded. One part of the evaluation involved people: three raters at our institution, unaffiliated
with this work, judged $900$ screens of that same publicly released footage to check our readiness
labels (Appendix~\ref{app:raters}). They volunteered without compensation, were told what the task
was for, and could stop at any point; we collected no personal or demographic data from them, release
no participant-level data, and they are not identifiable in anything we release. No crowdsourcing
platform was used. Because the study collected no personal data and involved only viewing already
public video, we did not seek institutional review. The egocentric footage in HD-EPIC
depicts identifiable people and private home interiors, and we use it only under its own release
terms, report no frame-level content beyond what those releases already make public, and release no
derived media. The readiness readout we study is a diagnostic on frozen activations rather than a
deployed capability, but we note the dual-use direction it points in: a reliable internal signal for
``the evidence needed to answer has now arrived'' is also a signal for when a continuously observing
system has acquired what it was watching for, and applying it to camera streams of people raises
surveillance and privacy concerns that are separate from the accuracy questions studied here. We see
no conflicts of interest or sponsorship to disclose.

\section*{Reproducibility statement}

The video-language models used here are frozen: no weights are trained, fine-tuned, or adapted at any
point, and the only fitted object is a linear probe on frozen activations. Readiness labels are derived
from each benchmark's own evidence timestamps before any model output is inspected, and
Appendix~\ref{app:raters} reports the independent-rater check on those labels together with its
instrument and its limits. Every reported
generalization number is held out at the level of whole videos, with the readout layer and the
regularization strength selected on an inner held-out-group loop, and every confidence interval is a
cluster bootstrap over whole groups. Appendix~\ref{app:probes} gives the probe protocol and the full
control battery, Appendix~\ref{app:geometries} the streaming geometries and label definitions,
Appendix~\ref{app:datasets} the benchmarks, pools, and models, and Appendix~\ref{app:repro} the
determinism and compute details, including the two respects in which our numbers are \emph{not}
bit-reproducible. Code, configurations, the frozen item sets of Appendix~\ref{app:constructed}, and
the result payload behind every number we report are provided as supplementary material, together
with a script that re-reads each headline number from those payloads. Video frames, activation
caches, and rater-level data are not redistributed, for the reasons given in
Appendix~\ref{app:licences}.

\bibliography{references}
\bibliographystyle{plainnat}

\clearpage
\appendix
\numberwithin{figure}{section}
\numberwithin{table}{section}

\section*{Appendix guide}

The appendices follow the argument from the readiness label to its use in an answer-timing policy.
The following routes connect each question to the evidence that addresses it.

\begin{itemize}\itemsep2pt\parskip0pt
\item \textbf{What does readiness measure?} Appendix~\ref{app:geometries} defines delivered evidence
coverage; Appendix~\ref{app:raters} compares it with independent judgments of sufficiency and answer
correctness.
\item \textbf{How is the signal measured?} Appendices~\ref{app:probes} and~\ref{app:datasets} specify
the held-out probe protocol, controls, evaluation pools and task provenance.
\item \textbf{Is it specific to the question rather than confidence or correctness?}
Appendix~\ref{app:signal} gives the identical-pixel crossed-question tests, conditional analyses,
uncertainty comparisons and graded-evidence results.
\item \textbf{What generalizes?} Appendices~\ref{app:generality} and~\ref{app:transfer} separate
readability across models and footage from transfer of a usable operating threshold.
\item \textbf{When does it improve answer timing?} Appendix~\ref{app:utility} defines the policy
comparison and relates its gains to task headroom, with a controlled distractor manipulation.
\item \textbf{What do the additional tests establish?} Appendices~\ref{app:rivals}
and~\ref{app:causal} examine released triggers and activation interventions.
Appendices~\ref{app:repro} and~\ref{app:limitations} give reproducibility details and scope limits;
Appendix~\ref{app:related} expands the related work.
\end{itemize}
\section{Streaming geometries and readiness labels}
\label{app:geometries}

Readiness labels measure coverage of the question's annotated evidence in the delivered frames,
before inspecting any model output. This is distinct from \emph{perceived sufficiency}, a judgment
that the frames support an answer, and from \emph{correctness}, the outcome of answering. A question
may be inferable before its annotated evidence appears, or answered incorrectly after it appears.
Appendix~\ref{app:raters} compares these quantities on the same delivered frames.

For a fixed-span window, availability concerns the current input, not everything that has appeared
earlier in the stream. Its label can return to not ready when the evidence leaves the window.
The geometries and coverage conventions below make that distinction explicit.

\subsection{The geometries}
\label{app:geom-zoo}

Every evaluation delivers frames to the model under one of the geometries in
\autoref{app-tab:geometries}. Two properties matter: span and checkpoint placement. A
\emph{growing prefix} $[0,t]$ expands as the stream advances, making elapsed time informative by
construction: the clock reads $0.63$--$0.74$ on the prefix pools and $0.956$ on the shot-ordering
prefixes. A \emph{fixed-span window} $[t-W,t]$ does not expand. The clock reads $0.500$--$0.506$
on the evidence-agnostic policy ladders; fixed-span probing pools can still carry a timing shortcut
through their negative sampling (Appendix~\ref{app:geom-neg}).

Checkpoint placement is a separate issue. Prefix endpoints placed at evidence onset and offset
make the checkpoint index itself informative, so those ladders support policy \emph{ordering}
rather than absolute latency. The reference containment window instead combines a constant span
with an evidence-agnostic grid. Prefix results provide the graded $\mathrm{seen}/K$ accumulation
test and a different frame-delivery regime, but retain their stronger timing confound.

\begin{table}[t]
\centering
\scriptsize
\setlength{\tabcolsep}{2pt}
\begin{tabular}{lllll}
\toprule
geometry & shape & frames & label & what it is for \\
\midrule
growing prefix & $[0,t]$, evidence-relative ends & $32$ & onset, $\mathrm{seen}/K$ & graded accumulation; earliest evidence \\
recent window & $[t-W,t]$, fixed span, $50\%$ overlap & $32$ & containment & first evidence-agnostic ladder \\
\textbf{containment window} & $16$\,s span, free $2$\,s stride & $32$ @ $2$\,fps & containment & the reference format; model axis, utility \\
short window & $2$\,s span, $0.5$\,s stride & $8$ @ $4$\,fps & containment & point-onset footage \\
matched pair & $2$ rows per item, all else fixed & $32$ @ $2$\,fps & availability & availability vs.\ correctness; causal work \\
crossed quad & 2 questions $\times$ 2 windows & $32$ @ $2$\,fps & containment & question-conditioning; visual controls at $0.5$ \\
region window & jittered context cut around evidence & $32$ @ $2$\,fps & containment & the manipulated difficulty contrast \\
sparse bursts & $16$ frames in 3--6 bursts per block & $16$ & availability & sparse multi-burst streaming \\
controlled composite & counter-prior cue on real footage & $16$ @ $2$\,fps & cue present & the existence test, full control \\
\bottomrule
\end{tabular}
\caption{\textbf{Every streaming geometry, its shape, its frame budget, and its label.} The
containment window (bold) is the primary geometry: constant span and an evidence-agnostic checkpoint grid. Sampled
frame count is constant within each geometry, so no readout can substitute ``how many frames'' for
``which frames''. Reported AUROCs are \emph{not} comparable across rows: the rows differ both in how
strict the label is and in how the not-ready rows are drawn (Appendix~\ref{app:geom-neg}), so each
result in \autoref{sec:results} names the row it belongs to.}
\label{app-tab:geometries}
\end{table}

\subsection{Where the coverage threshold sits}
\label{app:geom-contain}

A readiness label is a threshold on delivered coverage, and the threshold can be placed at either end of
the range: at \emph{onset}, where any delivered frame falls inside an evidence segment, or at
\emph{containment}, where the delivered frames cover the whole evidence hull. This paper places it at
containment on every fixed-span geometry, and at onset on the growing prefix, where a prefix that
contains the hull is simply a prefix that has run past the evidence. This subsection reports what the
choice is worth, separating the effect on the \emph{monitor} from the effect on the \emph{oracle},
because they are not the same size.

The containment builder emits three labels per row from a single activation cache, so the same
activations can be scored under each without re-running the model:

\begin{table}[H]
\centering
\footnotesize
\setlength{\tabcolsep}{4pt}
\begin{tabular}{lll}
\toprule
label & definition & used for \\
\midrule
containment & coverage contains the evidence hull; \emph{null} on partial & probe fitting \\
deployed & the same, partial folded into not-ready & policy evaluation \\
onset & any delivered frame inside an evidence segment & threshold sensitivity \\
\bottomrule
\end{tabular}
\caption{\textbf{Three labels from one cache.} Rows with partial coverage are excluded from binary probe
fitting, because their state is genuinely intermediate; a deployed policy may not drop a row, so it
treats them as not ready.}
\label{app-tab:labels}
\end{table}

\paragraph{The monitor barely moves.} On the two natural prefix pools the same frozen activations can be
scored under either threshold. Under the convention used on every window geometry, containment is
required and partial rows are excluded. This reads $0.7224$ over $3{,}486$ rows on NExT-GQA and $0.7292$ over $4{,}333$
rows on ST-Evidence-MCQ, against $0.733$ and $0.722$ at onset: a change of at most $0.011$, in opposite
directions on the two pools. Folding partial rows into not-ready instead, the deployed convention, reads
lower at $0.656$ and $0.637$, because on a prefix every checkpoint between evidence onset and offset then
becomes a not-ready row that already shows part of the evidence. Where the threshold sits changes which
rows count as intermediate; it does not change whether the state is present, and the prefix readings in
\autoref{tab:core} are not an artefact of the choice.

\paragraph{Partial rows are ordered, not ignored.} Nulling partial rows in the binary label does not
leave the readout undefined on them. Scored against fractional hull coverage, a continuous target on
$[0,1]$, the same activations track coverage within an item at Spearman $\rho$ of $0.454$
$[0.424,0.482]$ over $1{,}165$ items on NExT-GQA, $0.490$ $[0.444,0.536]$ over $349$ on ST-Evidence,
and $0.558$ $[0.494,0.619]$ over $214$ on ST-Evidence-Gen, with median item correlations of $0.584$,
$0.659$ and $0.739$, and a positive correlation on $85.3\%$, $86.8\%$ and $87.9\%$ of items. Strictly
partial rows are $15.6\%$, $15.0\%$ and $22.5\%$ of those pools. The nulls hold: a layer-$0$ readout
reads $-0.043$ to $-0.019$ and a within-video shuffle $0.013$ to $0.041$. Elapsed time within an item
reads $-0.081$ and $-0.094$ on the first two pools, so the ordering is not a restatement of the clock
there, although it reads $+0.094$ on the third. The containment ladder over ActivityNet-Captions
builds no partial rows, so this target is degenerate on it and it is left out. A gate crossing these
rows therefore sees a graded score rather than an arbitrary one.

\paragraph{The oracle moves a great deal.} The threshold matters far more for what the task makes
attainable than for what the probe can read, and this is the reason the fixed-span geometries use
containment. On the reference $16$\,s window, moving the label from onset to containment moves the
evidence-oracle advantage from $+2.1$ to $+9.0$\,pp and the attainable band from $3.7$ to $10.5$\,pp;
on the same $1{,}728$ items the latency-matched selection edge over confidence moves from $+1.33$\,pp
$[-0.18,+2.96]$ to $+1.68$\,pp $[+0.12,+3.28]$. These are paired results from one run, not the
$+3.85$\,pp median over the $21$ window configurations quoted in \autoref{sec:utility}. An onset-labelled oracle is allowed to stop at a
checkpoint that has shown part of the evidence and calls it ready, which understates the headroom an
evidence-timing policy is competing for. The band therefore depends on the label convention and supplies
a reference for policy gain rather than a guaranteed upper bound (\autoref{sec:utility}).

\paragraph{How much a threshold at onset actually withholds is a property of the geometry.} Where the
window is narrow, the evidence can routinely outrun it. On the measured pool, a
$4$\,s window faced a median evidence duration of $8.4$\,s, with the evidence longer than the window
on $77\%$ of items. An onset-positive checkpoint
holds a median of only $18.5\%$ of the evidence and just $2.49\%$ of such rows hold all of it, while
answer accuracy tracks delivered coverage closely ($r = 0.957$, $0.9867$ and $0.9968$ on three
benchmarks). On that geometry, an onset positive can therefore be \emph{no} usable evidence rather than weak
evidence. On the paper's reference geometry the same statistic reads very differently: a $16$\,s window
with the hull capped at $12$\,s makes onset rows carry a median of $100\%$ and a mean of $80.2\%$ of the
evidence, and $57.8\%$ of them are full containment. The two labels are therefore close to interchangeable
on the geometry this paper reports and far apart on a narrower one, which is why the numbers above are
small and why the threshold is stated for every result rather than assumed.

\subsection{Where the not-ready rows come from}
\label{app:geom-neg}

On footage whose evidence is effectively an onset rather than an interval, the not-ready rows have to be
drawn from somewhere, and the choice is load-bearing rather than incidental. We expose it as a parameter
and score one frozen activation cache under four regimes: not-ready windows drawn strictly
\emph{before} the evidence, immediately before it (\emph{near}), strictly \emph{after} it, or from
\emph{both} sides.

Before-only sampling makes elapsed time almost sufficient. Every ready row is then late and every
not-ready row early, so the end of the window predicts the label directly: a clock fitted on that pool
reaches AUROC $0.9403$ on one build and $0.9497$ on the frozen manifest. Jittering the boundary does not
remove this shortcut: the pool collapses to 26--69 items and the fitted clock stays between $0.92$ and $1.00$.
Drawing from both sides weakens this timing shortcut, because a not-ready window may now be early or
late, and the same fitted clock falls to $0.5154$ and $0.5271$ respectively. Drawing strictly after the
evidence gives $0.53$.

We therefore report the both-sided regime as primary throughout and use the full sweep to compare the
readiness readout's sensitivity to negative sampling with that of the fitted clock.

\subsection{Clock tracking}
\label{app:geom-clock}

The sweep in Appendix~\ref{app:geom-neg} yields a statistic that separates a readiness readout from a
stopwatch on any substrate where the clock cannot be removed by design. It is read on the
\texttt{before}$\rightarrow$\texttt{both} contrast, the one that moves the clock furthest, with the
\texttt{near} cell excluded as a content control rather than a clock cell:
\[
  \mathrm{clock\ tracking}
  = \frac{\text{probe AUROC drop from \texttt{before} to \texttt{both}}}
         {\text{fitted clock AUROC drop over the same contrast}} .
\]
A readout of elapsed time would move one-for-one with the clock and score near $1$. Measured on one
build the probe moves $0.0335$ ($0.9109 \rightarrow 0.8774$) while the clock moves $0.4249$, giving
$0.0788$; on the frozen manifest the same statistic reads $0.1065$ (probe $0.9190$; clock $0.9497
\rightarrow 0.5271$). Across the seven models of the shared evaluation it ranges from $0.0864$ to
$0.2286$ (Appendix~\ref{app:generality}).

\subsection{Graded evidence}
\label{app:geom-graded}

Where a question's evidence is split into $K$ segments and $K$ varies across items, the raw count of
segments seen is not a readiness axis: one count level mixes opposite states, since ``one of one'' is
ready and ``one of five'' is barely started. The graded target is therefore the fraction
$\mathrm{seen}/K$, which places items with different $K$ on one accumulation scale. On the same pool the
count axis gives a non-monotonic pooled curve ($\rho = 0.133$) and a partial correlation of $-0.009$.
This is a false null: the fraction axis gives $+0.378$. We also report per-$K$ staircases as the honest
per-level view, each with its minority-class count, since a stratum with very few positives cannot
support a stable AUROC.

To ask whether the readout carries information a stopwatch does not, we regress both the probe score and
the graded target on fractional video progress and correlate the residuals, with a video-cluster
bootstrap interval. One caveat travels with this: a partial correlation cannot fully exclude the probe
reading the confounder more finely than the variable used to control for it. That is why the
identical-pixel manipulation, not the partial correlation, is the decisive control
(\autoref{sec:signal}).

Finally, a constant sampled frame count does not by itself guarantee a constant count of \emph{distinct}
frames, because a short early window can repeat source frames. Frame budgets are fixed per geometry
(\autoref{app-tab:geometries}) and graded ladders are constructed so that revealed evidence segments are
actually sampled, with at least two frames per revealed segment on the graded pools.

\section{Independent raters on the readiness label}
\label{app:raters}

The coverage labels of Appendix~\ref{app:geometries} specify where annotated evidence is delivered;
they do not certify the first moment at which an answer can be inferred. Independent raters check
how those labels relate to perceived sufficiency. We compare the probe and raters against the
timestamp label, then score the probe against the raters' own verdicts. Their answers to the
underlying questions provide a separate check on correctness. These comparisons distinguish the
three quantities rather than treating agreement with any one of them as agreement with all three.

\subsection{Instrument}
\label{app:raters-design}

\paragraph{What a rater saw.}
Three raters at our institution, unaffiliated with this work, judged $\mathbf{900}$ distinct screens.
The pool comprises $180$ items, with $60$ each from NExT-GQA, ActivityNet-Captions and Oops!, drawn by a
deterministic digest of the item id. \emph{Each item appears five times}, once per streaming
window, for $180 \times 5 = 900$ separate judgments. The five windows per item span the range the
probe is evaluated over: two before the evidence, one at partial coverage, one at first full
coverage, and one after. A screen carried the question, the answer options, and the exact frames the
frozen model received at that window, re-hashed from disk against the extraction digest so that what
a rater saw is byte-identical to what the model was given. Windows of the same item were not
presented together and carried no marker of their order.

For each screen a rater answered two questions: whether the delivered footage is enough to answer the
question (yes, no, or uncertain), and what the answer is. No rater saw the gold answer, the readiness
label, the checkpoint identity, any timestamp, any frame index, or any model output.

\paragraph{Consensus.}
Two raters judged all $900$ screens independently. A third judged only the $317$ screens on which the
first two disagreed on either question and resolved all of them; $153$ of those disagreements were on
the three-category readiness verdict. The consensus verdict is therefore the first two raters' judgment where they
agree and the third's where they do not; it is a tie-break rather than a three-way vote, and treating
it as a majority over three full passes would mix two different measurement processes. Annotation
took $12.6$ hours in total.

\paragraph{The question is well posed.}
Before adjudication, the two primary raters agree on the three-category readiness verdict
(yes, no, uncertain) on $83.0\%$ of all $900$ screens, with Cohen's $\kappa$
\citep{cohen1960agreement} of $0.717$ pooled and $0.700 / 0.771 / 0.680$ on NExT-GQA,
ActivityNet-Captions and Oops!. The third rater does not enter either agreement measure. This is
substantial agreement between people who never conferred, on a judgment neither of them was given
a rule for.

\subsection{The probe reads the annotation at rater level}
\label{app:raters-parity}

\begin{table}[H]
\centering
\small
\setlength{\tabcolsep}{4pt}
\begin{tabular}{lccc}
\toprule
scored against the benchmark's evidence label & NExT-GQA & ActivityNet-Cap. & Oops! \\
\midrule
AUROC, readiness probe            & $0.671$ $[.607,.733]$ & $0.752$ $[.693,.804]$ & $0.791$ $[.712,.860]$ \\
AUROC, independent rater verdict  & $0.778$ $[.731,.822]$ & $0.681$ $[.628,.734]$ & $0.722$ $[.655,.787]$ \\
\midrule
rater accuracy                    & $0.760$ & $0.697$ & $0.733$ \\
rater $\kappa$                    & $0.527$ & $0.364$ & $0.385$ \\
rater sensitivity / specificity   & $0.687 / 0.868$ & $0.600 / 0.761$ & $0.700 / 0.744$ \\
screens ($n$)                     & $300$ & $300$ & $240$ \\
\bottomrule
\end{tabular}
\caption{\textbf{A person shown the delivered frames recovers the benchmark's own evidence label
about as well as the probe does.} Intervals overlap on all three evaluations, with the raters ahead
on NExT-GQA and behind on the other two. The $240$ on Oops! is because $60$ of its $300$ screens sit
at partial coverage and carry no binary label there; the rater-verdict comparisons in
\autoref{app-tab:rater-target} use all $300$.}
\label{app-tab:rater-parity}
\end{table}

These results highlight the distinction between timestamp coverage and perceived sufficiency, rather
than establishing a ceiling on representation performance. The comparison is descriptive: a binary
rater verdict has AUROC $(\mathrm{sensitivity} + \mathrm{specificity})/2$, while the probe contributes
a continuous score. Raters disagree with the annotation on $24$--$30\%$ of screens, at
$\kappa = 0.364$--$0.527$.

\subsection{The rater verdict as the target}
\label{app:raters-target}

Turning the comparison around and scoring each candidate against what the raters said gives a
readiness target that owes nothing to a timestamp convention.

\begin{table}[H]
\centering
\small
\setlength{\tabcolsep}{5pt}
\begin{tabular}{lccc}
\toprule
scored against the raters' readiness verdict & NExT-GQA & ActivityNet-Cap. & Oops! \\
\midrule
AUROC, readiness probe               & $\mathbf{0.871}$ $[.832,.909]$ & $\mathbf{0.944}$ $[.916,.967]$ & $\mathbf{0.923}$ $[.891,.949]$ \\
AUROC, the model's own confidence    & $0.643$ & $0.691$ & $0.674$ \\
AUROC, an external LLM asked directly & $0.629$ & $0.594$ & $0.576$ \\
\midrule
model accuracy, rater-ready screens     & $81.7\%$ & $71.6\%$ & $62.6\%$ \\
model accuracy, rater-not-ready screens & $65.2\%$ & $34.6\%$ & $34.6\%$ \\
screens ($n$), probe and confidence     & $300$ & $300$ & $300$ \\
screens ($n$), external LLM             & $262$ & $256$ & $252$ \\
\bottomrule
\end{tabular}
\caption{\textbf{The readout tracks a human readiness verdict; the model's own confidence and an
external LLM asked the question directly do not.} All three rows are continuous scores on the same
screens. The lower block shows the raters' verdict is not a private opinion: the frozen model answers
correctly $16.5$, $36.9$ and $28.0$ pp more often on the screens they called ready.}
\label{app-tab:rater-target}
\end{table}

\subsection{Asking a model directly}
\label{app:raters-judge}

The third row of \autoref{app-tab:rater-target} is a baseline, not a second validation of the label.
An external multimodal LLM (\texttt{gemini-3-flash-preview}) was shown the same frames as the raters
and asked the same question, with three independently seeded repetitions per screen over a $600$-item
superset of the rater pool; $8{,}881$ of the $8{,}907$ queries issued completed, and the comparison above
uses the subset the raters also judged. The judge answers consistently across its repetitions, with Fleiss'
$\kappa$ \citep{fleiss1971agreement} of $0.698$ pooled and $0.673$--$0.734$ per evaluation, so the low agreement with people is not
query noise. It is also not a matter of calibration: the judge is lenient, calling $46$--$81\%$ of
pre-evidence windows sufficient, but the AUROCs above are computed from its continuous rate of
answering ``enough'' and are therefore threshold-free. Removing its yes-bias entirely leaves it
ordering screens the way people do at $0.576$--$0.629$. The probe's advantage over overt responses
(\autoref{sec:notconf}) therefore extends to this external judge under the tested configuration.

Two bounds on that claim. This is one judge in one configuration, run with a small reasoning budget,
so it bounds what asking \emph{this} model directly achieves rather than what any such judge could;
and the judge is zero-shot while the probe is fitted, the same asymmetry that limits the comparison
with released triggers in \autoref{sec:rivals}.

\subsection{What the raters' own answers show}
\label{app:raters-answers}

Raters also answered the question on every screen, which tests the window construction independently
of any stated readiness judgment. Their accuracy tracks the labelled states, rising into the
evidence-containing windows and, where the design moves the evidence back out of a fixed-span window,
falling again after it.

\begin{table}[H]
\centering
\small
\setlength{\tabcolsep}{6pt}
\begin{tabular}{lcccccc}
\toprule
rater answer accuracy & pre (far) & pre (near) & partial & first ready & post & overall \\
\midrule
NExT-GQA           & $0.700$ & $0.717$ & $0.867$ & $0.917$ & $0.950$ & $0.830$ \\
ActivityNet-Cap.   & $0.433$ & $0.450$ & $0.883$ & $0.833$ & $0.483$ & $0.617$ \\
Oops!              & $0.733$ & $0.933$ & $0.950$ & $0.983$ & $0.883$ & $0.897$ \\
\bottomrule
\end{tabular}
\caption{\textbf{The windows sit where the design says they do.} Accuracy rises into the
evidence-containing windows on all three evaluations. It stays high in the post window on NExT-GQA,
where the delivered frames still contain the evidence, and falls on the other two, where by that
window the evidence has left the span. This reproduces the non-monotonicity of the labels themselves.}
\label{app-tab:rater-answers}
\end{table}

\subsection{Limits of this check}
\label{app:raters-limits}

Each evaluation contributes $60$ items at five windows, so the individual per-evaluation gaps carry
wide intervals; what we claim is the consistent direction across all three, not the size of any one
gap. ActivityNet-Captions is the hardest pool for raters, at $0.617$ own-answer accuracy against
$0.830$ and $0.897$, and is correspondingly the noisiest row above. The raters are not reproducing
the model: their answers coincide with the model's on $0.653$, $0.433$ and $0.443$ of screens. Three
raters at one institution are not a demographically sampled population, and the readiness question
was posed in English over multiple-choice items, so this establishes that the target notion is legible
and reproducible between independent people rather than that it is universal. Finally, this is a check
on the \emph{label}, run on three of the paper's evaluations; it does not extend to the controlled
composites, the model axis, or the transfer results, whose labels are validated only by construction.

\section{Probes, cross-validation, and controls}
\label{app:probes}

This appendix expands the probe protocol of \autoref{sec:method} and gives the full control battery.

\subsection{Where we read}
\label{app:probe-extract}

After each decoder layer, every token position holds a hidden state: the running residual the layer
read from and wrote to. It is $4096$-dimensional for Qwen3-VL-8B. We write $L$ for the residual stream after
layer $L$, so that $L=0$ denotes the raw input embeddings, before any transformer layer has run.

Four readout positions are captured. The \emph{final} prompt token, from which the model is about to
answer, is the primary position throughout. The \emph{option}-letter tokens and the pooled \emph{vision}
tokens provide complementary views, and on the controlled stimuli, where the question names a specific
object, the \emph{object-name} tokens are available as well. Comparing positions and layers localises
where a signal lives: readiness is at chance at layer $0$ at every language position and rises with
depth, and it is carried by the language stream rather than by the pooled visual representation.

Layers are snapshotted at fixed fractions of decoder depth ($0.40$, $0.50$, $0.65$, $0.80$ and $0.90$),
plus layer $0$. We use fractions rather than absolute indices because depth differs across architectures
(Qwen3-VL-8B and InternVL3.5-8B have $36$ decoder layers, Qwen3-VL-32B $64$, LLaVA-OneVision $28$).
Cross-model comparisons are therefore made at matched depth fractions. The depth fractions reported
for the crossed-question replications in Appendix~\ref{app:gen-layers} rescale this grid so the deepest
probed layer sits at $1.0$, which is why the shallowest sampled site reads $0.44$--$0.45$ rather than
$0.40$.

\subsection{Probe and evaluation}
\label{app:probe-cv}

The probe is an L2-regularized logistic regression fitted on top of the frozen model's activations to
predict the objective readiness label. Features are standardized to zero mean and unit variance using
\emph{training-fold statistics only}. The model's weights are never touched. Because the probe is
linear, it can exploit only a single direction in activation space, so success means the information is
linearly decodable. The structure was already computed by the network, not taught to the probe.

We report held-out ROC AUROC, the threshold-free probability that a randomly chosen ready example outscores
a randomly chosen not-ready one. AUPRC accompanies it wherever the classes are imbalanced, and expected
calibration error is reported wherever an operating threshold is under discussion, since a ranking result
and an operating-point result are different claims (Appendix~\ref{app:transfer}). For graded targets we
use ridge regression on the count directly, reported as mean held-out prediction per true level alongside
Spearman $\rho$, mean absolute error, and exact and within-one accuracy.

Within-benchmark generalization holds out whole groups: \texttt{video\_id} on natural footage and
the cue object for the controlled crossed-composite headline. An inner held-out-group loop selects
the readout layer and regularization, so test groups influence neither standardization nor model
selection. \autoref{app-tab:protocols} distinguishes these fits from the explicit transfer tests.

On natural matched-pair and crossed-quad designs, \textbf{both members of a pair, and all four
members of a quad, remain in one fold}. Splitting byte-identical twins can produce an inverted
visual readout rather than a valid null. The controlled composites are the explicit exception:
holding out cue objects separates questions about different objects on the same frames. Their
pair-preserving visual null is therefore a separate evaluation, not the grouping used for the
headline probe (Appendix~\ref{app:probe-quad}).

\begin{table}[t]
\centering
\footnotesize
\setlength{\tabcolsep}{3pt}
\begin{tabular}{@{}
	>{\raggedright\arraybackslash}p{3.1cm}
	>{\raggedright\arraybackslash}p{2.3cm}
	>{\raggedright\arraybackslash}p{7.0cm}
@{}}
\toprule
evaluation & held-out unit & fitting and scoring \\
\midrule
controlled-composite headline & cue object & Nested within-pool fitting; evaluation on held-out objects. \\
controlled-composite visual null & pair-preserving group & Both questions on identical frames share a fold; distinct from the headline protocol. \\
natural within-benchmark probes, including the model axis & video & Video-grouped inner selection and outer evaluation; matched pairs and crossed quads stay together. \\
held-out-family transfer & footage family & Selection on source families only; the target family's footage is excluded from fitting. \\
external transfer & disjoint source and target videos & Source scaler, layer and probe are frozen on the target; a target-fitted probe is a separate comparison. \\
\bottomrule
\end{tabular}
\caption{\textbf{Which groups are held out, and where fitting occurs.} The geometry, label and scored
row counts for the within-benchmark representation results are listed together in
\autoref{app-tab:decode}. \autoref{app-tab:pools} inventories the source and builder pools, whose
counts can differ from the scored subset. Transfer details are in Appendices~\ref{app:generality}
and~\ref{app:transfer}.}
\label{app-tab:protocols}
\end{table}

Confidence intervals resample whole groups with replacement $1{,}000$ times and take the $2.5$--$97.5$
percentiles of the recomputed statistic. Ordinary error bars would understate uncertainty because examples
are clustered, with many windows per video and many items per object. The same machinery gives intervals
for Spearman and partial correlations.

One disclosure belongs with every number in this paper: the pipeline was run at a \textbf{single seed}.
The reported intervals are cluster bootstraps conditional on the fitted cross-validation splits.
A same-seed, same-host re-run reproduces a
published AUROC to about three decimal places: $0.8055$ against a recorded $0.8056$ on one check.
The residual drift is attributable to BLAS threading rather than to the seed. We therefore quote
probe AUROCs to three decimals and treat fourth-decimal differences as noise.

Concretely, the classifier is a \texttt{StandardScaler}\,+\,\texttt{LogisticRegression} pipeline with
an L2 penalty, \texttt{class\_weight="balanced"} and \texttt{max\_iter}$=3000$, searching
$C \in \{0.01, 0.03, 0.1, 0.3, 1, 3, 10\}$. The nesting is a $5$-fold outer \texttt{GroupKFold} over
videos with a $3$-fold inner \texttt{GroupKFold} inside each outer training fold; the inner criterion
is a single AUROC pooled over the inner folds' out-of-fold predictions rather than a mean of per-fold
AUROCs. \texttt{GroupKFold} performs no shuffling, so for a given grouping column the splits are
deterministic and there is nothing to seed at the fold-assignment level.

\subsection{The control battery}
\label{app:probe-controls}

Probe performance needs controls for what the probe itself can learn \citep{hewitt2019designing}.
Some comparisons sit at chance by construction and only verify the
wiring; the informative ones are the baselines that \emph{can} carry real signal.

\paragraph{Layer-0 cells are two different objects.}
At the final language position, layer $0$ is the same embedded token on every row of an item, so it is at
exactly $0.500$ by construction. This standard control is worth reporting because its exact value of
$0.500$ is verifiable. But it cannot fail and should never be offered as evidence on its
own. The pooled-vision position is different in kind: its layer-$0$ embedding already carries raw visual
features, so on inputs where ready and not-ready conditions contain different pixels it can and does
discriminate them. It reads $0.686$--$0.819$ under before-only negative sampling and
$0.579$--$0.683$ once the clock is balanced. It is a measured baseline to be exceeded, not a null.

\paragraph{The shuffled-label null is measured, not assumed.}
Permuting readiness labels within groups and re-probing collapses a real signal to chance. The resulting
value is not automatically $0.500$: where the readiness base rate varies by source and source is itself
decodable, the within-group shuffle sits above chance. It reaches $0.578$ on the graded pool, so the
graded result there is read as ``$0.827$ against $\approx 0.58$'' rather than against $0.500$. Every probe
AUROC in this paper is accompanied by its own measured shuffle.

\paragraph{Timing, confidence, and nuisance baselines.}
A readout must beat what a stopwatch achieves: elapsed time, and on graded data fractional video progress,
which is the primary timing control there because absolute seconds is only a coarse proxy across
variable-length videos. The model's own answer confidence is fitted as a competing predictor. Beyond
these we decode answer letter, answer content, question type, correctness and source domain from the same
activations, to record what else is linearly available. These are not always clean nulls: where item
identity is constant across an item's rows while readiness varies, they are inflated by construction.
We therefore report them as measurements rather than as controls.

Two further baselines are model-independent. A pixel-statistics probe over luminance, contrast and motion,
fitted under the same cross-validation, sits at $0.5761$--$0.6590$ across the four negative-sampling
regimes of the seven-model evaluation, and the weakest model cell there clears its own baseline by $+0.16$.
On a variable-$K$ benchmark the evidence-segment count alone predicts readiness at $0.58$--$0.70$, so it is
reported beside any binary readiness AUROC on that pool.

\paragraph{The build-time placement audit.}
Synthesising the input geometry determines which burst falls in which block, which window becomes a
negative, and which context region is cut. These choices can themselves predict the label. Before any GPU time
is spent, the builder computes the AUROC of every input-visible nuisance column against the label and
\emph{refuses to write the manifest} if one exceeds its threshold. The audit tests $|\mathrm{AUROC}-0.5|$
rather than AUROC, because an inverse leak is still a leak: a burst-count column at $0.428$ is as
informative as one at $0.572$. A column that fails is fixed in the sampler, never by relaxing the
threshold, and the general remedy is to make a nuisance property constant \emph{within} an item rather
than randomising it per row.

Some columns must leak by construction: a growing prefix carries a clock, and an external benchmark's own
grid carries its anchoring. Those are moved from gated to reported \emph{by name} in the configuration and
echoed into the audit artefact, while every other column stays gated at its full threshold. The probing
pool and the policy ladder are audited separately and can differ enormously, so a clock value always names
which of the two it was measured on. On the crossed design the audit is set $20\times$ tighter than usual
($0.005$ rather than $0.10$), because both rows of a crossed pair share one window and any window-only
column is arithmetically $0.500$; it met that bar at exactly $0.0$.

\subsection{The crossed-quad guarantee}
\label{app:probe-quad}

For two questions $q_1, q_2$ on one video whose evidence hulls are \emph{disjoint}, we construct window
$W_1$ containing $\mathrm{hull}(q_1)$ and disjoint from $\mathrm{hull}(q_2)$, and $W_2$ the reverse, with a
$0.5$\,s separation margin enforced between hull and window boundary. All four $(\text{window},
\text{question})$ rows are emitted. Both orderings are required: a pool in which only one of the two
questions is ever the contained one is not admissible, because the question that can be contained is
systematically the shorter-hulled one and the resulting position skew cannot be removed by construction.

Writing $s(W,q)$ for the probe's readiness probability on window $W$ under question $q$, and $W_i$ for
the window containing the evidence for $q_i$, the \emph{difference-in-differences} (DiD) is the
$2\times2$ interaction
\begin{equation}
\mathrm{DiD} \;=\; \bigl[\,s(W_1,q_1) - s(W_1,q_2)\,\bigr] \;-\; \bigl[\,s(W_2,q_1) - s(W_2,q_2)\,\bigr],
\label{eq:did}
\end{equation}
averaged over quads. The design yields four reported quantities: held-out AUROC, the \emph{flip rate}
(how often changing only the question flips which row scores higher), the DiD of
\autoref{eq:did}, and the model's own accuracy on the contained question against the other, which is
$+0.1402$ $[0.1159, 0.1634]$ at no extra forward pass.

Its decisive property is algebraic rather than empirical. Each window appears once as ready and once as
not ready, so the positive and negative sets receive the \emph{same multiset} of windows. Any statistic
that is a function of the video alone therefore receives identical inputs under both labels. This includes
elapsed time, motion, brightness, segment count, pooled vision, and the visual layer-$0$ cell; each is \emph{forced} to
AUROC $0.500$. This is a design guarantee, not a measurement, and it is a stronger claim than clearing an
empirical baseline: the objection ``perhaps the probe reads whether anything salient is visible'' has nothing
left to attach to. A single question swap does not have this property, because it can be carried by a
fixed difference between two question embeddings; the $2\times2$ design subtracts that fixed difference
and then asks whether it reverses with the evidence-containing window.

The guarantee is conditional on the fold assignment: both members of a pair must fall in one fold, or
the two rows are scored by different fold models and the tie that forces $0.500$ is broken. This holds
on every natural crossed pool, where folds group by video and a pair shares its video. It does not hold
on the controlled composites, whose held-out-object protocol asks the two members of each pair about
different objects and so separates them; there the visual layer-$0$ cell is measured rather than forced
and reads $0.069$, which is as far from chance as $0.931$ since a linear readout is free to flip its
sign. \autoref{tab:core} therefore quotes that cell under a pair-preserving grouping, where it is
$0.500$.

\begin{table}[H]
\centering
\scriptsize
\setlength{\tabcolsep}{3pt}
\begin{tabular}{lccccccccc}
\toprule
& & \multicolumn{3}{c}{layer $0$} & & & & & \\
\cmidrule(lr){3-5}
evaluation & probe & final & vision & option & shuffle & clock & conf. & resid.\ drop & $\Delta$ over conf. \\
\midrule
NExT-GQA                     & $0.733$ & $0.500$ & $0.544$ & $0.500$ & $0.502$ & $0.633$ & $0.573$ & $0.0001$   & $+0.106$ \\
ST-Evidence-MCQ              & $0.722$ & $0.500$ & $0.552$ & $0.500$ & $0.484$ & $0.628$ & $0.589$ & $-0.0001$  & $+0.088$ \\
ST-Evidence-Gen              & $0.827$ & $0.500$ & $0.651$ & $0.507$ & $0.578$ & $0.722$ & $0.622$ & $0.0001$   & $+0.180$ \\
\quad the streaming cohort & $0.808$ & $0.500$ & $0.602$ & $0.485$ & $0.529$ & $0.737$ & $0.626$ & $0.0000$   & $+0.164$ \\
HERBench shot-ID             & $0.918$ & $0.500$ & $0.534$ & $0.500$ & $0.498$ & $0.505$ & $0.738$ & $0.0000$   & $+0.155$ \\
HERBench (shot ordering)     & $0.980$ & $0.500$ & --      & --      & $0.503$ & $0.956$ & $0.763$ & $0.0001$   & $+0.212$ \\
ActivityNet-Captions         & $0.764$ & $0.500$ & $0.517$ & $0.500$ & $0.493$ & $0.505$ & $0.579$ & $-0.0002$  & $+0.096$ \\
Oops!                        & $0.874$ & $0.500$ & $0.654$ & $0.500$ & $0.496$ & $0.527$ & $0.642$ & $-0.0001$  & $+0.108$ \\
external benchmark           & $0.748$ & $0.500$ & --      & --      & $0.509$ & $0.451$ & $0.350$ & --         & --       \\
\bottomrule
\end{tabular}
\caption{\textbf{The dissociation battery.} One row per evaluation, each read against its own
comparisons; values are \emph{not} comparable down a column, because the rows differ in geometry and in
label strictness (Appendix~\ref{app:geom-zoo}). The three layer-$0$ cells are different objects
(Appendix~\ref{app:probe-controls}): at the final language position, and at the option tokens wherever
the option count is constant, layer $0$ is the same embedded state on every row and is forced to
$0.500$, whereas the vision cell is a baseline that can carry real signal. The two option cells that depart from
$0.500$ belong to the pools that mix two- and four-option items, where that cell becomes an indicator
of the option count rather than a control. A dash marks a cell not available on that pool; such a cell
is never assumed to be $0.500$. For shot ordering, the activation cache was not retained. The clock column names a different pool in each row: it is
high wherever evidence onset is strongly time-correlated, which is why those rows carry the
qualification stated in Appendix~\ref{app:signal-decode} and are not read as readiness results on their
own. ``Conf.'', ``resid.\ drop'' and the last column, which gives the AUROC gain of the readiness score over a
confidence-only baseline, all come from the pooled confidence-separation estimator on the same rows
(\autoref{app-tab:conf}), which is why the confidence figure here can differ in the third decimal from
the fitted baseline in \autoref{tab:core}. Every row uses the pixel-cleaned pool that
\autoref{sec:results} reports on. The indented row is the same graded shot-boundary prefix ladder as
the row above it, restricted to the streaming eligibility cohort; it is a growing prefix, which is why
its clock sits with the other prefixes.}
\label{app-tab:battery}
\end{table}

\section{Datasets, pools, models, and prompts}
\label{app:datasets}

This appendix gives the data behind every evaluation: the benchmark families and the pools built from
them, what footage they share, the models run, the frame budgets and pixel quality control applied,
and what is documented about the prompt. The streaming geometries themselves are in
Appendix~\ref{app:geometries} and are not repeated here.

\subsection{Benchmark families and evaluation pools}
\label{app:data-pools}

\autoref{app-tab:pools} inventories the source and builder pools. These are not always the rows
used to fit or score a probe: filtering and negative sampling can select a subset. Use
\autoref{app-tab:decode} for each representation result's geometry, label and scored row count,
and \autoref{app-tab:protocols} for its fitting and grouping protocol. In particular, the
composites' source-video count is not their held-out-object evaluation unit.

{\scriptsize
\setlength{\tabcolsep}{3pt}
\begin{longtable}{@{}
  >{\raggedright\arraybackslash}p{1.95cm}
  >{\raggedright\arraybackslash}p{1.7cm}
  >{\raggedright\arraybackslash}p{2.15cm}
  >{\raggedright\arraybackslash}p{1.15cm}
  >{\raggedright\arraybackslash}p{1.15cm}
  >{\raggedright\arraybackslash}p{1.05cm}
  >{\raggedright\arraybackslash}p{1.25cm}
  >{\raggedright\arraybackslash}p{1.35cm}
@{}}
\caption{\textbf{Source and builder pools, evidence annotations, and sizes.} ``Rows'' counts
delivered checkpoints, not items. Readiness gives ready\,/\,not-ready counts where recorded and
marks the balanced matched and crossed designs. Source groups describe the listed pool, not
necessarily the split or bootstrap key. A dash marks a value not recorded. Pixel filtering and
its pool-specific effects are detailed in Appendix~\ref{app:data-frames}; scored representation
pools are identified in \autoref{app-tab:decode}.}
\label{app-tab:pools}\\
\toprule
pool & footage & evidence timestamps & items & rows & source groups & readiness & model acc. \\
\midrule
\endfirsthead
\toprule
pool & footage & evidence timestamps & items & rows & source groups & readiness & model acc. \\
\midrule
\endhead
controlled composites, crossed & HD-EPIC egocentric, P01--P09 & cue presence by construction, 17-object library & 260 (520 prefixes) & 1,040 & 31 videos, 90 clips & 0.5 by constr. & -- \\
matched \present/\absent\ windows & HD-EPIC egocentric, P01--P09 & official fine-grained action segments & 2,400 pairs & 4,800 & 142 videos & 0.5 by constr. & -- \\
fine-grained action VQA & HD-EPIC egocentric & the benchmark's own items & 1,200 & -- & 86 videos & -- & -- \\
NExT-GQA prefixes & NExT-QA & human temporal grounding & 1,000 & 5,695 & 473 videos & 3,600 / 2,095 & 0.729 five-opt., 0.695 six-opt. \\
NExT-GQA crossed quads & NExT-QA & human temporal grounding & 1,208 quads (1,196 intact) & 4,808 & 277 videos & 0.5 by constr. & -- \\
ST-Evidence-MCQ prefixes & NExT-QA & independent human evidence & 1,298 & 6,756 & 348 videos & 4,400 / 2,356 & 0.704 \\
ST-Evidence-MCQ crossed quads & NExT-QA & independent human evidence & 311 quads & -- & 88 videos & 0.5 by constr. & -- \\
ST-Evidence-Gen prefixes & pt, star, vicas, ego4d & released segments, $K{=}1$--$11$ & 1,643 & 4,346 & 1,529 videos & 2,894 / 1,452 & 0.702 \\
ST-Evidence-Gen containment windows & pt, star, vicas, ego4d & released segments, evidence-agnostic clock & 1,643 (1,244 at 8 frames) & 92.6k over 5 formats & 1,529 videos & -- & 0.586--0.677 \\
containment rebuild, three benchmarks & NExT-QA; pt, star, vicas, ego4d & released segments, hull $\le 12$\,s & 1,728 & 37,129 & 588 clusters & -- & -- \\
shot-ordering prefixes & movie trailers & shot timestamps, $K{=}4$ constant & 2,144 & 12,582 & 87 trailers & -- & 0.227 five-way \\
shot-identification bursts & movie trailers, the same 87 & shot timestamps and per-shot descriptions, $K{=}1$ & 956 with a full ladder & 7,648 per variant & 86 trailers, 77 after filtering & -- & 0.887 delivered, 0.223 not \\
ActivityNet windows & ActivityNet-Cap\-tions & caption spans & 600 per arm & 16,109 & 447 clusters & -- & -- \\
ActivityNet held-out cell & ActivityNet-Cap\-tions & caption spans & 250 & 6,664 & 179 clusters & -- & -- \\
Oops!\ short windows & Oops!\ compilation clips & human failure-onset instants & 312 & 6,475 (+1,248) & 312 clusters & -- & -- \\
OVO-Bench, external & COIN tutorials, MovieNet films & the benchmark's own steps and clue instants & 145 + 48 & 11,086 & 42 + 10 videos & -- & not comparable \\
\bottomrule
\end{longtable}
}

Readiness positive rates vary widely across pools and are reported with each result rather than
assumed: $0.500$ by construction on the matched-pair and crossed designs, $0.63$--$0.67$ on the three
prefix benchmarks, $0.34$ on the shot-ordering pool, and $0.13$--$0.19$ on the short-window
substrates and the external benchmark. Two pools have no single scalar rate, being intrinsically
multi-cell; their components are given where they are used.

\paragraph{Controlled composites on egocentric footage.}
Ninety clean HD-EPIC cooking windows from $31$ videos (P01--P09) carry a rendered cue object in a
\emph{counter-prior} colour, so the question (``what colour is the object?'') cannot be answered from
the object's usual colour. The cue library holds $17$ objects, each with a canonical variant in its
real colour alongside its recolours. An instance takes two of those objects and one shared clip and
renders a prefix for each at matched cue salience; both colour questions are then asked of both
prefixes, so over byte-identical frames the queried object is present in half the rows and absent in
the other half. The $260$ instances give $520$ prefixes on this design. Object identity is crossed with
answer colour, canonical and counter-prior variants are balanced, and option order is shuffled per
item. The library's food-like objects, which could occur naturally in kitchen footage, are screened
against the clips by asking the frozen model whether the object appears in the clean window; this
excludes $41$ of the $540$ (object, clip) pairs the screen covers. This checks the background's
content; it never filters on correctness. The headline crossed-question probe holds out cue objects,
not videos. The pair-preserving grouping used for its visual null is reported separately
(\autoref{app-tab:protocols}). The evidence annotation here is exact rather than estimated: the queried object is
composited across the whole observed prefix or not at all, so readiness is set by construction.

\paragraph{Grounded and evidence-backed VideoQA on everyday video.}
NExT-GQA extends NExT-QA \citep{xiao2021nextqa} with human temporal grounding \citep{xiao2023trust}: five reasoning types,
five options, and an annotation that may be several disjoint segments ($891$ of the sampled $1{,}000$
questions are single-window). Videos run a median $35.5$\,s, with evidence from a median $9.8$\,s to
$19.0$\,s. ST-Evidence ships human-verified spatio-temporal evidence per question--answer pair
\citep{wang2026evidence}; we use its multiple-choice task and its temporal segments only, and its items
carry more multi-segment evidence, which is why its crossed-quad pool is proportionally larger. In both
the annotation marks where a human looked to justify the answer, and readiness thresholds the delivered
fraction of it.

\paragraph{The generation companion, and the two filters it needs.}
The generation companion covers five sources with released segments and per-item counts from $1$ to
$11$, and is the only natural pool where the graded $\mathrm{seen}/K$ axis has real support. Two item
filters run before anything else: a lexical filter removes questions whose wording leaks \emph{where} a
segment sits (``first'', ``last'', ``finally''; ``before'' and ``after'' are kept, being relative
order rather than fixed position), dropping $376$ items, and a $10.0$\,s minimum-duration floor drops a
further $686$ and removes one source entirely, whose items are all the same length and whose sub-second
segments cannot meet the guarantee of two frames per revealed segment. The four that remain are
Perception Test \citep{patraucean2023perception}, STAR \citep{wu2021star}, ViCaS \citep{athar2025vicas}
and a trimmed QaEgo4D \citep{di2024grounded}, listed as \texttt{pt}, \texttt{star}, \texttt{vicas} and
\texttt{ego4d} in \autoref{app-tab:pools}. Per-source readiness positive
rates then run from $0.573$ to $0.903$, which is why this pool's within-group shuffle null sits at
$0.578$ rather than $0.500$ and why its decoding result is quoted against that null.

\paragraph{Multi-evidence questions on movie trailers.}
The temporal shot-ordering task names four scenes and asks for their chronological order, one of five
options being ``two or more descriptions do not match'' \citep{benami2025herbench}; the annotation is
the four shot windows, so evidence is constant at $K{=}4$. One structural property bounds it: the shots
fall roughly one per video quarter and the evidence spans nearly the whole trailer (median start
$22.6$\,s, end $120.7$\,s of $149$\,s), so delivered fraction and revealed-shot count are close to the
same variable. The shot-identification task on the same $87$ trailers asks instead which described shot
has been seen, which makes evidence one segment per item and gives the burst ladder exactly one ready
checkpoint per item.

\paragraph{Caption spans, and failure onsets.}
ActivityNet-Captions supplies dense caption spans over YouTube footage \citep{krishna2017dense}, from
which five-way items are constructed with the caption span as the evidence annotation; two distractor
arms are built over \emph{identical pixels}, one from the same action class and one uniform, making
question difficulty a manipulated rather than an observed variable. Oops!\ annotates the instant an
intentional action goes wrong, from three annotators per clip \citep{epstein2020oops}, and is the only
pool whose evidence is a \emph{point} rather than an interval, which motivates its short, dense window. Onsets
sit at a median $3.50$\,s and $0.45$ of the clip (p5--p95 fractional $0.10$--$0.88$), wide enough that
no fixed-time policy can lock onto them, and items are kept only where annotator disagreement is at most
$0.5$\,s. Its multiple-choice task is constructed rather than shipped
(Appendix~\ref{app:constructed}). Because these are compilation trims, $40.0\%$ of
the pool carries a hard cut, so the delivered region is the detected shot containing the onset, and
over-long shots are dropped rather than trimmed because trimming would place a synthetic edge at a fixed
offset from the onset.

\paragraph{An external streaming benchmark.}
OVO-Bench is used only as a zero-shot transfer target, with its own prompt text carried over verbatim
\citep{li2025ovobench}. Its two relevant tasks carry its own labels, a step interval and a single clue
instant. Its format is Yes/No, so its trivial baseline is not chance and its accuracy is not comparable
with the multiple-choice pools: on its own step rungs, always answering Yes scores $0.6689$ against the
model's $0.6290$, and on the evidence-agnostic clock ladder always answering No scores $0.8694$ against
the model's $0.8025$.

\subsection{Constructed items: what we authored, and how}
\label{app:constructed}

Four of our evaluations pose questions that do \emph{not} ship with the source dataset. In each case
the source supplies footage and a timestamped annotation: caption spans, action segments, shot
windows, or failure onsets. We author the question, the answer and the distractors from that annotation.
\autoref{app-tab:provenance} states which is which, and the paragraphs below give each construction.
Nothing about the readiness label depends on this: labels come from the source's timestamps in every
case. What depends on it is the accuracy numbers, which are accuracy on \emph{our} items wherever
the provenance column says constructed.

\begin{table}[H]
\centering
\scriptsize
\setlength{\tabcolsep}{4pt}
\begin{tabular}{@{}
  >{\raggedright\arraybackslash}p{4.3cm}
  >{\raggedright\arraybackslash}p{1.9cm}
  >{\raggedright\arraybackslash}p{6.4cm}
@{}}
\toprule
evaluation & questions & source of the items \\
\midrule
HD-EPIC controlled composites & constructed & synthetic cues composited onto real footage \\
HD-EPIC matched \present/\absent & constructed & from the official fine-grained action segments \\
HD-EPIC fine-grained action VQA & shipped & the benchmark's own items, read verbatim \\
NExT-GQA & shipped & NExT-QA's questions with human grounding \\
ST-Evidence-MCQ & shipped & the benchmark's own QA and evidence \\
ST-Evidence-Gen & shipped & the benchmark's own items, two item filters applied \\
HERBench temporal shot ordering & shipped & the released task, read verbatim \\
HERBench shot identification & constructed & re-posed from the released shot timestamps \\
ActivityNet-Captions & constructed & from dense caption spans \\
Oops! & constructed & from the failure-onset and free-text annotations \\
OVO-Bench & shipped, external & its own prompts and labels, carried verbatim \\
\bottomrule
\end{tabular}
\caption{\textbf{Task provenance.} ``Constructed'' means we authored the question, the answer and the
distractors from the source's annotation; the footage and the evidence timestamps are always the
source's. Readiness labels are derived from source timestamps throughout.}
\label{app-tab:provenance}
\end{table}

\paragraph{ActivityNet-Captions.}
One item per caption span, with one fixed question, \texttt{Which event happens in this clip?}. The
correct option is the caption sentence itself, changed only by stripping whitespace; the four
distractors are other caption sentences, so answer and distractors have identical provenance and
identical surface form. The arms differ only in the pool they are drawn from: the same-action-class arm draws
from other videos carrying the same ActivityNet-200 label set. The class attaches to the video, the
only granularity the release offers, giving $208$ distinct class keys. The uniform arm draws
from the whole split. Both exclude the target's own video, require the candidate to fall within $30$
characters of the target's length, and reject any candidate whose content-word Jaccard similarity to
the target exceeds $0.5$; four are then taken from a per-item seeded shuffle. \textbf{No model is
involved at any point in this build}, and that lexical screen is the only check that a distractor is
false. This is weaker than the verifier chain used on Oops!, making the text-only control important
here: a text-only model scores $0.155$ and $0.147$ on the two arms against chance $0.200$, so the
options are not separable without the video. The length screen matches answer and distractor lengths
to within about one character on average. Answer position is assigned from a balanced vector seeded
on the split, giving $\chi^2 = 0.00$ and placing the answer on the same letter in both arms.
The funnel, per split: $37{,}421$ raw train caption spans reduce to $974$
items, capped to $600$, under filters for video length, boilerplate captions, a $12$\,s span cap and the
requirement of room before the evidence and a placeable region; the span cap ($21{,}786$ spans) and the
$60$\,s video-length floor ($7{,}426$) are the two largest losses. The held-out split runs
$17{,}505 \rightarrow 444$, capped to $250$. Videos are
disjoint across splits, and a video contributes at most $8$ items.

\paragraph{Oops!}
One fixed question, \texttt{What happens in this video?}, chosen neutral rather than ``what goes
wrong'' so the question does not presuppose the failure. The answer is \emph{not} the annotator's
sentence verbatim: it is a rewrite of the annotator's free-text
description of what went wrong, produced by the same model that writes the distractors, required to
entail it in both directions, and required to be verified against the evidence window. This verification drops
$73$ items outright and switches which of the two annotators' sentences is used for $13.9\%$ of the
survivors. The ground truth is therefore model-mediated, which is a real qualification on this pool
and is why we do not treat it as a human-authored benchmark. The four distractors are generated by
\textbf{InternVL3.5-8B} from the last clean pre-onset window, using exactly the frames the model under
test sees at the hardest negative rung, together with the clip's stated goal. Generation has three passes: the
ordinary continuation, a near-miss of that continuation, and mishaps. Every candidate then passes five screens, all
run by the same model: a text check rejecting any candidate that could also describe the same clip; a
vision check over $16$ frames of the whole clip rejecting any candidate that actually happens; a
pre-evidence check rejecting any candidate the visible scene makes impossible, which removes $1{,}550$
options; a lexical
containment cut at $0.6$ against options already accepted; and a blind five-option screen that drops
the item entirely if the generator recovers the answer with no video under both of two option
shuffles. The frozen item set retains $312$ of $621$ attempted items.
\textbf{InternVL3.5-8B scores no row of this benchmark}: the models evaluated on it are the three
Qwen3-VL sizes, GLM-4.1V-9B-Thinking, LLaVA-OneVision-1.5-8B, Idefics3-8B-Llama3 and Pixtral-12B, none
of which had any part in writing or screening the options. Because one question string serves every
item, the question-swap control on this pool is an \emph{option} swap, and is reported as such.

\paragraph{HERBench shot identification.}
The released ordering task names four scenes per question; we re-pose it as identification. The four
shot descriptions are recovered by parsing them out of the released question string (all $2{,}144$
parse) and re-paired with their own timestamps, one shot becomes the target and three become
distractors, all from the same trailer, each at least $2.5$\,s from the target and from each other,
with the target between $0.8$ and $2.0$\,s long. The question presents the four descriptions in a
per-item shuffled order and asks which of them appears in the frames delivered so far; the options are
the literal strings \texttt{Shot 1} to \texttt{Shot 4}, so no option carries content and the model
must read the descriptions. The deferral option, where used, reads \texttt{None of these shots has
appeared yet}. One property is worth stating because it would otherwise be a confound: a distractor
shot could in principle have been delivered by the burst sampler, making two options true at that
checkpoint. The construction prevents this: the sampler guards all four candidate windows with a
$\pm 2.5$\,s exclusion zone and places every non-target burst clear of it. We verified this on the
frozen manifest by recovering each row's delivered timestamps from its frame files: no distractor shot
is delivered on any of the $7{,}648$ rows, and no row has more than one satisfiable option. The guard
is paid for at build time, in items that cannot be placed at all and are skipped rather than shipped
with an ambiguous label.

\paragraph{HD-EPIC matched \present/\absent.}
One fixed question, \texttt{Which action does the person perform in this clip?}, over five options
that are all canonical class surfaces: for each (verb class, noun class) pair we take the wording
annotators used most often for that pair, so the correct option's surface has the same provenance as
the distractors' and cannot be identified by how it is phrased. The distractors are one sharing the
target's verb class, one sharing its noun class, and the
rest drawn from the $800$ most frequent class pairs while excluding both of the target's classes; any
candidate whose class pair appears on \emph{any} narration overlapping the $16$\,s window is rejected,
so exactly one option can be true, and this is established from the annotation rather than by a model
call. Class-level matching means a synonymous narration cannot silently make a second option true.
The action's position inside the \present\ window is drawn uniformly over the placements that keep it
whole, rather than centred. The \absent\ window is anchored on a real action of a different class
and jittered exactly as the \present\ window is, matching the placement rule across labels.
Negatives are stratified into $1{,}824$ where the target's object is off screen, $508$ hard ones where
the object is visible and the queried action is not happening, and $68$ where both classes occur but
never together; the hard fraction runs $0.203$--$0.224$ across the three video-disjoint splits. Items
are selected by scanning a shuffled pool of actions between $1$ and $6$\,s long and stopping at
$2{,}400$ pairs. \textbf{No correctness filter runs}: nothing is dropped for being answered wrongly
when the evidence is present, which is what leaves the present-but-wrong cell populated and makes the
availability-versus-correctness test of \autoref{sec:notconf} possible.

\subsection{Footage independence}
\label{app:data-footage}

The paper spans \emph{eight benchmark families}. Families are the correct unit, not eight
independent datasets. Footage is shared in three places.

The grounded and the evidence-backed multiple-choice evaluations both run on NExT-QA video: all $348$
videos in the evidence-backed pool resolve to the same local NExT-QA files ($269$ validation, $79$
test), and the two pools share $221$ videos and $240$ raw item identifiers; every prefix,
crossed-quad, gating and containment-rebuild pool built from either inherits that overlap. These two
evaluations therefore replicate under \emph{independent annotations of shared footage}, not under
independent video: two annotation efforts, working from different task definitions, marked evidence
on overlapping videos. That is a real but weaker form of replication and is labelled as such wherever
it is used, including in Appendix~\ref{app:generality}, where holding out a \emph{benchmark} and
holding out a \emph{footage family} give different numbers and the second is the one reported.

HD-EPIC footage is shared by the controlled composites, the matched \present/\absent\ windows, and the
causal and head-level analyses built on splits of the matched-pair pool that fitting never touched
(Appendix~\ref{app:causal}); movie-trailer footage is shared by the two HERBench evaluations, which
differ in task, annotation granularity and evidence structure but not in video. The remaining families
are the generation companion's sources, ActivityNet-Captions, Oops!\ and OVO-Bench. These are independent
of the shared-footage families and of each other.

\subsection{Models}
\label{app:data-models}

\autoref{app-tab:models} lists every model evaluated. Throughout, \emph{model family} means the
\textbf{language backbone}, not the vision tower: the probe reads language-model states, so two systems
with different vision towers over the same backbone are not two families for our purpose.

\begin{table}[H]
\centering
\scriptsize
\setlength{\tabcolsep}{3pt}
\begin{tabular}{@{}
  >{\raggedright\arraybackslash}p{2.0cm}
  >{\raggedright\arraybackslash}p{2.8cm}
  >{\raggedright\arraybackslash}p{1.15cm}
  >{\raggedright\arraybackslash}p{1.6cm}
  >{\raggedright\arraybackslash}p{0.55cm}
  >{\raggedright\arraybackslash}p{2.0cm}
  >{\raggedright\arraybackslash}p{2.0cm}
@{}}
\toprule
model & HuggingFace id & vision tower & language backbone & dec.\ lay. & how frames are fed & probed layers \\
\midrule
Qwen3-VL-8B-Instruct & \texttt{Qwen/\allowbreak Qwen3-\allowbreak VL-\allowbreak 8B-\allowbreak Instruct} & Qwen-ViT & Qwen3-8B & 36 & native video path, video metadata & 0, 14, 18, 23, 29, 32 \\
Qwen3-VL-2B-Instruct & \texttt{Qwen/\allowbreak Qwen3-\allowbreak VL-\allowbreak 2B-\allowbreak Instruct} & Qwen-ViT & Qwen3-2B & -- & native video path & 0, 11, 14, 18, 22, 25 \\
Qwen3-VL-32B-Instruct & \texttt{Qwen/\allowbreak Qwen3-\allowbreak VL-\allowbreak 32B-\allowbreak Instruct} & Qwen-ViT & Qwen3-32B & -- & native video path & 0, 26, 32, 42, 51, 58 \\
InternVL3.5-8B \citep{wang2025internvl35} & \texttt{OpenGVLab/\allowbreak InternVL3\_5-\allowbreak 8B-\allowbreak HF} & InternViT & Qwen3-8B & 36 & multi-image, one $448$\,px tile per frame & 0, 14, 18, 23, 29, 32 \\
LLaVA-OneVision-7B \citep{li2024llavaonevision} & \texttt{llava-hf/\allowbreak llava-\allowbreak onevision-\allowbreak qwen2-\allowbreak 7b-\allowbreak ov-\allowbreak hf} & SigLIP \citep{zhai2023siglip} & Qwen2-7B & 28 & native video path, no metadata & by depth fraction \\
LLaVA-OneVision-1.5-8B \citep{an2025llavaonevision15} & \texttt{lmms-lab/\allowbreak LLaVA-\allowbreak OneVision-\allowbreak 1.5-\allowbreak 8B-\allowbreak Instruct} & -- & Qwen3-8B & -- & -- & 0, 14, 18, 23, 29, 32 \\
GLM-4.1V-9B-Thinking \citep{hong2025glm41v} & \texttt{zai-org/\allowbreak GLM-\allowbreak 4.1V-\allowbreak 9B-\allowbreak Thinking} & -- & GLM-4-9B & -- & native video path & 0, 16, 20, 26, 32, 36 \\
Idefics3-8B-Llama3 \citep{laurencon2024idefics3} & \texttt{HuggingFaceM4/\allowbreak Idefics3-\allowbreak 8B-\allowbreak Llama3} & -- & Llama-3.1-8B & -- & -- & 0, 13, 16, 21, 26, 29 \\
Pixtral-12B \citep{agrawal2024pixtral} & \texttt{mistral-community/\allowbreak pixtral-\allowbreak 12b} & -- & Mistral-Nemo-12B & -- & image-native; no video path & 0, 16, 20, 26, 32, 36 \\
\bottomrule
\end{tabular}
\caption{\textbf{Every model evaluated.} Layers are probed at fractions of each model's own depth
rather than at transcribed indices, because decoder depth differs across architectures; the indices
shown are the resolved ones where they are recorded, and a dash marks a value not recorded.
Qwen3-VL-8B-Instruct \citep{qwen2025qwen3vl} is the default everywhere unless a result names another.}
\label{app-tab:models}
\end{table}

Frame packing is fixed by the architecture: InternVL must be driven as multi-image, because dynamic
tiling at $32$ frames overruns the context window, and Pixtral has no video path at all. Every model
nonetheless receives \emph{the same sampled frames from the same manifest}; only the processor packing
differs, and each architecture is verified per run for non-empty visual token identifiers and for
coverage of the vision, option and final positions.

Two of the systems above carry a Qwen3-8B backbone whose text-side geometry is identical to
Qwen3-VL-8B's, and the consequence reaches backwards through the paper: the matched-pair transfer arm
and the crossed-quad evaluation vary the \emph{vision tower} on one Qwen3 language backbone, and must be
described that way rather than as separate model families. The three-model comparison on the grounded
prefixes is not of that kind, its third model being LLaVA-OneVision-7B on a Qwen2 backbone. The model-axis evaluation on the frozen short-window manifest is the one that varies the
backbone: four language-backbone families (Qwen3 at 2B, 8B and 32B; GLM-4; Llama-3.1; Mistral-Nemo)
over a $2$B to $32$B span on a single manifest. GLM-4.1V, Idefics3 and Pixtral are the first
genuinely independent language backbones in the paper. LLaVA-OneVision-7B, on the NExT-QA pools, adds
the fifth family the paper reports, Qwen2, which is also the backbone both scored rival triggers run
on (Appendix~\ref{app:rivals}).

Three details complete the table. No checkpoint revision is pinned in configuration: models are
loaded by identifier, so exact reproduction requires pinning the revisions we report resolved on our
hosts. All models are loaded in \texttt{bfloat16} except LLaVA-OneVision-7B, whose checkpoint ships
\texttt{float16} natively and is loaded as such; nothing is quantised. No attention implementation is
selected, so each architecture's library default applies. Probed layer indices follow from the depth
fractions in Appendix~\ref{app:probe-extract}: for a $28$-layer decoder they are
$\{0,11,14,18,22,25\}$, for $32$ layers $\{0,13,16,21,26,29\}$, for $36$ layers $\{0,14,18,23,29,32\}$,
for $40$ layers $\{0,16,20,26,32,36\}$, and for $64$ layers $\{0,26,32,42,51,58\}$.

\subsection{Frame budgets and pixel quality control}
\label{app:data-frames}

\autoref{app-tab:budgets} gives the delivery budget per geometry; what each geometry is for is in
Appendix~\ref{app:geometries}. The property that matters for reading any decoding result is that
\emph{the sampled frame count is constant within every geometry}, so a probe cannot read ``how many
frames'' in place of ``which frames'': across the rows of a pool the tensor handed to the model always
has the same shape and only its contents change.

\begin{table}[H]
\centering
\scriptsize
\setlength{\tabcolsep}{3pt}
\begin{tabular}{@{}
  >{\raggedright\arraybackslash}p{3.3cm}
  >{\raggedright\arraybackslash}p{1.2cm}
  >{\raggedright\arraybackslash}p{1.2cm}
  >{\raggedright\arraybackslash}p{2.0cm}
  >{\raggedright\arraybackslash}p{1.7cm}
  >{\raggedright\arraybackslash}p{2.4cm}
@{}}
\toprule
geometry & frames & rate & span & stride & decode resolution \\
\midrule
controlled composites & 16 & 2\,fps & fixed 8\,s & -- & 448\,px long side \\
growing prefixes & 32 & -- & $[0,t]$ & evidence-relative ends & 448\,px long side \\
containment windows & 32 & 2\,fps & 16\,s & free 2\,s & -- \\
short windows & 8 & 4\,fps & 2\,s & 0.5\,s & -- \\
sparse bursts & 16 & 2--3\,fps & 3--6 bursts of 1--2\,s & one block & -- \\
\bottomrule
\end{tabular}
\caption{\textbf{Frame budgets per geometry.} Frame count is constant across every row of a geometry.
Prefixes sample uniformly across $[0,t]$, except on the segment-aligned pools, where sampling
guarantees at least two frames per revealed evidence segment. A dash marks a value not recorded.}
\label{app-tab:budgets}
\end{table}

A model-independent filter reads raw frame luminance and drops rows whose delivered frames are flat or
black, before any model runs and without reference to any outcome; on one pool it removed the
\emph{exact same} rows for all three models compared there, which independently confirms its
model-independence. Drop counts are in \autoref{app-tab:pixqc}.

\begin{table}[H]
\centering
\scriptsize
\setlength{\tabcolsep}{3pt}
\begin{tabular}{@{}
  >{\raggedright\arraybackslash}p{5.0cm}
  >{\raggedright\arraybackslash}p{3.2cm}
  >{\raggedright\arraybackslash}p{2.0cm}
  >{\raggedright\arraybackslash}p{1.6cm}
@{}}
\toprule
pool & dropped & of & rate \\
\midrule
NExT-GQA prefixes & 115 prefixes (21 items entirely) & 5,695 & 2.0\% \\
ST-Evidence-MCQ prefixes & 210 prefixes & 6,756 & 3.1\% \\
ST-Evidence-Gen prefixes & 5 prefixes & 4,346 & 0.1\% \\
matched \present/\absent\ windows & 0 & 4,800 & 0\% \\
model-axis short windows & 0 & -- & 0\% \\
Oops!\ short windows & -- & -- & 0.3\% \\
shot-ordering prefixes & not run & -- & -- \\
shot-identification bursts & 1,439 prefixes & 7,648 & 18.8\% \\
\bottomrule
\end{tabular}
\caption{\textbf{Pixel quality-control drop counts.} The filter is model-independent and runs before
any outcome exists. Only the last row has a rate large enough to change what a pool can support.}
\label{app-tab:pixqc}
\end{table}

\paragraph{Where the filter had to be handled asymmetrically.}
On the trailer-based shot-identification ladder the drop rate is $18.8\%$, and it is not uniform along
the ladder: the keep rate is $0.762$ in the first block and $0.499$ in the last, roughly $0.9$ in
between, because trailers open on a studio logo over black and close on titles. On a ladder with
exactly one ready checkpoint per item, this has structural rather than cosmetic consequences: it costs $137$ of
$956$ items their only reachable stop, and only $247$ items keep all eight rows. A filter that
preferentially removes late rungs could by itself make an early-stopping policy look better, so the
probing pool and the policy ladder are handled differently here: the probe is fitted and scored on the
pixel-cleaned rows, where a flat or black window would otherwise contribute a meaningless activation,
while the answer-timing ladder is scored on the \emph{intact} ladder with only the behavioural
eligibility criterion applied, so no rung is removed from among the policy's choices. The filter
therefore cannot manufacture the timing effect, because it never touched the rows the policy chose
among. Elsewhere the rates are small enough that the distinction does not arise and the cleaned pool is
used throughout. Where its count differs from the raw count in \autoref{app-tab:pools}, the reported
cleaned pool has $979$ items and $5{,}580$ rows for the grounded prefixes and $6{,}546$ rows over $338$ videos for the
evidence-backed prefixes.

\subsection{Prompts}
\label{app:data-prompts}

Every evaluation uses a multiple-choice prompt with a fixed option count for the pool it runs on, so
chance is constant within a pool and the answer is always a single option token. Two forms exist. The
\emph{five-option} form presents the benchmark's substantive options only; the \emph{six-option} form
appends one further option, the explicit deferral ``not enough evidence yet'', which the model may
select instead of committing. Which form is used is a configuration setting rather than a per-item
decision, and the same rows can be run under both: the grounded prefix pool scores $0.729$ under the
five-option form and $0.695$ under the six-option form.

The deferral option is what makes the behavioural comparison in \autoref{sec:notconf} possible. With it
in the option set, the model's own decision about whether to commit becomes an observable event with a
score attached, so the readiness readout can be compared against what the model does overtly rather
than only against its confidence in whichever answer it was forced to give. It also supplies the
commitment quantity used by the activation interventions of \autoref{sec:causal}, defined as the gap
between the substantive options and the deferral option, which does not exist without it.

Option counts are not uniform across benchmarks, because some ship their own: the grounded and
evidence-backed multiple-choice pools, shot ordering, and the constructed ActivityNet and Oops!\
items are five-option; shot identification is four-option; and the generation companion's items are
$84\%$ two-option and $16\%$ four-option, giving chance rates of $0.200$, $0.250$ and $0.460$. This
is why every pooled number is also reported per benchmark.

Every prompt is built by one dataset-agnostic function and wrapped as a \emph{single user turn}
containing the video followed by the text; no system message is used anywhere. The text is the
question, the lettered options, and one fixed closing instruction, after which the literal string
\texttt{Answer:} is appended so the answer is read from the next token position:

\begin{quote}\ttfamily\footnotesize\obeylines
why is the baby smiling

A) excited about the food
B) playing with the dog
C) lady tickles him
D) playing with adult
E) happy

Answer with the letter of the correct option.
\end{quote}

The six-option form is byte-identical except that the deferral option is appended last, as
\texttt{F)}. Its wording is set per evaluation: \texttt{Not enough evidence yet} on the prefix
benchmarks, \texttt{Not enough visual evidence has appeared yet.} on the matched-pair substrate, and
\texttt{None of these shots has appeared yet} on shot identification. No benchmark overrides the
closing instruction or adds a question prefix. The option count varies: five on the prefix
benchmarks, four on shot identification, and two on the external benchmark. The external
benchmark also carries over its own published prompt text verbatim as the question
string. Architecture changes only how frames are packed: three models take the native video path and
four receive the identical frames as separate images, leaving the option text untouched.

\subsection{Licensing, data availability, and human subjects}
\label{app:licences}

\paragraph{Human subjects.}
Every readiness label used to fit or evaluate a probe is derived programmatically from evidence
timestamps that already ship with the benchmarks we use; no annotation collected for this work enters
any label, model, or fitted object. Human judgment enters at two points, both as checks. The first is
the authors' own comparison of an automated shot-transition detector against $28$ sampled
transitions, which audits a tool rather than collecting data from participants. The second is the
rater study of Appendix~\ref{app:raters}: three people at our institution, unaffiliated with this
work, judged $900$ screens of publicly released benchmark footage, volunteering their time without
compensation and with the purpose of the task explained to them beforehand. We collected no personal
or demographic data from them, we release no participant-level data, and the raters are not
identifiable in anything we release. No crowdsourcing platform was used. Because no personal data was
collected, we did not seek institutional review.

\paragraph{Licences.}
All benchmarks are used under the terms of their public research releases, and we redistribute no
video frames. Terms differ materially across them: several are non-commercial or share-alike, one
composite source carries a restrictive custom licence prohibiting raw-video redistribution, and one
benchmark is built from third-party compilation footage whose underlying rights sit with the original
uploaders. We therefore point to each benchmark's own repository for authoritative terms rather than
asserting a single licence string. The model checkpoints are permissively licensed: Apache~2.0 for the
Qwen3-VL family, InternVL3.5, both LLaVA-OneVision checkpoints and Pixtral, MIT for GLM-4.1V, and
Apache~2.0 for Idefics3 subject to the additional obligations of its Llama-3 base. Of the two released
streaming triggers we score, one is MIT-licensed and the other non-commercial share-alike.

\section{Extended results: signal, confidence, and correctness}
\label{app:signal}

This appendix gives the per-evaluation detail behind \autoref{sec:signal} and \autoref{sec:notconf}:
every decodability cell with the comparisons measured beside it, the identical-pixel evidence at full
precision, the confidence and correctness dissociations evaluation by evaluation, the uncertainty
baselines one row at a time, and the graded ladder.

\subsection{Decodability, evaluation by evaluation}
\label{app:signal-decode}

\begin{table}[H]
\centering
\scriptsize
\setlength{\tabcolsep}{3pt}
\begin{tabular}{@{}lllccccc@{}}
\toprule
pool & geometry & label & rows / clusters & probe AUROC & shuffle & L$0$@vision & clock \\
\midrule
controlled composites & controlled composite & cue presence & 1,040 / 17 obj. & $0.985$ $[.965,.999]$ & $0.524$ & $0.500$ & -- \\
matched windows & matched pair & availability & 4,800 / 142 & $0.827$ $[.812,.840]$ & $0.495$ & $0.495$ & $0.506$ \\
\quad the same, InternVL3.5-8B & matched pair & availability & 4,800 / 142 & $0.815$ $[.802,.829]$ & $0.504$ & $0.510$ & $0.506$ \\
NExT-GQA prefixes & growing prefix & onset & 5,580 / 458 & $0.733$ $[.715,.749]$ & $0.502$ & $0.544$ & $0.633$ \\
ST-Evidence-MCQ prefixes & growing prefix & onset & 6,546 / 338 & $0.722$ $[.706,.739]$ & $0.484$ & $0.552$ & $0.628$ \\
ST-Evidence-Gen prefixes & growing prefix & $\mathrm{seen}/K$ & 4,346 / 1,529 & $0.827$ $[.813,.840]$ & $0.578$ & $0.651$ & $0.722$ \\
NExT-GQA crossed quads & crossed quad & containment & 4,808 / 277 & $0.727$ $[.710,.745]$ & $0.498$ & $0.500$ & $0.500$ \\
\quad the same, InternVL3.5-8B & crossed quad & containment & 4,808 / 277 & $0.700$ $[.684,.716]$ & $0.500$ & $0.500$ & $0.500$ \\
ST-Evidence-MCQ quads & crossed quad & containment & 1,244 / 88 & $0.707$ $[.670,.747]$ & $0.499$ & $0.500$ & $0.500$ \\
ActivityNet windows & containment window & containment & 2,390 / 447 & $0.764$ $[.742,.787]$ & $0.493$ & $0.517$ & $0.505$ \\
Oops!\ short windows & short window & containment & 1,204 / 301 & $0.874$ $[.855,.892]$ & $0.496$ & $0.654$ & $0.527$ \\
shot-ordering prefixes & growing prefix & availability & 12,582 / 87 & $0.980$ $[.974,.984]$ & $0.503$ & -- & $\mathbf{0.956}$ \\
\bottomrule
\end{tabular}
\caption{\textbf{Every decodability cell with its own comparisons.} All AUROCs are nested,
cluster-held-out, and read at the final prompt token. \textbf{The column is not a ranking:} the rows
differ in geometry (\autoref{app-tab:geometries}), in label strictness (\autoref{app-tab:labels}) and
in where the not-ready rows are drawn from (Appendix~\ref{app:geom-neg}), so each cell is
interpretable only against the comparisons printed beside it. The shuffle is the cluster-grouped label
shuffle \emph{measured} on that pool, not assumed to be $0.500$; on the graded prefixes it is
$0.578$, a base-rate artefact, so that row reads ``$0.827$ against ${\sim}0.58$''. The clock is a
fitted elapsed-time baseline on the same rows: it is informative by construction on a growing prefix,
exactly $0.500$ on a balanced crossed pool, and \textbf{$0.956$ on the shot-ordering prefixes}, where
evidence onset is strongly time-correlated and the readiness reading is therefore not creditable on
its own. Layer $0$ at the final language position is exactly $0.500$ in every row by construction; the
visual layer-$0$ baseline, which \emph{can} carry real signal, is the adjacent column. It is forced to $0.500$ on the
crossed quads by the design and is quoted on the controlled composites under a pair-preserving grouping
(Appendix~\ref{app:probe-quad}); the dash on the shot-ordering prefixes marks a cell that is not
measurable, its activation cache not having been retained, rather than one assumed to be $0.500$, and
the composited stimuli carry no elapsed-time column at all. Row counts identify the reported
decoding pools; pixel quality control is documented in Appendix~\ref{app:data-frames}, and grouping
in \autoref{app-tab:protocols}. Oops!\ is quoted in the both-sided negative
regime; the before-only regime reads $0.919$ but carries a clock at $0.950$.}
\label{app-tab:decode}
\end{table}

The controlled-composite cell is uniform across the cue library rather than carried by a few easy
objects: under held-out-object cross-validation $16$ of the $17$ objects read above $0.98$.

\subsection{Question-conditioning, at full precision}
\label{app:signal-question}

\paragraph{Identical pixels.}
On the controlled composites, the two members of a pair are the same $16$ frames asked two different
questions. Across $520$ identical-prefix pairs the readout reverses on $0.9808$ of them, with a mean
signed margin of $0.9296$ and a crossed DiD of $1.859$ in score units; against an equally
salient but irrelevant cue the relevant question is ranked first on $0.9981$ of $520$ matched pairs,
with mean readiness $0.9587$ when the asked-about cue is present against $0.0119$ when only the
distractor is.

On natural footage the same manipulation is run as a $2\times2$. Each crossed quad pairs two
disjoint-evidence questions with two windows, one containing each evidence hull, and every pair is
verified byte-identical at build time. On NExT-GQA, over the $1{,}196$ of $1{,}208$ quads left intact by pixel QC, from $277$ videos, the
score flips with the question on $0.6614$ $[0.6411,0.6805]$ of swaps ($1{,}590$ of $2{,}404$ pairs
positive) and the DiD is $\mathbf{0.4288}$ $[0.3783,0.4819]$, positive on $945$
of $1{,}196$ quads ($p = 5.2\times10^{-95}$, sign test). The same rows move the model's own behaviour
in the predicted direction: accuracy is $0.7787$ when the window contains evidence for the asked
question against $0.6385$ for the other question, a difference of $+0.1402$ $[0.1159,0.1634]$ from one
forward pass each on identical frames, while output confidence moves only $+0.0446$. On
ST-Evidence-MCQ, whose evidence annotations are independent of NExT-GQA's, the flip rate is $0.7074$
$[0.6685,0.7500]$, the DiD is $\mathbf{0.4874}$ $[0.3931,0.5935]$ over $311$ quads, and
accuracy moves $+0.1302$ $[0.0819,0.1824]$. A second model on row-identical inputs flips at nearly the
same rate ($0.6535$) with a DiD of $0.4099$, but its per-pair margins correlate only
$r = 0.22$ and $0.20$ with the first model's: the two models agree on how often the question decides
the readout, not on which pairs it decides.

\paragraph{Why the crossed design is stronger than a single swap.}
A one-sided swap compares one question against another and can be moved by any fixed difference
between the two prompts, including their length, register, and the probability mass they leave on the
answer letters. The $2\times2$ subtracts that fixed difference twice over and asks only whether it
\emph{reverses} with the evidence-containing window, so the DiD measures the alignment between
which evidence is visible and which question is asked. It also makes the visual controls a design
guarantee rather than an empirical baseline: both labels receive the same multiset of windows, so elapsed
time, motion, brightness, segment count, and the pooled visual representation are forced to exactly
$0.500$ rather than merely measured near it. This is why the crossed quads are built in a fixed-span
window geometry: the four cells of one quad cannot all exist on a growing prefix.

\paragraph{The donor-distance ladder.}
How far a question has to move before the readout follows is itself measurable. We use one frozen
short-window manifest whose corpus carries a single question string for every row, so only the donor
manipulates the question. The same $624$ evidence-carrying pairs are re-scored on
byte-identical frames under three donors at increasing distance: another item's \emph{identical}
question, a \emph{paraphrase} of it drawn from another corpus, and a \emph{distinct}, specific
question drawn from a pool with one question per item.

\begin{table}[H]
\centering
\scriptsize
\setlength{\tabcolsep}{3pt}
\begin{tabular}{@{}llcccc@{}}
\toprule
& & \multicolumn{4}{c}{paired drop on byte-identical frames, $624$ pairs} \\
\cmidrule(l){3-6}
model & language backbone & identical question & paraphrase & distinct question & \quad register-matched \\
\midrule
Qwen3-VL-2B      & Qwen3        & $+0.196$ & $+0.448$ & $\mathbf{+0.594}$ & $+0.477$ \\
Qwen3-VL-8B      & Qwen3        & $+0.080$ & $+0.153$ & $\mathbf{+0.778}$ & $+0.732$ \\
Qwen3-VL-32B     & Qwen3        & $+0.300$ & $+0.579$ & $\mathbf{+0.418}$ & $+0.680$ \\
LLaVA-OV-1.5-8B  & Qwen3        & $+0.305$ & $+0.435$ & $\mathbf{+0.708}$ & $+0.431$ \\
GLM-4.1V-9B      & GLM-4        & $+0.003$ & $+0.140$ & $\mathbf{+0.269}$ & $+0.458$ \\
Idefics3-8B      & Llama-3.1    & $-0.026$ & $-0.087$ & $\mathbf{+0.383}$ & $+0.408$ \\
Pixtral-12B      & Mistral-Nemo & $-0.086$ & $-0.072$ & $\mathbf{+0.325}$ & $+0.325$ \\
\bottomrule
\end{tabular}
\caption{\textbf{Replacing the question collapses the readout in every model tested.} Each cell is
the mean paired drop in readiness score between a row and the same row re-asked with a donor
question, on the same $624$ evidence-carrying pairs and byte-identical frames. Distance increases left
to right: another item's identical question, a paraphrase from a different corpus, and a specific
question from a pool carrying one question per item, with a register-matched build of that last donor
in the final column. Under a genuinely distinct question every model drops by $+0.27$ to $+0.78$, on
$64$--$98\%$ of pairs. Register is not what drives it: Pixtral's raw and register-matched drops agree
to four decimals ($0.3252$, $0.3253$) and Idefics3's register-matched drop is the larger of its two,
so the two builds should be read as a range rather than a point.}
\label{app-tab:donor}
\end{table}

The ranking statistic moves with the level. Scoring the swapped rows against the source rows' labels
gives $0.641$--$0.732$ under the paraphrase donor and $0.555$--$0.645$ under the distinct one; nothing
reaches $0.500$, which is expected, because the frames still carry their own evidence and the probe
still reads it. Two consequences travel with any question-conditioning number. First, whether the
substrate's question varies per item is part of the measurement: on a corpus with one question string
for every row, a donor swap manipulates the options and leaves the question generic, and a generic
``what happens here'' is answerable from any frames. Second, the unit that carries the claim is the
ladder rather than any single donor. An independent route agrees: on a pool whose question genuinely
varies per item, identical-prefix pairwise accuracy against a $0.500$ floor is $0.8024$, $0.7244$ and
$0.6732$ for three of these models, with layer $0$ at exactly $0.500$ in all eight cells and every
shuffle within $0.031$ of chance.

\subsection{Confidence and correctness}
\label{app:signal-conf}

\paragraph{Conditioning on confidence.}
Within fixed confidence bins the readout stays informative: AUROC runs $0.688$--$0.757$ over six bins on
the matched windows, $0.615$--$0.807$ on the graded prefixes, $0.60$--$0.69$ on the NExT-GQA prefixes
and $0.91$--$0.98$ on the shot-ordering prefixes. Residualizing the score on confidence changes
nothing anywhere (\autoref{app-tab:conf}), and the score adds large incremental AUROC over a
confidence-only baseline on every pool.

\begin{table}[H]
\centering
\scriptsize
\setlength{\tabcolsep}{3pt}
\begin{tabular}{@{}lccccc@{}}
\toprule
pool & readiness AUROC & residualized on conf. & drop & confidence alone & increment \\
\midrule
controlled composites, factorial & $0.9317$ & $0.9317$ & $\phantom{-}0.0000$ & $0.534$ & $+0.397$ \\
controlled composites, crossed & $0.9920$ & $0.9920$ & $\phantom{-}0.0000$ & $0.845$ & $+0.147$ \\
matched windows & $0.7701$ & $0.7702$ & $-0.0001$ & $0.636$ & $+0.135$ \\
\quad the same, InternVL3.5-8B & $0.7506$ & $0.7506$ & $\phantom{-}0.0000$ & $0.567$ & $+0.184$ \\
NExT-GQA prefixes & $0.6793$ & $0.6792$ & $+0.0001$ & $0.573$ & $+0.106$ \\
ST-Evidence-MCQ prefixes & $0.6761$ & $0.6762$ & $-0.0001$ & $0.589$ & $+0.088$ \\
ST-Evidence-Gen prefixes & $0.8019$ & $0.8018$ & $+0.0001$ & $0.622$ & $+0.180$ \\
ActivityNet windows, containment & $0.6856$ & $0.6858$ & $-0.0002$ & $0.590$ & $+0.096$ \\
shot-ordering prefixes & $0.9747$ & $0.9746$ & $+0.0001$ & $0.763$ & $+0.212$ \\
\bottomrule
\end{tabular}
\caption{\textbf{Confidence explains none of the readout.} Each row residualizes the readiness score
on the model's output confidence over the same examples and re-scores; the drop is raw minus
residualized, and the largest movement in either direction is $0.0002$. These cells use a pooled
estimator, and the ActivityNet row is the containment-label build, so the levels are not comparable
with \autoref{app-tab:decode} and only the drop and the increment are read here. The increment is the
AUROC gain of the activation score over a confidence-only baseline on the same rows.}
\label{app-tab:conf}
\end{table}

\paragraph{Conditioning on correctness.}
Availability and correctness enter the readout differently. On the $4{,}800$ matched windows the four
cell means are $0.2667$ for \absent/wrong ($1{,}553$ rows), $0.4062$ for \absent/correct ($847$),
$0.5780$ for \present/wrong ($681$) and $0.7362$ for \present/correct ($1{,}719$): the availability
contrast is roughly twice the correctness contrast within either level. Availability decodes at
$0.7224$ among wrong answers and $0.7534$ among correct ones, with model accuracy at $53.5\%$ and the
same-example clock at $0.506$. The controlled composites give the same shape on $480$ factorial rows:
cell means are $0.0793$, $0.0587$, $0.6593$, $0.8644$, with availability AUROC $0.8611$ among wrong and
$0.9256$ among correct, and $0.9028$ with the cells balanced. Thus, the dissociation is not an artefact of
one label. The two factors are not orthogonal: within a fixed availability level the score still
separates correct from wrong answers ($0.737$ inside ready, $0.726$ inside not-ready). Availability
sets the level; correctness modulates it.

The complementary case is a pool on which the model is near chance at the task, so that the readout
cannot be riding on correctness at all. On the shot-ordering prefixes accuracy is $22.7\%$ against a
five-way chance level of $20.0\%$, and availability still decodes at $\mathbf{0.9695}$ among $2{,}169$
present-but-wrong rows and $0.9473$ among correct ones, with cell means $0.0579$, $0.1867$, $0.8342$
and $0.9190$. This evaluation contributes the within-wrong comparison and nothing else: its evidence
onset is strongly time-correlated, and the elapsed-time baseline on the same rows is $0.956$ against
readiness $0.980$, so its absolute level is not a readiness result. Taken together the competent and
the near-chance regimes close both ways the $2\times2$ could fail: a pool where correctness is too
rare to populate the off-diagonal, and a pool where correctness and availability coincide.

\paragraph{A note on cosines.}
None of these tests is geometric, deliberately. In $4096$ dimensions two unrelated directions have
$\mathrm{SD}(\cos) \approx 1/\sqrt{4096} \approx 0.016$, so a cosine near zero between a readiness
direction and a confidence direction is exactly what chance predicts and is not evidence that the two
are independent. Every claim of separation in this paper is therefore prediction-level, supported by
within-bin decoding, residualization, incremental validity, and downstream selection. The same
caution applies wherever directions from two different systems are compared
(Appendix~\ref{app:rivals}).

\subsection{Uncertainty estimators, one row at a time}
\label{app:signal-baselines}

\autoref{app-tab:baselines} expands \autoref{tab:uncertainty} and \autoref{fig:baselines}: every
alternative signal, what it is, its selection accuracy at an identical deadline, and the probe's
margin over it with a video-cluster bootstrap interval. The two substrates are the Oops!\ evidence
windows and the same-action-class ActivityNet-Captions condition, the two settings in which readiness
selection has its largest effect; the probe is fitted on these substrates and scored out of fold,
which is why the comparison is read as ``these signals do not substitute for it here'' rather than as
a system-level win.

\begin{table}[H]
\centering
\scriptsize
\setlength{\tabcolsep}{3pt}
\begin{tabular}{@{}l>{\raggedright\arraybackslash}p{3.6cm}cccc@{}}
\toprule
& & \multicolumn{2}{c}{Oops!\ windows} & \multicolumn{2}{c}{ActivityNet} \\
\cmidrule(lr){3-4}\cmidrule(l){5-6}
signal & what it is & acc. & probe $-$ this & acc. & probe $-$ this \\
\midrule
\texttt{sc\_agree} & share of $K$ sampled answers, each with a rationale, agreeing with the mode & $0.417$ & $+0.260$ $[.205,.314]$ & $0.393$ & $+0.253$ $[.208,.301]$ \\
\texttt{sc\_neg\_entropy} & negative entropy of that sampled answer distribution & $0.417$ & $+0.260$ $[.205,.314]$ & $0.393$ & $+0.253$ $[.208,.301]$ \\
\texttt{se\_neg} & semantic entropy: $K$ open-ended answers clustered by bidirectional entailment & $0.478$ & $+0.199$ $[.144,.253]$ & $0.413$ & $+0.233$ $[.191,.276]$ \\
\texttt{se\_neg\_lennorm} & the same, length-normalized & $0.487$ & $+0.189$ $[.135,.244]$ & $0.427$ & $+0.220$ $[.177,.264]$ \\
\texttt{se\_neg\_loose} & the same, one-direction clustering & $0.446$ & $+0.231$ $[.176,.288]$ & $0.423$ & $+0.223$ $[.180,.267]$ \\
\texttt{naive\_neg\_entropy} & clustering-free predictive entropy over sequence log-probabilities & $0.458$ & $+0.218$ $[.160,.276]$ & $0.413$ & $+0.233$ $[.191,.277]$ \\
\texttt{ask\_pyes} & $P(\text{yes})$ from asking the model directly whether it has enough evidence yet & $0.590$ & $+0.087$ $[.045,.131]$ & $0.408$ & $+0.238$ $[.195,.283]$ \\
\texttt{order\_agree} & answer agreement across cyclic option rotations & $0.494$ & $+0.183$ $[.128,.240]$ & $0.417$ & $+0.230$ $[.184,.277]$ \\
\texttt{order\_mean\_maxp} & mean top-option probability across the same rotations & $0.596$ & $+0.080$ $[.035,.125]$ & $0.565$ & $+0.082$ $[.046,.117]$ \\
\texttt{sig\_max} & SigLIP retrieval, max frame--option similarity in the window & $0.474$ & $+0.202$ $[.147,.256]$ & $0.505$ & $+0.142$ $[.103,.182]$ \\
\texttt{sig\_margin} & the same, top-$1$ minus top-$2$ option margin & $0.468$ & $+0.208$ $[.154,.263]$ & $0.478$ & $+0.168$ $[.125,.211]$ \\
\texttt{confidence} & the model's own answer probability & $0.606$ & $+0.071$ $[.029,.112]$ & $0.540$ & $+0.107$ $[.068,.145]$ \\
\texttt{ens\_vote} & pooling what the model already says across the ladder: vote & $0.481$ & $+0.196$ $[.144,.250]$ & $0.435$ & $+0.212$ $[.171,.253]$ \\
\texttt{ens\_mean\_prob} & \dots\ mean probability & $0.503$ & $+0.173$ $[.125,.224]$ & $0.445$ & $+0.202$ $[.161,.241]$ \\
\texttt{ens\_conf\_prob} & \dots\ confidence-weighted probability & $0.519$ & $+0.157$ $[.109,.205]$ & $0.455$ & $+0.192$ $[.152,.233]$ \\
\midrule
\texttt{composite} & supervised head over all baseline columns, out of fold by video & $0.631$ & $+0.045$ $[.003,.087]$ & $0.565$ & $+0.082$ $[.046,.117]$ \\
\texttt{composite+probe} & the same head with the probe's column added & $0.673$ & $+0.003$ $[-.016,.022]$ & $0.648$ & $-0.002$ $[-.018,.015]$ \\
\bottomrule
\end{tabular}
\caption{\textbf{Every alternative signal, and the probe's margin over it.} Selection accuracy is
measured at an identical deadline, so the comparison is latency-matched by construction; the probe
reads $0.676$ and $0.647$ on these substrates and answering only at the end reads $0.487$ and
$0.358$. Intervals are video-cluster bootstraps of the paired difference. \emph{Every} alternative
loses on both substrates; adding the supervised head back to the probe moves selection by
$+0.003$ and $-0.002$, both intervals spanning zero. Signal names match \autoref{fig:baselines}.}
\label{app-tab:baselines}
\end{table}

The table groups the signals by what they read: two sampled-answer agreement columns, four
entropy-based columns, one direct self-assessment, two option-order columns, two retrieval columns,
the model's own answer probability, and three no-forward-pass ensembles over the window ladder, with
the supervised head and the head-plus-probe row beneath them.
The eleven estimators are adaptations and variants from five families: self-consistency
\citep{wang2023selfconsistency} over $8$ sampled answers
(temperature $1.0$, top-$p$ $0.95$); semantic and predictive entropy over $10$ samples (temperature
$1.0$, top-$p$ $0.9$), clustered by bidirectional entailment under a DeBERTa MNLI model
\citep{he2021deberta,kuhn2023semanticuncertainty,farquhar2024semantic}; direct self-assessment,
adapting probability-based self-evaluation \citep{kadavath2022language} to the probability of answering
``yes'' when asked whether enough evidence is available; option-order robustness, motivated by
documented option-position bias \citep{zheng2024selectors}; and SigLIP frame--text retrieval
\citep{zhai2023siglip}. The table also carries the
model's own confidence, three window-ladder ensembles, and two supervised
combinations, which are reference columns rather than members of the eleven.

\paragraph{Tie-breaking.}
The probe keeps the published convention of resolving an \texttt{argmax} tie in favour of the earliest
window, while each alternative is additionally scored under \emph{its own} most favourable tie-break.
This matters: sampled-answer agreement takes about $5.5$ distinct values across a median of $19$
windows, so it ties constantly and moves from $0.417$ to $0.583$ on the Oops!\ substrate on the
convention alone, and option-order agreement moves from $0.494$ to $0.609$. The ordering survives the
correction designed to break it: every alternative re-scored this way stays positive on both
substrates, with a smallest surviving margin of $+0.067$ on Oops!\ and $+0.082$ on ActivityNet.

\paragraph{Why competent uncertainty estimators still rank evidence badly.}
The estimators are not broken; they are answering a different question. Output confidence,
sampled-answer agreement and option-order consistency all read the model's \emph{own correctness} well
above chance on both substrates ($0.572$--$0.687$) while reading evidence availability worse
($0.547$--$0.655$) and selecting the wrong window. Being a good uncertainty detector does not make a
signal a readiness detector. The alternative that asks the model directly about readiness
is also the only one that reads availability ($0.712$) above correctness
($0.573$), which is what it was designed to do, and it is the strongest single baseline on Oops!; it
still loses selection by $+0.087$, and on ActivityNet it falls to chance on both axes ($0.503$ and
$0.479$), so a direct question is substrate-dependent rather than a substitute.

Semantic entropy \citep{farquhar2024semantic} sits at chance on both axes on both substrates
($0.495$/$0.494$ against availability, $0.482$/$0.473$ against correctness), and the reason is
specific to this application rather than to the method. The sampled answers are coherent and
video-grounded and parse at rate $1.0$ from $10$ samples per row, but two descriptions that differ
only in an incidental detail do not mutually entail, so strict bidirectional entailment returns a mode
of $9$--$10$ clusters from those $10$ samples (mean $7.1$) and the estimator sits near its ceiling with
nothing left to discriminate. A one-direction clustering variant and a clustering-free predictive
entropy agree with it, so the null is not an artefact of the clustering criterion. Semantic entropy
was built for short factoid answers; on descriptive video answers its estimator saturates. That is a
limit of applying it here, and we report it as one.

\subsection{Graded evidence}
\label{app:signal-graded}

On the graded prefixes, where a question's evidence is split into $K = 1$--$11$ segments, readiness
reads $0.827$ $[0.813,0.840]$ against a \emph{measured} within-video shuffle of $0.578$. The shuffle reflects a base-rate
artefact of the pool, not the $0.500$ a balanced design would give. The elapsed-time baseline
reads $0.722$ and confidence $0.622$. Score rises with the graded target $\mathrm{seen}/K$ after
regressing both on fractional video progress: the partial Spearman correlation is $\mathbf{0.3775}$
$[0.3501,0.4019]$, against a confound correlation of $0.293$ between progress and segments seen. The
axis has to be the fraction: with the raw segment count in its place, the same control gives
$-0.0092$ $[-0.0575,+0.0393]$, a false null produced by pooling ``one of one'' with ``one of five''
(Appendix~\ref{app:geom-graded}).

\begin{table}[H]
\centering
\scriptsize
\setlength{\tabcolsep}{3pt}
\begin{tabular}{@{}ccccc@{}}
\toprule
$K$ & rows & ready & not ready & AUROC \\
\midrule
$1$ & 2,309 & 1,877 & \textbf{432} & $0.852$ \\
$2$ & 1,121 & 667 & \textbf{454} & $0.787$ \\
$3$ & 379 & \textbf{174} & 205 & $0.716$ \\
$4$ & 211 & \textbf{81} & 130 & $0.699$ \\
$5$ & 156 & \textbf{51} & 105 & $0.729$ \\
$6$ & 75 & \textbf{24} & 51 & $0.555$ \\
\bottomrule
\end{tabular}
\caption{\textbf{Per-$K$ strata, each with its minority-class count.} Read on the pixel
quality-controlled build of the graded prefixes, which reads $0.831$ $[0.817,0.844]$ overall. The
minority class is in bold and falls from $432$ rows at $K=1$ to $24$ at $K=6$; the $K=6$ cell rests on
$24$ positives and is printed for completeness rather than read as a decline. Strata above $K=6$ are
not tabulated for the same reason. Within each $K$ the mean score rises monotonically with the number
of segments seen for $K = 1$--$5$, which is $96\%$ of the rows, with within-$K$ Spearman correlations
of $0.489$, $0.521$, $0.505$, $0.599$ and $0.614$.}
\label{app-tab:perk}
\end{table}

The graded structure is question-conditioned rather than a property of the footage. Re-asking the same
$4{,}346$ prefixes, using the same JPEGs on disk but an unrelated video's question, collapses
within-arm AUROC from $0.832$ to $0.546$ and the mean score from $0.595$ to $0.169$, a mean paired drop
of $0.4268$ $[0.4123,0.4423]$ with $85.5\%$ of pairs falling. Progress tracking goes with it:
the correlation between score and $\mathrm{seen}/K$ falls from $0.554$ to $0.064$, and between score
and fractional video progress from $0.435$ to $0.085$. A generic accumulator of visual events would
have reproduced both curves unchanged.

The ladder itself is monotone over the bulk of the mass: mean probe score rises from $0.150$ at
$\mathrm{seen}/K = 0$ ($130$ rows) through $0.291$ at one third ($107$), $0.431$ at one half ($395$)
and $0.477$ at two thirds ($115$), to $0.688$ when all segments have been seen ($1{,}014$). The
intermediate fractions carrying only a handful of rows are not read individually.

\section{Generality across models and benchmarks}
\label{app:generality}

Section~\ref{sec:general} reports the endpoints of each generality axis; this appendix gives the cells
behind them.

\subsection{The full model axis, all four negative-sampling regimes}
\label{app:gen-modelaxis}

The seven models read one frozen manifest of Oops! footage \citep{epstein2020oops}: $8$ frames over a
$2$\,s window, with $6{,}475$ ladder rows over $312$ video clusters. Frames and option
order are byte-identical across models. A mask then selects which \emph{not-ready} rows the probe is fitted against, leaving a matched
subset of $728$--$1{,}248$ rows over $182$--$312$ clusters:

\begin{itemize}\itemsep1pt
\item \texttt{before}: not-ready rows only from before the evidence; a fitted elapsed-time clock on
      those same rows reads $0.9497$.
\item \texttt{near}: a same-scene content control, $1$--$2$\,s before the evidence onset, so only the
      evidence is missing. Clock $0.8663$.
\item \texttt{after}: not-ready rows only from after the evidence has passed. Clock $0.9069$.
\item \texttt{both}: rows from both sides. This is the regime Section~\ref{sec:general}
      reports, and the one that breaks the clock: the same fitted readout falls to
      $0.5271$.
\end{itemize}

\autoref{app-tab:modelaxisfull} gives all $28$ cells. Two spans matter, and neither is interpretable
without its regime named: under the reported \textbf{both-sided} regime, readiness runs
$0.7329$--$0.9046$, and under the \textbf{before-only} regime it runs $0.7896$--$0.9411$. The before-only
figures are the larger ones and we do not lead with them, for a reason worth stating plainly: before-only
is the regime in which elapsed time is nearly sufficient on its own. A fitted clock scores $0.9497$
there, above \emph{every} readiness cell in that regime, whose best is $0.9411$. A strong
reading there therefore says little about readiness. The both-sided regime is the least favourable of the four in
six of seven models and holds the lowest cell in the grid ($0.7329$); in the seventh, Idefics3-8B, the
after-only cell is lower still at $0.7406$. No cell, under any regime, falls below $0.7329$. The
clock-tracking column divides each model's readiness drop from \texttt{before} to \texttt{both} by the
clock's drop over the same contrast ($0.4226$): a readout reading elapsed time would land near $1$, and
the measured range is $0.0864$--$0.2286$. The \texttt{near} cell is excluded from that ratio by
construction, being a content control rather than a clock cell.

{\scriptsize
\setlength{\tabcolsep}{3pt}
\begin{longtable}{llccccccc}
\caption{\textbf{The full model axis: seven models $\times$ four negative-sampling regimes.} Readiness
AUROC is nested cross-validated with video-disjoint folds; the interval is a video-cluster bootstrap. The
final-language-position layer-$0$ column is $0.500$ in all $28$ cells and is forced to be: at layer $0$ a
fixed text position holds the same embedded token on every row, so the readout is degenerate before any
frame is consulted. The visual layer-$0$ column \emph{can} carry real signal because it reads raw
patch embeddings. It spans $0.5790$--$0.8194$, and the language-position readout clears it in every
cell, by at least $0.0740$. The within-video shuffle stays inside $0.4669$--$0.5206$ throughout. The
fitted elapsed-time clock is a property of the regime rather than the model: $0.9497$ (\texttt{before}),
$0.8663$ (\texttt{near}), $0.9069$ (\texttt{after}), $0.5271$ (\texttt{both}). Clock tracking is one
number per model, on the \texttt{before}$\rightarrow$\texttt{both} contrast.}
\label{app-tab:modelaxisfull} \\
\toprule
backbone & model & regime & AUROC & $95\%$ CI & L$0$@final & L$0$@vision & shuffle & clock track \\
\midrule
\endfirsthead
\toprule
backbone & model & regime & AUROC & $95\%$ CI & L$0$@final & L$0$@vision & shuffle & clock track \\
\midrule
\endhead
\bottomrule
\endfoot
Qwen3 & Qwen3-VL-2B & \texttt{before} & $0.8934$ & $[0.8727,\,0.913]$ & $0.500$ & $0.8194$ & $0.4920$ & $0.1039$ \\
 &  & \texttt{near} & $0.8930$ & $[0.8722,\,0.9126]$ & $0.500$ & $0.8085$ & $0.5078$ &  \\
 &  & \texttt{after} & $0.8648$ & $[0.8364,\,0.8923]$ & $0.500$ & $0.6974$ & $0.4986$ &  \\
 &  & \texttt{both} & $0.8495$ & $[0.8277,\,0.8705]$ & $0.500$ & $0.6828$ & $0.5021$ &  \\
\addlinespace
Qwen3 & Qwen3-VL-8B & \texttt{before} & $0.9190$ & $[0.9022,\,0.9356]$ & $0.500$ & $0.7848$ & $0.4925$ & $0.1065$ \\
 &  & \texttt{near} & $0.9106$ & $[0.8925,\,0.9275]$ & $0.500$ & $0.7834$ & $0.5006$ &  \\
 &  & \texttt{after} & $0.8764$ & $[0.8503,\,0.9022]$ & $0.500$ & $0.6714$ & $0.4930$ &  \\
 &  & \texttt{both} & $0.8740$ & $[0.855,\,0.8922]$ & $0.500$ & $0.6535$ & $0.4962$ &  \\
\addlinespace
Qwen3 & Qwen3-VL-32B & \texttt{before} & $0.9411$ & $[0.9258,\,0.9549]$ & $0.500$ & $0.7373$ & $0.5078$ & $0.0864$ \\
 &  & \texttt{near} & $0.9277$ & $[0.9102,\,0.9442]$ & $0.500$ & $0.7436$ & $0.5206$ &  \\
 &  & \texttt{after} & $0.9067$ & $[0.8832,\,0.9295]$ & $0.500$ & $0.6595$ & $0.4705$ &  \\
 &  & \texttt{both} & $0.9046$ & $[0.888,\,0.9211]$ & $0.500$ & $0.6324$ & $0.4866$ &  \\
\addlinespace
Qwen3 & LLaVA-OV-1.5-8B & \texttt{before} & $0.8927$ & $[0.8735,\,0.9112]$ & $0.500$ & $0.7734$ & $0.5041$ & $0.1574$ \\
 &  & \texttt{near} & $0.8877$ & $[0.8673,\,0.907]$ & $0.500$ & $0.7669$ & $0.5008$ &  \\
 &  & \texttt{after} & $0.8312$ & $[0.7999,\,0.8624]$ & $0.500$ & $0.6286$ & $0.5115$ &  \\
 &  & \texttt{both} & $0.8262$ & $[0.8042,\,0.8486]$ & $0.500$ & $0.6446$ & $0.5037$ &  \\
\addlinespace
GLM-4 & GLM-4.1V-9B & \texttt{before} & $0.8306$ & $[0.8057,\,0.8525]$ & $0.500$ & $0.7071$ & $0.4971$ & $0.1649$ \\
 &  & \texttt{near} & $0.8149$ & $[0.7902,\,0.8375]$ & $0.500$ & $0.7186$ & $0.5088$ &  \\
 &  & \texttt{after} & $0.7779$ & $[0.7467,\,0.8101]$ & $0.500$ & $0.6470$ & $0.4916$ &  \\
 &  & \texttt{both} & $0.7609$ & $[0.737,\,0.784]$ & $0.500$ & $0.6153$ & $0.4985$ &  \\
\addlinespace
Llama-3.1 & Idefics3-8B & \texttt{before} & $0.8669$ & $[0.8425,\,0.8898]$ & $0.500$ & $0.7823$ & $0.4913$ & $0.2286$ \\
 &  & \texttt{near} & $0.8553$ & $[0.8321,\,0.8775]$ & $0.500$ & $0.7775$ & $0.5092$ &  \\
 &  & \texttt{after} & $0.7406$ & $[0.7088,\,0.7748]$ & $0.500$ & $0.6534$ & $0.5138$ &  \\
 &  & \texttt{both} & $0.7703$ & $[0.7471,\,0.7953]$ & $0.500$ & $0.6274$ & $0.4669$ &  \\
\addlinespace
Mistral-Nemo & Pixtral-12B & \texttt{before} & $0.7896$ & $[0.7651,\,0.8137]$ & $0.500$ & $0.6865$ & $0.4935$ & $0.1342$ \\
 &  & \texttt{near} & $0.7840$ & $[0.7585,\,0.8098]$ & $0.500$ & $0.6862$ & $0.4885$ &  \\
 &  & \texttt{after} & $0.7373$ & $[0.7053,\,0.7712]$ & $0.500$ & $0.5813$ & $0.4934$ &  \\
 &  & \texttt{both} & $0.7329$ & $[0.7106,\,0.757]$ & $0.500$ & $0.5790$ & $0.5062$ &  \\
\end{longtable}
}

\subsection{Where in depth the signal appears}
\label{app:gen-layers}

\autoref{fig:depth} shows the seven-model depth profiles on the clock-balanced Oops!\ evaluation.
Two crossed-question replications on natural footage also show increasing decodability with depth.
Their depth fractions are normalized to the deepest sampled layer, not to the whole decoder
(Appendix~\ref{app:probe-extract}). At the shallowest sampled site ($0.44$--$0.45$ on this scale),
readiness AUROC is near chance ($0.5488$ and $0.5271$), rising to $0.6597$ and $0.6477$ at the
deepest sampled site.

The input controls have different roles (Appendix~\ref{app:probe-controls}). The final language
token is constant at layer $0$, so its $0.500$ is a protocol check. The visual layer-$0$ readout is
a measured baseline: it reaches $0.5790$--$0.8194$ across the grid, and the language-position
readout exceeds it in all $28$ cells.

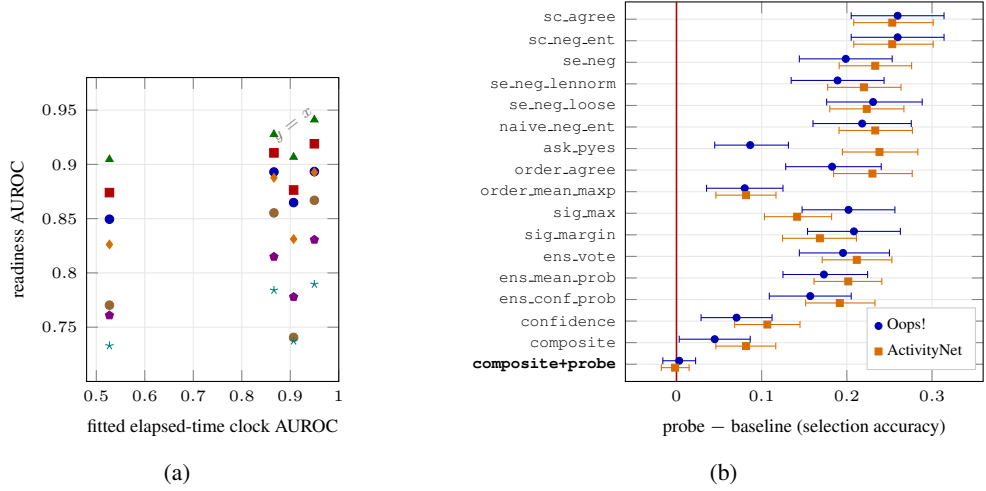
\begin{figure}[t]
\centering
\begin{subfigure}{0.40\linewidth}
\centering
\begin{tikzpicture}
\begin{axis}[
  width=0.88\linewidth, height=5.6cm,
  xlabel={fitted elapsed-time clock AUROC},
  ylabel={readiness AUROC},
  xmin=0.48, xmax=1.0, ymin=0.70, ymax=0.98,
  xtick={0.5,0.6,0.7,0.8,0.9,1.0},
  ytick={0.75,0.8,0.85,0.9,0.95},
  grid=major, grid style={gray!18},
  tick label style={font=\tiny}, label style={font=\scriptsize},
  mark size=1.6pt, only marks,
]
\addplot[gray, dashed, line width=0.5pt, mark=none, forget plot,
         domain=0.70:0.98, samples=2]{x};
\addplot[blue!70!black,   mark=*]         table[col sep=comma,x=clock,y=auc]{figures/data/clock_Qwen3_VL_2B.csv};
\addplot[red!70!black,    mark=square*]   table[col sep=comma,x=clock,y=auc]{figures/data/clock_Qwen3_VL_8B.csv};
\addplot[green!45!black,  mark=triangle*] table[col sep=comma,x=clock,y=auc]{figures/data/clock_Qwen3_VL_32B.csv};
\addplot[orange!85!black, mark=diamond*]  table[col sep=comma,x=clock,y=auc]{figures/data/clock_LLaVA_OV_1_5_8B.csv};
\addplot[violet,          mark=pentagon*] table[col sep=comma,x=clock,y=auc]{figures/data/clock_GLM_4_1V_9B.csv};
\addplot[brown!80!black,  mark=otimes*]   table[col sep=comma,x=clock,y=auc]{figures/data/clock_Idefics3_8B.csv};
\addplot[teal,            mark=star]      table[col sep=comma,x=clock,y=auc]{figures/data/clock_Pixtral_12B.csv};
\node[font=\tiny, gray, rotate=38, anchor=south] at (axis cs:0.925,0.925) {$y=x$};
\end{axis}
\end{tikzpicture}
\caption{}
\label{fig:clock}
\end{subfigure}
\hfill
\begin{subfigure}{0.565\linewidth}
\centering
\begin{tikzpicture}
\begin{axis}[
  width=0.80\linewidth, height=6.6cm,
  xlabel={probe $-$ baseline (selection accuracy)},
  xmin=-0.06, xmax=0.36,
  ytick={0,1,2,3,4,5,6,7,8,9,10,11,12,13,14,15,16},
  yticklabels={\texttt{sc\_agree},\texttt{sc\_neg\_ent},\texttt{se\_neg},
    \texttt{se\_neg\_lennorm},\texttt{se\_neg\_loose},\texttt{naive\_neg\_ent},
    \texttt{ask\_pyes},\texttt{order\_agree},\texttt{order\_mean\_maxp},
    \texttt{sig\_max},\texttt{sig\_margin},\texttt{ens\_vote},\texttt{ens\_mean\_prob},
    \texttt{ens\_conf\_prob},\texttt{confidence},\texttt{composite},
    \textbf{\texttt{composite+probe}}},
  ymin=-0.8, ymax=16.8, y dir=reverse,
  grid=major, grid style={gray!18},
  tick label style={font=\tiny}, label style={font=\scriptsize},
  legend style={font=\tiny, at={(0.98,0.03)}, anchor=south east,
                draw=gray!40, fill=white},
  legend cell align=left,
]
\draw[red!70!black, line width=0.6pt] (axis cs:0,-0.8) -- (axis cs:0,16.8);
\addplot[only marks, mark=*, mark size=1.3pt, blue!70!black,
  error bars/.cd, x dir=both, x explicit]
  table[col sep=comma, x=pm, y expr=\thisrow{idx}-0.16,
        x error plus expr=\thisrow{hi}-\thisrow{pm},
        x error minus expr=\thisrow{pm}-\thisrow{lo}]{figures/data/pm_oops.csv};
\addlegendentry{Oops!}
\addplot[only marks, mark=square*, mark size=1.3pt, orange!85!black,
  error bars/.cd, x dir=both, x explicit]
  table[col sep=comma, x=pm, y expr=\thisrow{idx}+0.16,
        x error plus expr=\thisrow{hi}-\thisrow{pm},
        x error minus expr=\thisrow{pm}-\thisrow{lo}]{figures/data/pm_anet.csv};
      \addlegendentry{ActivityNet}
\end{axis}
\end{tikzpicture}
\caption{}
\label{fig:baselines}
\end{subfigure}
\caption{\textbf{Readiness is neither a stopwatch nor an uncertainty estimate.}
\textbf{(a)} Readiness AUROC against the AUROC of a fitted elapsed-time clock, one point per
model~$\times$~negative regime. Colours identify Qwen3-VL-2B (blue), Qwen3-VL-8B (red),
Qwen3-VL-32B (green), LLaVA-OV-1.5-8B (orange), GLM-4.1V-9B (violet), Idefics3-8B (brown)
and Pixtral-12B (teal). Before-only sampling makes
the clock ($0.950$) stronger than every probe. With both-sided sampling the clock falls to $0.527$,
while all seven probes remain informative: their AUROCs change much less than the clock's.
\textbf{(b)} Eleven modern uncertainty estimators, plus a supervised head fitted over all
of them (\texttt{composite}), compared on fixed-deadline answer selection. Points are Readiness
Gating's margin over each baseline with a cluster-bootstrap interval, using earliest-window
tie-breaking; Appendix~\ref{app:signal-baselines} reports the favourable-tie-break comparison.
Every estimator loses on both substrates. The last row compares the combined head
(\texttt{composite+probe}) with Readiness Gating alone: adding the baseline scores yields no
detectable selection gain, with both intervals spanning zero.}
\label{fig:clockbase}
\end{figure}

\autoref{fig:clock} displays the negative-sampling comparison: the clock falls from $0.9497$ to
$0.5271$ between before-only and both-sided sampling, while each probe changes much less. The
before-only clock is stronger than every probe; the evidence for readiness is its persistence when
that shortcut weakens. \autoref{fig:baselines} shows the complementary answer-selection comparison
with uncertainty estimators, detailed in Appendix~\ref{app:signal-baselines}.

\subsection{Holding out a benchmark family}
\label{app:gen-lobo}

Every cell above is a probe fitted on the pool it is scored on, which supports \emph{the phenomenon
recurs} rather than \emph{one direction underlies it}. For the stronger statement we fit a single probe
on the union of all-but-one benchmark family, weighting each family equally. Layer and regularization are chosen
by leave-one-source-family-out cross-validation inside the source union only. The probe is then applied frozen to
the held-out family. Source unions run $4{,}450$--$9{,}639$ rows over $889$--$1{,}131$ videos; selected
sites are late ($0.72$--$1.0$ of depth) at $C = 0.01$ throughout. Two hold-out definitions are possible
and they disagree. Holding out a \emph{benchmark} leaves NExT-GQA \citep{xiao2023trust} and
ST-Evidence-MCQ in each other's training unions. Because they share $221$ NExT-QA videos, footage the
frozen probe was fitted on reappears in the cell it is scored on. Holding out the whole \emph{footage
group} removes that leak, and it is the honest reading (\autoref{app-tab:lobo}).

\begin{table}[t]
\centering
\scriptsize
\setlength{\tabcolsep}{3pt}
\begin{tabular}{lccccccc}
\toprule
held-out family & transfer AUROC & $95\%$ CI & own AUROC & retention & L$0$@vision & shuffle & clock \\
\midrule
NExT-GQA          & $0.7073$ & $[0.6893,\,0.7261]$ & $0.7464$ & $0.9476$ & $0.5139$ & $0.5006$ & $0.4586$ \\
ST-Evidence-MCQ   & $0.7153$ & $[0.6904,\,0.7425]$ & $0.7448$ & $0.9604$ & $0.5310$ & $0.4976$ & $0.4548$ \\
ST-Evidence-Gen   & $0.7501$ & $[0.7211,\,0.7808]$ & $0.7692$ & $0.9752$ & $0.5389$ & $0.5208$ & $0.5446$ \\
ActivityNet       & $0.6525$ & $[0.6335,\,0.6745]$ & $0.7637$ & $0.8544$ & $0.4974$ & $0.5040$ & $0.4942$ \\
Oops!             & $0.7610$ & $[0.7382,\,0.7843]$ & $0.8740$ & $0.8707$ & $0.5507$ & $0.5022$ & $0.5278$ \\
\midrule
\multicolumn{4}{l}{mean retention, footage-group hold-out} & $\mathbf{0.9217}$ & & & \\
\multicolumn{4}{l}{mean retention, benchmark hold-out} & $0.9491$ & & & \\
\bottomrule
\end{tabular}
\caption{\textbf{One probe, fitted without a family's footage, read on that family.} Rows are the
footage-group hold-out, in which NExT-GQA and ST-Evidence-MCQ are removed together because they share
$221$ videos. Retention is transfer AUROC over the family's own fitted AUROC, the comparable quantity given
that the family ceilings themselves run $0.7448$--$0.8740$. Holding out benchmarks singly raises the mean
to $0.9491$, and the whole of that difference is the two NExT-QA families, whose retention rises by
$0.0575$ and $0.0799$ once their shared footage is allowed back into the training union. Every cell sits
fully inside the source union's delivered frame-density envelope.}
\label{app-tab:lobo}
\end{table}

The baselines hold in every held-out cell and are best read as a block: visual layer-$0$ within
$0.4974$--$0.5507$, within-video shuffle within $0.4976$--$0.5208$, elapsed-time clock within
$0.4548$--$0.5446$. The frozen probe is not recovering the held-out family through frame density, video
identity, or the clock. Nor is the pooling carrying the result, although one family has a different
geometry: Oops! delivers $8$ frames over $2$\,s where the others deliver $32$ over $16$\,s. Rebuilding
the matrix with that family dropped, and comparing like with like against the benchmark-wise hold-out on
the four remaining families, moves each transfer AUROC by at most $0.0190$: $0.7472$, $0.7873$, $0.7311$
and $0.6421$ against $0.7502$, $0.7748$, $0.7501$ and $0.6525$.

\subsection{Reading one model's probe in another model's basis}
\label{app:gen-procrustes}

Going off-diagonal across benchmarks still leaves the models separate. To ask whether these are one
readout in different coordinates, we fit an orthogonal map between two models' activations
\citep{schonemann1966procrustes} on a video-disjoint half of the frozen Oops!
manifest \emph{using no labels}, then read model $S$'s frozen probe in model $T$'s basis on the held-out
half: $1{,}204$ rows over $301$ clusters, the map fitted on $150$ videos and scored on $604$ rows at base
rate $0.5$, over all $7 \times 6 = 42$ ordered pairs. An orthogonal map in high dimension aligns almost
anything, so an aligned value means nothing without two companions quoted \emph{at the same rank}: a
random orthogonal map of that rank, and the un-aligned transfer that skips the map
(\autoref{app-tab:procrustes}).

\begin{table}[t]
\centering
\scriptsize
\setlength{\tabcolsep}{4pt}
\begin{tabular}{lcccc}
\toprule
rank & aligned (median) & random-rotation null & row-shuffled alignment & pairs beating own null \\
\midrule
$4$   & $0.5135$ & $0.4932$ & $0.5109$ & $32/42$ \\
$8$   & $0.6733$ & $0.4822$ & $0.4801$ & $41/42$ \\
$16$  & $0.6938$ & $0.5175$ & $0.5114$ & $42/42$ \\
$32$  & $0.6935$ & $0.4940$ & $0.5237$ & $42/42$ \\
$64$  & $0.6925$ & $0.5060$ & $0.4925$ & $42/42$ \\
$\mathbf{128}$ & $\mathbf{0.7134}$ & $\mathbf{0.4963}$ & $0.5009$ & $\mathbf{42/42}$ \\
$256$ & $0.7181$ & $0.5037$ & $0.5048$ & $42/42$ \\
\bottomrule
\end{tabular}
\caption{\textbf{The cross-model rotation, with both controls read at the same rank.} Medians over the
$42$ ordered model pairs on the frozen Oops! manifest. Rank $128$ is the primary rank; reading an aligned
value against a null taken from a different rank is exactly the error these columns exist to prevent.
In \textbf{un-aligned} transfer, the source probe is applied to the target's activations with no map at
all. It reads $0.4668$ over the $14$ ordered pairs whose two models share $d_{\text{model}}$, i.e.\ at
chance. Recovering the basis therefore needs paired activations on shared inputs, and this is not a
shippable probe.}
\label{app-tab:procrustes}
\end{table}

At rank $128$ the aligned median is $0.7134$, against a same-rank random-rotation null of $0.4963$, a
row-shuffled alignment of $0.5009$, and un-aligned transfer, which skips the map entirely, at
$0.4668$. All $42$ ordered pairs beat
their own null, as they do at every rank from $16$ upward, across four language-backbone families and a
$2$B--$32$B span, and against each target model's own fitted ceiling the aligned readout retains a median
$0.8983$, range $0.8189$--$0.9889$.

\paragraph{What this bounds.} The same label-free map preserves properties that are decodable but are
\emph{not} readiness at least as well as it preserves readiness. Applied to the gold answer letter, the
clock, and the source model's own correctness, those survive the rotation as well or better, and on two
of three substrates a control is preserved better than readiness is: median preservation $0.8030$ for
readiness against $0.8948$ for the source model's correctness on Oops!, and $0.7346$ against $0.8960$ for
the gold answer letter on ST-Evidence-MCQ, with HD-EPIC \citep{perrett2025hdepic} close to a tie
($0.9504$ against $0.9074$). The map carries generic shared structure, so the supported claim is limited:
\emph{the readiness readout is not model-idiosyncratic}. We do not claim that readiness is the direction
these models share.

\subsection{Transfer across the attribute the readiness is about}
\label{app:gen-attribute}

A last axis holds model and footage fixed and varies what the question asks about. On controlled HD-EPIC
composites \citep{perrett2025hdepic}, where a counter-prior cue is placed on real footage so the evidence
for the answer is known exactly, a probe fitted only on \emph{colour} questions, with its
weights, scaler, layer and threshold then frozen, reads readiness on \emph{size} questions at AUROC $0.9380$ ($0.9502$
pairwise on identical prefixes; $0.8756$ with the focal object held out); the reverse direction reads
$0.9954$ ($0.9942$; $0.9874$), against a jointly trained ceiling of $0.9957$ and $0.9889$. On the same
size rows the model's own confidence reads $0.6939$, and elapsed time, cue salience, and visible-cue
count each read exactly $0.5000$.

The geometry is the point: the colour and size readiness directions are \emph{not} aligned, with cosine
$0.2656$, yet each predicts the other close to the joint ceiling. Two independent
attribute-specific detectors would not transfer; two non-aligned readouts of one underlying state would,
which is why our test is cross-task \emph{predictive} transfer rather than vector cosine.

The bound, plainly: a single source, controlled synthetic composites, one model, and a held-out-object
arm that holds out the focal object but not every relational one, since colour and size draw from the
same $17$-object library. It is an existence result under full control, not a claim about natural
footage.

\section{External transfer and the operating point}
\label{app:transfer}

Appendix~\ref{app:generality} varies the model, the footage, and the benchmark while holding the label
definition fixed. This appendix goes one step further out, to an external benchmark with its own task
definition and its own readiness labels, neither of which the probe has seen. The result splits in two,
and the split is the point: the probe's \emph{ranking} survives the move, its \emph{operating point}
does not.

\subsection{Carrying a frozen probe to an external benchmark}
\label{app:tr-zeroshot}

We take the standardizing scaler, the direction, and the read-out layer from a source probe fitted on the
ST-Evidence-Gen containment-window pool. This uses $32$-frame, $16$\,s windows advanced on an evidence-agnostic clock,
with a window counting as ready when it holds at least one released evidence segment ($3{,}059$ rows over
$693$ videos, $68.8\%$ ready; held-out source AUROC $0.7828$). We apply all three unchanged to
OVO-Bench \citep{li2025ovobench}, scoring against OVO's own labels and its own task definition. Nothing
is fitted, selected, or rescaled on the target. The target pool is $4{,}440$ windows over $145$ items and
$42$ videos, of which the $3{,}913$ rows carrying labels are scored, at a positive rate of $0.1306$.

The frozen probe reads AUROC $\mathbf{0.7484}$ $[0.6890, 0.7994]$. A probe fitted directly on the target
with video-grouped nested cross-validation reaches $0.8322$ $[0.7789, 0.8754]$, so the carried-over
direction recovers about $90\%$ of what fitting on the target would buy. The controls on the same rows:
the within-video shuffle is $0.5085$ and the fitted elapsed-time clock is $0.4511$.

The trivial columns deserve a sentence of their own: on this target they sit \emph{below} chance,
with their rankings inverted. OVO's own probe-grid index reads $0.4790$ and fraction-of-video
$0.4796$, the model's own confidence reads $0.3496$, and a probe fitted on a different, wrong notion of
readiness and carried over the same way reads $0.3134$ $[0.2698, 0.3545]$. Where the obvious proxies
point the wrong way, $0.7484$ has to come from somewhere else.

\paragraph{The operating point does not come with it.} The same frozen probe is badly miscalibrated on
the target: expected calibration error $\mathbf{0.8079}$, accuracy $\mathbf{0.1377}$ at the carried-over
threshold against $\mathbf{0.8681}$ at a target-chosen one, on a target whose base rate is $0.1306$. The
mechanism is visible in the scores: the mean carried-over score is $0.9385$, with $99.28\%$ of rows called
ready. This failure is easy to miss in a favourable case, since on a majority-positive target the
same defect reads as $0.674 \rightarrow 0.727$ and looks cosmetic. That is why those four numbers must
travel together. A second external leg behaves the same way: the frozen probe reads the official
Fine-grained Action benchmark at $0.738$ $[0.725, 0.752]$ over $1{,}200$ items and $86$ unseen videos,
$94\%$ of that pool's $0.787$ refit ceiling, at ECE $0.334$, converting none of a demonstrated
$+9.0$-point ground-truth-oracle headroom there.
Two positive rates describe this pool and they count different rows: $0.1948$ over the $601$ rows
carrying the benchmark's own readiness annotation, and $0.1306$ over the $3{,}913$ evidence-agnostic
rows the transfer probe is scored on. The accuracies quoted here are computed on the latter.

\subsection{Repairing the operating point, and what the repair is worth}
\label{app:tr-threshold}

The failure is a threshold, not a direction, so we price it. Holding the direction, the layer, and the
scaler untouched, we refit \emph{only} the decision threshold on $k$ target videos drawn without
replacement, evaluate on the complement, and resample $200$ times. \autoref{app-tab:threshold} gives
$k=8$ on four deployment cells.

\begin{table}[h]
\centering
\scriptsize
\setlength{\tabcolsep}{4pt}
\begin{tabular}{lcccccc}
\toprule
cell & base rate & carried-over & $J$ at $k=8$ & \textbf{Platt at $k=8$} & oracle threshold & constant predictor \\
\midrule
OVO-Bench (window)   & $0.1306$ & $0.1377$ & $0.6955$ & $\mathbf{0.8648}$ & $0.8681$ & $0.8694$ \\
Fine-grained Action  & $0.2733$ & $0.5764$ & $0.6456$ & $\mathbf{0.7277}$ & $0.7311$ & $0.7267$ \\
NExT-GQA ladder      & $0.1968$ & $0.6744$ & $0.6515$ & $\mathbf{0.8006}$ & $0.8062$ & $0.8032$ \\
ActivityNet ladder   & $0.1850$ & $0.4715$ & $0.5982$ & $\mathbf{0.8099}$ & $0.8151$ & $0.8150$ \\
\bottomrule
\end{tabular}
\caption{\textbf{Accuracy under a threshold refitted on eight labelled target videos.} Direction, layer,
and scaler are frozen throughout; only the threshold moves. The two refitting arms differ in objective:
our intercept-only variant of Platt scaling \citep{platt1999probabilistic} moves the crossing point, whereas Youden's
$J$ \citep{youden1950index} maximizes TPR $-$ FPR, which is not an accuracy objective and loses by roughly $0.17$ on the same
eight labels ($0.6955$ $[0.6083, 0.7854]$ against $0.8648$ $[0.8473, 0.8791]$ on the first row). The last
column is the accuracy of always predicting the majority class, and it is the number that bounds how much
this repair can be worth.}
\label{app-tab:threshold}
\end{table}

The gain is entirely a base-rate effect, and it is confined to cells whose deployment base rate differs
from the balanced pool the probe was standardized on. Across the twelve evaluated cells, seven are
base-rate-shifted (rates $0.1306$--$0.2881$); there the mean gap between the carried-over accuracy and an
oracle threshold is $0.2327$, and eight labelled videos close $92.7\%$ of it. The remaining five cells are
the balanced probing pools, at base rate $0.500$ by construction; there the gap is $0.0078$ to begin with,
and the same eight labels make accuracy \emph{worse}. Knowing which regime one is in requires knowing the
target's base rate, which a deployer may not.

\paragraph{What the labels actually buy.} Not thresholded skill. On the base-rate-shifted cells an
\emph{oracle} threshold, which gives the best any threshold could do with the labels free, beats a constant
``not ready'' predictor by at most $+0.0281$, and on the OVO cell by $-0.0013$, meaning the oracle
threshold loses to the constant predictor there. At base rates of $0.13$--$0.29$, thresholded accuracy
simply cannot express what the probe knows. What the eight labels remove is a deployment hazard: a probe
that calls $99.28\%$ of rows ready would commit at every rung of the ladder. What the probe knows is its
ranking, which reads AUROC $0.65$--$0.76$ on these cells and is what a policy can act on. This is why the
paper's decision results in Section~\ref{sec:utility} are selection and stopping statistics rather than
thresholded accuracy, and why we write that the ranking transfers and the operating point does not,
rather than that the probe transfers.

\begin{figure}[t]
\centering
\begin{tikzpicture}
\begin{axis}[
  width=0.62\linewidth, height=5.0cm,
  xlabel={probe score (predicted probability)},
  ylabel={observed positive fraction},
  xmin=0.28, xmax=1.02, ymin=-0.03, ymax=1.03,
  xtick={0.3,0.5,0.7,0.9}, ytick={0,0.25,0.5,0.75,1.0},
  grid=major, grid style={gray!18},
  tick label style={font=\scriptsize}, label style={font=\small},
  legend style={font=\scriptsize, at={(0.02,0.98)}, anchor=north west, draw=none, fill=white, fill opacity=0.85, text opacity=1},
]
\addplot[gray, dashed, domain=0.28:1.02, forget plot]{x};
\addplot[blue!70!black, mark=*, mark size=1.6pt, thick]
  table[col sep=comma,x=mean_pred_prob,y=observed_pos_frac]{figures/data/ovo_reliability.csv};
\addlegendentry{carried-over probe}
\draw[red!70!black, thick, dotted] (axis cs:0.5,-0.03) -- (axis cs:0.5,1.03);
\node[font=\tiny, red!70!black, anchor=west, rotate=90] at (axis cs:0.51,0.30) {carried-over threshold};
\end{axis}
\end{tikzpicture}
\caption{\textbf{The ranking transfers; the operating point does not.} Reliability of the
carried-over probe on the external benchmark over $3{,}913$ scored rows, against the diagonal. The
probe concentrates almost all mass above $0.9$, placing $3{,}093$ of $3{,}913$ rows there despite the pool's
far lower positive rate. At the carried-over threshold of $0.5$, almost every row is therefore called
ready. The curve is nonetheless monotone at the top end, which is why the ordering survives while the
calibration does not (ECE $0.808$).}
\label{app-fig:calibration}
\end{figure}
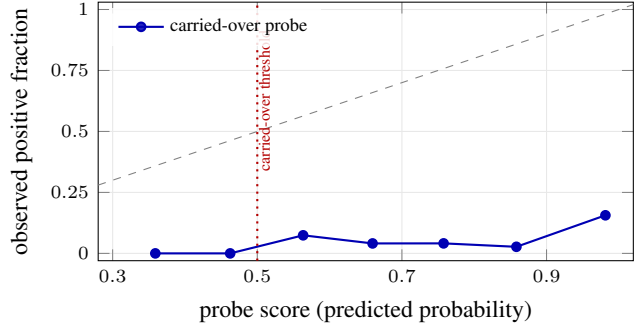

\section{Readiness Gating: answer timing and the attainable-band relationship}
\label{app:utility}

Policy gains depend on how much evidence-timed answering can improve the model's accuracy on a
given task. After defining the evaluation, we test that relationship with a controlled distractor
manipulation and then place it beside the descriptive comparison across configurations.

\subsection{The unit is a stop decision}

The unit of every result in this appendix is an \emph{item's stop decision}, not a row, and the metric is
accuracy at matched latency rather than AUROC. This analysis shares no code with the probing pipeline.

Readiness Gating has two modes, and they do not use the same monitor. \textbf{Gated stopping} maps the
readiness score and the model's confidence each to its train-fold quantile rank and walks the ladder,
stopping at the first rung where the smaller of the two ranks crosses a single training-selected
threshold, then emitting that rung's answer. It is compared against the same construction on confidence
alone at matched latency, and the reported quantity \dstar\ is the accuracy difference at the budget
where latency matches. Keeping one threshold across both arms holds them to the same number of free
parameters, and it makes \dstar\ an incremental-validity statistic: what the readiness readout adds to
the confidence the model already exposes, rather than what it beats. A \textbf{selection} policy instead
returns, at a fixed deadline, the answer cached from the highest-\emph{readiness} window observed so
far, with no confidence term and no threshold. Because every arm then stops on the same rung,
selection is latency-matched \emph{by construction}, with no budget grid and no latency guard needed. It
is the cleaner estimator, and it is the one used for the comparison against uncertainty estimators in
\autoref{sec:notconf}.

The two modes therefore carry two different statistics and they are not interchangeable. \dstar\ is a
stopping gain for the readiness-and-confidence conjunction over confidence alone at equal latency; the
\emph{selection edge} is a fixed-deadline accuracy difference between a readiness-only monitor and a
rival monitor on the same rungs. \autoref{sec:utility} quotes the selection edge for the per-format medians and the count of
configurations that improve, and \dstar\ everywhere else; \autoref{fig:bandlaw}b plots \dstar.

\subsection{Matching latency: rungs or seconds}

\dstar\ can be matched on the rung index or on wall-clock seconds. These are not the same quantity once
rung counts vary across items: $20$ to $33$ on one substrate and $4$ to $83$ on another. Every
\dstar\ the paper reports is matched on seconds. The choice can change the verdict: on the containment rebuild the
rung-matched figure is $+0.0067$ $[-0.0005, +0.0139]$, an interval spanning zero, while the
seconds-matched figure on the same run is $+0.0092$ $[+0.0028, +0.0158]$, which does not. Three of the
$26$ configurations disagree this way, and it is why the variable is named rather than assumed.

A second, subtler point: \textbf{matching the mean is not matching per item}. A confidence-style policy
allocates latency adaptively across items while Readiness Gating allocates it comparatively flat, so at
equal \emph{mean} latency the two are not directly comparable. Readiness Gating can beat confidence
inside every cohort and still lose on the pooled average, which is a genuine Simpson's paradox rather
than a rounding artefact. The two remedies used here are per-item latency standardization and the
selection view, which matches latency by construction.

\subsection{Guards}

Three checks run alongside every timing result and are reported rather than patched: that the two arms'
latencies actually match at the reported budget; that a fixed-time policy which never consults the model
does \emph{not} reproduce the readiness arm, which is what rules out the ladder having inherited an
anchoring leak from how its checkpoints were placed (Appendix~\ref{app:probe-controls}); and that the
count of distinct frames does not track the label. A result whose anchoring check does not pass is not
creditable as a timing result, and is not reported as one.

\subsection{The attainable band}

The attainable band is the accuracy difference between the evidence oracle and answering at the
end of the video:
\[
  \mathrm{band} = \text{evidence-oracle accuracy} - \text{accuracy from answering at the end},
\]
using the end rather than the first checkpoint because a stopping policy can always choose to wait.
This is a reference gap, not a guaranteed upper bound, and can be negative. It depends on the model,
question and answer format, delivered windows, evidence labels and oracle tie-breaking convention.

Evidence timestamps alone do not determine answerability. Removing the annotated span from the
complete video costs only $7.6$, $8.4$ and $5.2$ pp on three benchmarks, so a model can often
answer without those frames. Moving the readiness threshold from onset to containment also changes
which windows the oracle treats as ready (Appendix~\ref{app:geom-contain}). Both properties affect
the reference gap against which policy gain is measured.

\subsection{A controlled change in headroom}

\begin{figure}[H]
\centering
\begin{subfigure}{0.45\linewidth}
\centering
\begin{tikzpicture}
\begin{axis}[
  width=0.98\linewidth, height=4.8cm,
  xlabel={attainable band (pp)}, ylabel={$\Delta^{*}$ (pp)},
  xmin=11, xmax=41, ymin=0, ymax=9,
  xtick={17.5,34.5}, ytick={0,2,4,6,8},
  grid=major, grid style={gray!18},
  tick label style={font=\scriptsize}, label style={font=\small},
]
\addplot[only marks, mark=square*, mark size=2.5pt, red!75!black,
  error bars/.cd, y dir=both, y explicit]
  table[row sep=\\, x=band, y=gain,
        y error plus expr=\thisrow{hi}-\thisrow{gain},
        y error minus expr=\thisrow{gain}-\thisrow{lo}] {
    band gain lo hi\\
    17.5 1.10 0.55 1.67\\
    34.5 4.66 2.84 6.54\\
  };
\draw[red!75!black, -{Stealth[length=4pt]}, line width=0.7pt]
  (axis cs:17.5,1.10) -- (axis cs:34.5,4.66);
\node[font=\scriptsize, align=center, anchor=south] at (axis cs:17.5,2.1)
  {Uniform\\distractors};
\node[font=\scriptsize, align=center, anchor=south] at (axis cs:34.5,6.8)
  {Same-action\\distractors};
\end{axis}
\end{tikzpicture}
\caption{Controlled comparison on identical pixels}
\end{subfigure}
\hfill
\begin{subfigure}{0.52\linewidth}
\centering
\begin{tikzpicture}
\begin{axis}[
  width=0.98\linewidth, height=4.8cm,
  xlabel={attainable band (pp)},
  ylabel={$\Delta^{*}$ (pp)},
  xmin=-9, xmax=102, ymin=-6.5, ymax=25,
  grid=major, grid style={gray!18},
  tick label style={font=\scriptsize}, label style={font=\small},
  yticklabel style={/pgf/number format/fixed, /pgf/number format/precision=2},
]
\addplot[blue!65!black, line width=1pt, domain=0:100, samples=2, forget plot] {0.1482*x - 0.10};
\draw[gray, dashed, line width=0.4pt] (axis cs:-9,0) -- (axis cs:102,0);
\addplot[only marks, mark=*, mark size=1.5pt, black,
  error bars/.cd, y dir=both, y explicit]
  table[col sep=comma,
        x expr=100*\thisrow{band}, y expr=100*\thisrow{delta_star},
        y error plus expr=100*(\thisrow{ci_hi}-\thisrow{delta_star}),
        y error minus expr=100*(\thisrow{delta_star}-\thisrow{ci_lo})]
  {figures/data/bandlaw_fit.csv};
\addplot[only marks, mark=square*, mark size=2.4pt, red!75!black, forget plot]
  coordinates {(34.5,4.66) (17.5,1.10)};
\node[font=\scriptsize, align=left, anchor=north west] at (rel axis cs:0.03,0.98)
  {Pooled: $r^{2}=0.883$\\Without three high-band points: $r^{2}=0.479$};
\end{axis}
\end{tikzpicture}
\caption{Descriptive relationship across 26 configurations}
\end{subfigure}
\caption{\textbf{Changing distractors moves headroom and policy gain together.}
\textbf{(a)} ActivityNet-Captions conditions hold items, videos, windows, labels, frames and answer
positions fixed, changing only the distractors. The attainable band moves from $17.5$ to $34.5$ pp
and \dstar\ from $+1.10$ to $+4.66$ pp; error bars are the reported confidence intervals.
\textbf{(b)} The same two conditions (red) enter the $26$-configuration comparison over ten
substrates, each configuration entering once on its full item pool. The plotted quantity is the
seconds-matched \dstar\ of gated stopping, not the selection edge quoted in \autoref{sec:utility},
which is why some configurations sit below zero. The fitted slope is $+0.148$ pp per band pp. Three high-band configurations carry much of the pooled relationship; the
remaining $23$ give $r^2=0.479$. These are configuration-level observations, not $26$ independent
dataset replications. The band tracks gain but does not determine it.}
\label{fig:bandlaw}
\end{figure}
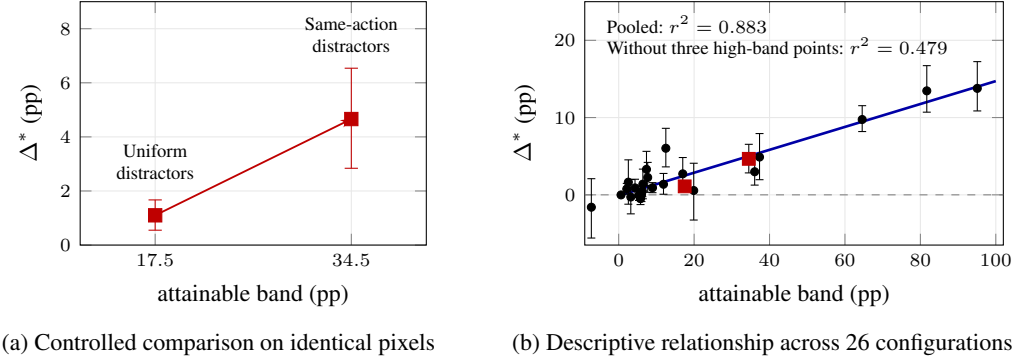

The ActivityNet-Captions comparison changes only the distractor options, keeping items, videos,
windows, frames, evidence labels and answer positions fixed. Same-action-class distractors yield a
$34.5$ pp band and \dstar\ of $+4.66$ pp $[+2.84,+6.54]$; uniform distractors yield a $17.5$ pp
band and $+1.10$ pp $[+0.55,+1.67]$. Gain differs by $4.2\times$ while the delivered visual
evidence is unchanged. This is a controlled change in answer format, not an isolated intervention
on the band itself: distractor difficulty can affect both headroom and the policy's ability to use
it. The two move together at an implied $+0.209$ pp of gain per pp of headroom
(\autoref{fig:bandlaw}a).

\subsection{Decidable items and flat ladders}

A stop-decision metric must never be averaged over all items without reporting how many of them a
stopping policy could possibly have changed. On several pools, three items in four have a
\emph{constant} right-or-wrong ladder: the answer is the same at every rung, so no timing policy can move
them. They dilute the measured effect and make the realized minimum detectable effect look far tighter
than the design's real sensitivity. We therefore report the decidable cohort beside the full set, and
report the flat-item share.

That share is itself informative. Under the controlled difficulty intervention of \autoref{sec:utility},
which holds videos, windows, frames, evidence labels and answer positions fixed and replaces only the
distractor options, the flat-item share moves from $25.7\%$ to $59.6\%$. This same manipulation
moves the attainable band. Where most ladders are flat, there is nothing for a timing policy to convert.

\subsection{Oracles are mostly ties}

A binary evidence oracle is nearly all ties: wherever an item has more than one ready window, every one
of them is an equally good stop, and an \texttt{argmax} convention silently picks the earliest. That
convention systematically handicaps the oracle. Oracle accuracy is therefore reported as a \emph{family}
over tie-breaking conventions rather than as a single number, and any comparison against it names the
convention used.

\subsection{Decision-rule levers}

Two levers are defined on a finished run with nothing refitted. \emph{Memory} emits the answer cached
from the best-scoring window so far, exactly as in the selection policy above. \emph{Per-item running
normalization} standardizes the score stream within an item; two variants exist and are not
interchangeable, one letting warm-up rows score zero and therefore fire, the other marking them as
ineligible so they cannot, and swapping one for the other changes the basis of a reported number.

Memory is significant on exactly those substrates where the thresholded policy already converted, and it
replicates out of sample ($+0.1087$ on a split never fitted on). The honest statement is therefore that a
decision-rule lever \emph{amplifies an existing effect and does not create one}. A lever cannot substitute
for attainable band.

\subsection{The band regression}

\autoref{fig:bandlaw}b places the controlled comparison within $26$ configurations spanning ten
substrates, with \dstar\ matched on seconds throughout. Each configuration enters once on its
full item pool; six cohort-filtered subsets of recent-window configurations are not counted again
beside their full-pool counterparts. Twenty-five of the twenty-six take confidence as the rival
monitor; one growing-prefix configuration takes a trained selector head instead, on both sides of
the contrast. Configurations can share footage, models and geometry, so they
are not independent dataset replications. The descriptive fit is
\[
  \dstar_{\mathrm{pp}} = 0.1482\,\mathrm{band}_{\mathrm{pp}}-0.10, \qquad r = 0.9399, \quad r^2 = 0.883 .
\]
The controlled contrast and the pooled fit agree in sign and scale. The fit summarizes the
observed configurations rather than establishing a universal conversion rate.

\paragraph{Two bounds on how far the line should be pushed.} The first is that the band alone does not
determine gain. The clearest case is a substrate the paper reports: on Oops! the gate converts $0.48$ of a
$12.5$ pp band where the fit predicts $0.15$ of it, the largest positive residual in the set. Substrates
with near-identical headroom can therefore differ several-fold in what a timing policy actually
extracts, so no reading of \autoref{fig:bandlaw} should treat a large band as a promise.

The second is leverage. Three configurations, all of them sparse four-option bursts, have bands past
$0.6$ while the remaining $23$ all sit below $0.38$, and those three carry the least-squares fit:
restricted to the other $23$ the slope is $+0.1139$ at $r^2 = 0.479$. Rank order is the more stable
summary. Spearman's $\rho$ is $0.793$ $[0.550, 0.926]$ over all $26$ and $0.700$ $[0.383, 0.874]$
over the $23$, both intervals from a bootstrap over configurations. What weakens when the three are
removed is the least-squares relationship, not the ordering.
The defensible claim is that the band tracks policy gain across substrates, not that it predicts
any single configuration's \dstar\ to a point.

\section{Comparison with published streaming triggers}
\label{app:rivals}

This appendix gives the detail behind \autoref{sec:rivals}: what each released trigger is, how it was
driven, the validation that establishes it was driven correctly, and the bounds on what the comparison
supports.

\subsection{The triggers, and how they were driven}

Streaming systems that decide when to respond ship that decision as a small learned head, and in every
released case we could obtain, the head is a linear readout of a residual stream
\citep{wang2024mmduet,qian2025dispider,chen2024videollm}.

\begin{table}[H]
\centering
\scriptsize
\setlength{\tabcolsep}{3pt}
\begin{tabular}{llll}
\toprule
system & head & shape & base model \\
\midrule
MMDuet & \texttt{relevance}, \texttt{informative} & $2\times3584$ ea. & LLaVA-OneVision (Qwen2-7B) \\
Dispider & \texttt{silent}                          & $1\times1536$      & Qwen2 \\
VideoLLM-online & streaming-EOS logit & unembedding row & Llama-3-8B \\
\bottomrule
\end{tabular}
\caption{\textbf{The released triggers are linear residual-stream readouts, the same kind of object as
our probe.} MMDuet ships as a $0.13$\,GB adapter and publishes a $2$\,fps rate; Dispider's head is
embedded in its checkpoint's root shards and decides once per $16$\,s, and its sign convention is that a
high score means \emph{respond}. VideoLLM-online was obtained but never run, so no number is asserted for
it anywhere in this paper.}
\label{app-tab:rivalheads}
\end{table}

Each rival is driven inside its own environment and framework pin ($4.41.2$ and $4.44.2$ respectively,
against $5.1.0$ for our own extraction path), so no rival is run under a library version it was not
released against. The two substrates are our own shot-identification and ActivityNet-Captions ladders,
replayed bit-identically; nothing was refitted, and no new footage was introduced for this comparison.
StreamReady and StreamOV had released no code at the time of writing, and StreamBridge releases its
implementation but withholds the activation model's weights for video copyright reasons, so none of
the three could be driven end-to-end here.

\subsection{Validation: the same head on its own benchmark}

A rival reading near chance invites the objection that it was driven incorrectly. We therefore ran the
same head, through the same code, on the benchmark its own paper uses to evaluate it: Charades-STA
temporal grounding \citep{gao2017tall}, $3{,}720$ test examples over $1{,}334$ videos, using their
annotations and their evaluation procedure. It reaches $\mathrm{R@IoU}{=}0.5$ of $53.2$ against the
$42.4$ reported in the original paper \citep{wang2024mmduet}, so the head is being driven at or above its
published level. Scored by \emph{our} metric on \emph{their} task it reads AUROC $0.6998$, against
$0.4837$--$0.5133$ on ours. This is the like-for-like contrast.

One reproducibility fact belongs here neutrally: the released evaluation script cannot read the released
inference script's output, as the two disagree on field names and on whether the relevance score is a
scalar or a pair. The metric was therefore reimplemented from their own code.

\subsection{Zero-shot readings on our labels}

The two substrates of \autoref{tab:rivals} are scored at the following scale. Shot identification
contributes $796$ items over $77$ video clusters and $6{,}368$ windows, at a ready base rate of
$0.125$; the ActivityNet-Captions condition contributes $600$ items over $447$ clusters and $16{,}109$
windows, at a base rate of $0.185$.

On natural ActivityNet-Captions footage, where MMDuet's own $2$\,fps frame grid coincides exactly with our
window grid and alignment staleness is $0.00$\,s, its \texttt{relevance} head reads $0.4837$, below
chance with the interval excluding $0.5$, against $0.753$ for readiness on identical rows. On
shot identification it reads $0.5133$, an interval spanning chance, against $0.9260$ for readiness.
Because that ActivityNet cell runs at the rival's own published rate with no staleness, the
off-distribution objection does not apply to it.

Dispider's head reads $0.4767$ on shot identification. It is reported as a secondary carrying no claim of
its own, for the reason given in Appendix~\ref{app:rivals-bounds}.

On latency-matched answer selection the ordering is the same. Our probe's edge over plain confidence is
$+0.26759$ on shot identification, reproducing the anchor established independently on that substrate,
which is what licenses the rival columns placed beside it; MMDuet's \texttt{relevance} head scores
$-0.21859$ against the same reference, significantly \emph{worse} than confidence.

\subsection{The trained direction, on its own base model}

The substrate difference can be removed entirely. MMDuet's base model is, by value, the same language
model in the same basis as a checkpoint our extraction path can load: of $339$ shared language-model
tensors, $337$ are identical to a maximum absolute difference of $2.98\times10^{-8}$, bf16 round-trip
precision. The two that differ are the token embedding and the output head, and they differ only along
the vocabulary axis ($152{,}064$ against $152{,}128$); the $3{,}584$-dimensional residual axis, which is
where a direction lives, is identical in both. That is what makes a cosine between its head and our probe
a defined quantity rather than an assumption. It licenses the cosine only: it says nothing about the
visual stream entering that basis, which the rival's released adapter changes outright by replacing the
visual-to-language projector. The readings below therefore do not rest on it, being taken on the rival's
own adapted stream.

The vector its \texttt{relevance} head actually consumes is captured by a forward pre-hook on the head's
own input, inside MMDuet's own runtime and framework pin, with its adapter applied, its replaced
visual-to-language projector in place and its query-first prompt order. The basis is therefore the one its head
reads by construction rather than by reimplementation. The pool is $1{,}204$ balanced windows over $301$
Oops!\ videos at a ready base rate of $0.500$, drawn under the clock-balanced \texttt{both} negative
regime of Appendix~\ref{app:geom-neg}, where a fitted elapsed-time clock on these rows reads
$0.5276$, and the readout cell is the layer and frame position its head uses.

At that cell the \texttt{relevance} direction reads $0.6072$ $[0.5791, 0.6339]$ against our labels:
above chance, with the interval excluding $0.5$, and above the $0.5484$ that a layer-$0$ readout at the
same frame position reaches, but only by $+0.059$. A probe fitted on the
identical rows and the identical activations reads $0.8433$ $[0.8220, 0.8639]$, clearing that same floor
by $+0.295$, against a within-video label shuffle of $0.5073$. Its absolute cosine with our probe is $0.0162$,
and all $56$ measured cosines lie inside a random-direction null ($99$th percentile $0.043$ at $d=3584$),
the largest being $0.0414$.
The state is thus linearly present in the trained system's own residual stream, while the direction that
system learned to trigger on is statistically indistinguishable from orthogonal to it.

Scored instead with not-ready rows drawn only from before the evidence, the same three readouts read
$0.6073$, $0.6098$ and $0.9001$ while a fitted clock on those rows reads $0.9497$. Moving to both-sided
sampling drops the clock by $0.422$ and those readouts by $0.000$, $0.061$ and $0.057$, giving
clock-tracking ratios (Appendix~\ref{app:geom-clock}) of $0.000$ for the trained direction and $0.135$
for our probe, both far from the $1$ a stopwatch would score. Most of the before-only floor was the
clock, which is why the trained direction's margin over it changes sign between the two regimes while
our probe's does not ($+0.290$ against $+0.295$). One bound on the contrast: the balanced pool also has
eleven fewer videos and after-side negatives containing the aftermath, so our probe's $0.057$ drop
cannot be attributed to the clock alone.

MMDuet's second head reads our label better than the head it triggers on: \texttt{informative} reaches
$0.6662$ at the same cell, clearing the layer-$0$ floor by $+0.118$, despite not being
question-conditioned. Reported as an observation; no claim about question-conditioning rests on it, that
result being established by the identical-pixel controls of \autoref{sec:signal}.

A by-product from the same weights points the same way and bounds it: MMDuet's two heads are mutually
near-orthogonal ($\cos = 0.0286$), so \emph{informative} and \emph{relevant to the current question} are
close to independent directions in one space. That supports readiness not being generic salience, and
cautions that readiness need not be a single thing.

\subsection{Coverage and temporal resolution}

Coverage is complete on both substrates. For MMDuet, alignment staleness is at most one of its own frame
intervals, and $0.00$\,s on the ActivityNet cell where the grids coincide. Dispider decides once per
$16$\,s by construction, giving a staleness of at most $16.00$\,s with a median of $6.05$\,s, and leaving
$83.4\%$ of our rungs more than one of its own intervals stale.

\subsection{Deployment cost}

On our primary geometry (Qwen3-VL-8B, $32$ frames, sequence length $2{,}462$), one monitored forward
pass costs $251.24$\,ms (p95 $251.85$, $3.98$ checkpoints per second, peak $17.80$\,GB), while the entire
readiness readout, one dot product on a residual vector the model has already computed, costs
$1.057$\,$\mu$s, or $0.00042\%$ of it. Running a separate activation model alongside the base model is a
second full forward, so $+100\%$, a ratio of roughly $237{,}000\times$.

Two caveats travel with these numbers. They are wall-clock \emph{inference} seconds, a different object
from the \emph{video} seconds in which every other latency figure in this paper is expressed. And the
separate-model figure is a proxy: the weights of the system it stands for are withheld, so our own
backbone stands in for theirs. No timing transfers across stacks.

\subsection{What the comparison does and does not support}
\label{app:rivals-bounds}

Four bounds travel with every number above.

\textbf{The comparison is asymmetric, deliberately.} Our probe is fitted on these substrates and scored
out of fold; each rival's direction is zero-shot and never saw this benchmark, this label, or this task.
The supported claim is that a foreign trained trigger does not transfer usefully to this task, not that
our monitor is a better system. On its own base, the trigger reads no better than a layer-$0$ floor. We did not refit either rival's head on our labels; the probe fitted on the
identical rows and activations from MMDuet's own base, above, is the closest stand-in we have for that
like-for-like upper bound.

\textbf{Only one rival is validated.} MMDuet can be checked against a published head-level number and
passes convincingly. Dispider's paper reports whole-system answer benchmarks with no head-level number to
reproduce, so its reading is consistent both with a weak evidence detector and with a detector of a
different notion entirely, and the comparison cannot separate those. It therefore carries no claim.

\textbf{Both scored rivals share a Qwen2 language backbone.} The only non-Qwen rival was obtained but not
run. The head-to-head therefore reproduces, on the rival axis, precisely the single-backbone-family
weakness that the seven-model evaluation removes on ours (Appendix~\ref{app:generality}). This is the
comparison's sharpest remaining limitation.

\textbf{A rival reading that does not clear the floor remains ambiguous} between ``the frozen state is
weaker on this substrate''
and ``the notion being triggered on differs''. The evidence favours the second: scored against a
clock-monotone label MMDuet's \texttt{relevance} inverts to $0.2733$--$0.3074$, and its trained direction
sits inside a random-direction null against ours. But the design cannot fully separate the two
readings, and we do not claim it does.

\section{Activation intervention detail}
\label{app:causal}

This appendix expands \autoref{sec:causal}: how the intervention site is chosen, what is added and what
it is compared against, and what the result does and does not establish.

\subsection{Choosing the site}

The site is determined by a measurement, not by a search over outcomes. The probe is fitted on a training
split, its site selected on a validation split, and then frozen. Alongside it we measure \emph{layer
eligibility}: how decodable the model's own answer letter is at each layer. The answer letter reads
$0.996$--$0.997$ at late depth across four footage families and five models, and every layer at which
readiness decodes above roughly $0.72$ also carries answer-letter AUROC between $0.90$ and $1.00$. Those
layers are ineligible, because an intervention there would move the output for a trivial reason: it
would be editing the answer, not the state that precedes it. A mid-depth site satisfies both constraints
and is re-derived independently in every evaluation that runs this procedure.

A second requirement is checked before any intervention runs: the frozen probe must read the
deferral-prompt variant of the same rows at AUROC $\ge 0.80$. A readout that does not survive the change of
prompt is not a stable enough object to intervene on. This requirement is binding rather than nominal:
where a candidate substrate or model falls below it, the threshold is not lowered and the site is not
re-selected, and that configuration simply contributes nothing to the results in \autoref{sec:causal}.

\subsection{What is added, and what it is compared against}

We estimate a rank-$32$ readiness subspace by taking the singular value decomposition of the
mean-centred matrix of matched \present\ minus \absent\ activation differences, computed on the
training split only, and keeping its leading $32$ right singular vectors. The edit moves a row's
coordinates on that basis toward a target coordinate; on treated rows the target is the row's own
partner in the matched pair, so the edit size is that item's observed evidence gap rather than a free
parameter. We add it at the frozen site on \absent\ rows measuring the change in the model's \emph{commitment}: the signed
change in the substantive-answer logits against the deferral logit. Three quantities are recorded. The
signed logit change carries direction directly. A divergence measure of the whole output change is
sign-blind by construction, so it is never used to support a directional statement on its own; a signed
companion in the same units is used for that.

An effect counts only as an \emph{excess over a matched control}, and how the control is matched is
load-bearing. A control matched only on the \emph{number} of edited units leaves the effect inflated,
because the tested units may simply write more into the residual stream than an arbitrary unit does.
Every comparison here uses a \textbf{per-item magnitude-matched} control: the target coordinate is
taken from a \emph{different} pair's partner, so the control's edit magnitude on each item is a real
evidence gap drawn from the same population, matched item by item rather than on average. Adopting per-item matching in place
of count matching absorbed roughly a third of one previously measured effect, which is why it is stated
rather than assumed.

Around that comparison sit four further controls. The first is a family of $35$ alternative
directions: $19$ random unit vectors, $8$ drawn orthogonal to the readiness subspace's leading
component, and $8$ recomputed from labels shuffled within video. All are evaluated under the identical
edit and dose, so the family is a null over \emph{which direction is moved} rather than over item
order. The reported $p$ is a one-sided max-statistic value over that family, controlling the
family-wise error rate directly; when the treatment exceeds all $35$, it takes its floor of
$1/36 = 0.0278$. Then: a dose--response curve over four doses, summarised by a Spearman
correlation; an equivalence test showing the distribution over answer content did not move; and a
replication restricted to the evidence-present-but-wrong subset.

\subsection{Results}

On the matched-pair test split, adding the direction produces the effects in
\autoref{app-tab:causal}. The response is monotone across doses in the primary model, it exceeds all
$35$ alternative directions, and the answer-content distribution stays inside its equivalence margin.

\begin{table}[H]
\centering
\scriptsize
\setlength{\tabcolsep}{4pt}
\begin{tabular}{lcc}
\toprule
quantity & Qwen3-VL-8B & InternVL3.5-8B \\
\midrule
commitment excess over a magnitude-matched control & $+0.3924$ $[+.331,+.454]$ & $+0.3495$ $[+.306,+.395]$ \\
deferral-to-answer flips, treatment vs.\ control    & $101$ vs.\ $3$ & $91$ vs.\ $7$ \\
exceeds all $35$ alternative directions             & yes, $p_{\max T}=0.0278$ & yes, same $p$ \\
dose--response Spearman over four doses             & $1.0$ & $0.8$ \\
answer-content equivalence test                     & passes & passes \\
replicates on evidence-present-but-wrong            & yes & yes \\
\bottomrule
\end{tabular}
\caption{\textbf{Adding the readiness direction changes answer commitment.} Nothing was re-selected for
the second model: it ran through the locked cell chosen on the first. The second model's dose curve is
not fully monotone, which is the one qualitative difference between them.}
\label{app-tab:causal}
\end{table}

\subsection{Head-level modulation}

Naming the units in advance turns a post-hoc search into a test. Two attention heads named before
evaluation rank first and second of $32$ at the intervention layer for modulating the addition effect,
giving the minimum attainable rank-sum of $3$ and an exact $p$-value of $1/496 = 0.0020$ from
the rank-sum distribution. A companion index-free statistic measures the share of total effect mass
carried by the named units against a within-item permutation null, so the question can travel to models
whose head indices do not correspond. A share statistic is blind to the magnitude of what is being
shared and needs a magnitude floor beneath it; ours is reported with one.

One implementation detail is a blocking check rather than a footnote. A residual-stream hook at a given
layer leaves that layer's hidden state byte-identical, with maximum absolute difference exactly $0.0$ over
$120$ measurements. The first observable change appears exactly one layer later. A head-level
forward pre-hook fires \emph{inside} the preceding decoder layer and lands at the layer itself. The two
therefore land at different effective depths, and an analysis run under the wrong convention is
mislabelled by one layer. This was verified on three models at three layers.

\subsection{Scope}

These interventions establish sufficiency at the tested site: adding the decoded direction changes answer
commitment, specifically, monotonically in dose, and beyond what a magnitude-matched edit of the same size
achieves on the same example. They do not establish necessity. Subtracting the subspace changes the model
broadly and no more specifically than a magnitude-matched wrong-target edit, and the strictest component
knockout amplifies the addition effect rather than removing it, while leaving the model's natural
evidence-present against evidence-absent commitment gap intact. The supported statement is that the
decoded direction has a behavioural role when added, not that this subspace or this head pair is the
necessary locus of readiness. The evidence is also narrow in coverage: one intervention site, one matched-pair
construction, and two video-language models.

\section{Reproducibility and compute}
\label{app:repro}

\subsection{What is and is not fitted}

The video-language model is frozen throughout. No weights are trained, fine-tuned, or adapted at any
point in this work. The only fitted object is the L2-regularized logistic-regression probe described in
Appendix~\ref{app:probe-cv}, whose features are standardized using training-fold statistics only. The
activation interventions of \autoref{sec:causal} are hooks applied to the residual stream during a
forward pass: they change activations for the duration of that pass and never change a parameter.

Evaluation is group-held-out throughout. The video identifier is the grouping key for every
natural-footage evaluation; nested cross-validation selects the readout layer and the regularization
strength on an inner held-out-group loop, so test groups influence neither the standardization, nor the
hyperparameters, nor the layer choice. Confidence intervals are cluster bootstraps over whole groups with
$1{,}000$ resamples.

\subsection{Determinism, and two exceptions worth stating}

Frozen weights and fixed, pre-decoded frames make each forward pass deterministic. Frame sets are
content-addressed by SHA-256, and the crossed designs verify byte-identity pair by pair at build time,
which is what licenses the identical-pixel claims in \autoref{sec:signal}. Sampling is deterministic
wherever the geometry permits it; the one geometry that requires randomness, the sparse-burst layout,
keys its generator per item rather than drawing from a global stream, so a shard, an item cap, or a re-run
reproduces the same bursts for the same item and no item's geometry depends on how many items preceded it.

\paragraph{The constructed item sets.}
The four constructed benchmarks of Appendix~\ref{app:constructed} are built by seeded, deterministic
loaders: one global seed, and a per-item generator keyed on the item's own identifier, so an item's
distractors and option order do not depend on how many items preceded it or on how the build was
sharded. Two of the four commit a frozen artefact alongside the code, by two different mechanisms.
The Oops!\ option set is a build input: the loader joins it by item identifier and hash rather than
regenerating it, which is what makes that build reproducible without re-running its generator model.
The matched \present/\absent\ pairs are rebuilt from the official annotations by the seeded loader,
and the committed manifest is the frozen record of that build, one row per item carrying the content
hash the rebuild must reproduce. For the other
two, the item population is frozen and committed while the per-arm frame manifests are rebuilt from it
by the same seeded loader; that build is annotation-driven and CPU-only, so it reproduces without a
GPU. The one build that cannot be reproduced from annotations alone is the Oops!\ option set, since
regenerating it would require a full pass of the generator model; the committed artefact is therefore
the definition of that benchmark rather than a cache of it.

Two exceptions are real and are stated rather than smoothed over.

First, a nominally deterministic solver is \textbf{not} bit-reproducible across hosts. Re-running one
fitted control reproduced $3$ of $26$ committed values exactly, with the remainder moving by up to
$0.0152$. Fitted control values are therefore not quoted to four decimal places anywhere in this paper.

Second, per-item activation values are \textbf{not} portable across GPU architectures: the maximum
per-item difference between architectures is $0.113$, against arm-level agreement of $0.0059$. Aggregates
are portable and per-item values are not, so any analysis whose claim is the \emph{absence} of an effect
was run on a single host, and per-item numbers are never compared across hardware.

\subsection{Compute}

Runtime depends on the software and hardware stack, so the timing measurements below specify both
their environment and measurement procedure. To reduce random-seek decode cost, the frame-build
pipeline uses pre-transcoding to a $640$\,px long side with a short GOP before resizing sampled
frames to $448$\,px. This preserves frame rate and frame count.

Extraction ran on a shared cluster of RTX PRO 6000, RTX 6000 Ada, and L40S GPUs. Two individual costs are
measured: the head-level intervention analysis at $\approx 40$ GPU-hours, and the uncertainty-baseline
sweep's ActivityNet cell at $41.07$ GPU-hours.

No programme-wide GPU-hour total was recorded, and we do not estimate one. The analysis and probing
pipeline runs under Python $3.10$ with PyTorch $2.8.0$/CUDA $12.8$ \citep{paszke2019pytorch},
\texttt{transformers} $5.1.0$ \citep{wolf2020transformers}, scikit-learn $1.7.2$
\citep{pedregosa2011scikit}, NumPy $2.2.6$ \citep{harris2020array} and PyAV $16.1$.
The GPU stack is not a single environment: the
timing measurement below ran under PyTorch $2.10.0$, a different build from the extraction stack, so we
cite the stack beside each timing number rather than presenting one global environment table.

The timing figures quoted in \autoref{sec:utility} were measured on one host under PyTorch
$2.10.0$/CUDA $12.8$ at batch size $1$, with $3$ discarded warm-up iterations and $40$ timed
repetitions, synchronising the device at both ends of every timed call. The forward pass has mean
$251.24$\,ms with p50 $251.25$, p95 $251.85$, and a min--max range of $250.06$--$255.40$\,ms; no
standard deviation was recorded. The monitor figure is the mean of $5{,}000$ repetitions of a
single-precision dot product on one residual vector, timed on CPU. The separate-activation-model
figure is a declared proxy rather than a measurement: it is defined as one additional identical
forward pass, because the weights of the system it stands for are not publicly released.

\section{Extended limitations}
\label{app:limitations}

This appendix expands the limitations stated in \autoref{sec:conclusion}. It records the limits of the
results we report, and the reasons several controls take the form they do.

\subsection{Coverage}

The representation results rest on the seven models of the shared frozen-manifest evaluation, spanning
four language-backbone families and $2$B to $32$B parameters, with a fifth, Qwen2, entering on the
NExT-QA pools. The remaining results are narrower, and the
difference matters. The graded accumulation results come from a single model. The activation
interventions cover two video-language models at one primary site. The cross-attribute transfer result comes
from one model on controlled stimuli. All models evaluated here are open-weight; we test nothing closed.
Breadth on the representation claim should not be read as breadth on the graded, causal, or
attribute-transfer claims.

\subsection{Evidence annotations are proxies}

A grounding interval marks \emph{where the answer is depicted}, which is a proxy for \emph{the earliest
moment at which the question becomes answerable}. The labels therefore measure timestamp coverage
rather than perceived sufficiency. The gap is measured directly in
Appendix~\ref{app:raters}: on the $840$ of $900$ screens that carry a binary label, independent raters
shown only the delivered frames recover the benchmark's own evidence label at $\mathrm{AUROC}$
$0.681$--$0.778$, which is the level the probe reads it at, while disagreeing with that label on
$24$--$30\%$ of screens. A separate label-cleaning diagnostic on the grounded pool points the same way:
independent models recover a human readiness label from their own behaviour at only
$\mathrm{AUROC} \approx 0.56$, while hidden states decode the same label at $\approx 0.73$. The internal representation tracks the annotated
window considerably better than any behaviour does, and crisper action boundaries on egocentric footage
decode higher still. Relatedly, on one benchmark family the annotated span uniquely decides the answer for
only about $5\%$ of items, which bounds what an evidence-timing policy can recover there
(Appendix~\ref{app:utility}).

\subsection{Shared footage between two natural-evidence evaluations}

One of the two natural-evidence multiple-choice evaluations is a human re-annotation of a subset of the
other's source corpus: all $348$ of its videos are NExT-QA files and $221$ of them also appear in the
other pool. The two are therefore
\emph{not} independent domains. That pairing replicates the \emph{labelling}, not the footage:
different annotators independently marked evidence timing under different instructions.
Genuinely independent video enters with the movie-trailer substrate, the multi-source graded pool, and the
independent ActivityNet-Captions evaluation, and this is why the paper says eight benchmark
\emph{families} rather than eight independent datasets.

\subsection{Scope of the controlled stimuli}

Two limits apply to the composited-cue experiments. Absence is \emph{model-screened} rather than
ground-truth annotated: $41$ of $540$ (object, clip) pairs were excluded by asking the frozen model
whether the object appears, and a genuinely annotated absence set would be stronger. And because the two
attributes share one $17$-object library, held-out-object cross-validation holds out the \emph{focal}
object but not every relational object appearing in the scene.

\subsection{Calibration}

Expected calibration error in domain is $0.104$, $0.144$ and $0.139$ on three evaluations. For a signal
intended as a deployed answer-timing trigger, calibration in that range is not good enough on its own.
Under distribution shift it is far worse: the carried-over probe on the external benchmark reaches ECE
$0.8079$, with accuracy $0.1377$ at the carried-over threshold against $0.8681$ at a target-chosen one.

Appendix~\ref{app:transfer} measures the repair rather than leaving this open, but three limits remain.
It needs labels, so a genuinely zero-label deployment still carries the miscalibration.
Label-free correction cannot substitute in at least two cases for structural reasons: a
label-shift estimator's input is the predicted positive rate at the very threshold that is broken, and
temperature scaling cannot move accuracy at a fixed threshold at all. It helps only under base-rate
shift, and where the deployment base rate already matches the pool the probe was standardized on,
spending the same labels makes accuracy slightly worse. And knowing which of those two regimes one is in
requires knowing the target's base rate, which a deployer may not.

\subsection{Methodological caveats}

\paragraph{Late layers couple to the output.}
Where the correct answer maps deterministically onto a fixed option, the late-layer readiness direction
nearly coincides with the direction that emits that letter. Every causal claim in this paper is therefore
made at a mid-depth site, chosen before the answer letter becomes decodable, and randomising option order
removes the confound and \emph{strengthens} the mid-depth effect rather than weakening it.

\paragraph{Cosine between high-dimensional directions is weak evidence.}
In $4096$ dimensions, unrelated vectors have $\mathrm{SD}(\cos) \approx 1/\sqrt{4096} \approx 0.016$, so a
cosine near zero is exactly what chance predicts and is not evidence of independence. The dissociation
results in \autoref{sec:notconf} are therefore prediction-level throughout, using within-bin decoding,
residualization, and incremental validity. Where we do report a cosine, we compare it with a measured
random-direction null at the relevant dimension. Conversely, the cross-attribute cosine of $0.266$ is
informative precisely \emph{because} transfer succeeds despite the directions not being aligned.

\paragraph{The crossed $2\times2$ is impossible on a growing prefix.}
A crossed design needs both time orderings: one question's evidence entirely after the other's, and the
reverse. For a growing prefix, this contradicts the fact that an evidence span starts before it ends. No
crossed pool built on a growing prefix can yield a DiD, so the prefix-based crossed
evaluations report the paired contrast only. A fixed-span window can sit before \emph{or} after a
given question's evidence, so both orders exist for one question pair, which is why the
DiD reported in \autoref{sec:signal} is a window result.

\paragraph{The same-pixels control is unavailable on both graded benchmarks, for opposite reasons.}
On the movie-trailer substrate there are many questions per video, but every question's evidence spans the
whole video, so no pair of questions has disjoint evidence hulls and there are zero separable pairs. On the
multi-source graded pool the opposite holds: there is essentially one question per video, giving only $22$
separable pairs over $20$ videos, too few to support video-disjoint cross-validation with a cluster
bootstrap. The fallback there is to ask a \emph{different video's} question on byte-identical frames,
which is if anything a sharper test, because the swapped question's evidence is absent by construction.

\subsection{Offline replay}
\label{app:limitations-replay}

Every accuracy result in this paper is an offline ladder replay: rows are pre-decoded, checkpoints are
pre-enumerated, and the policy chooses among them after the fact. One substrate was checked against a
live-decoding run of the same policy. Selection accuracy differed by $0.031$ (two items in $64$, over
$1{,}685$ rows from $40$ videos). This is not distinguishable from zero at that sample size, while the
\emph{rung} the policy selected agreed on only $44\%$ of items.

That instability is a property of the decision rather than of the harness. Re-running the identical
forward pass on the identical frames already flips about a quarter of the rungs while changing accuracy by
exactly zero, and at the activation level a re-run reproduces the cached vector at cosine $0.99997$ with
rank correlation $0.9825$. A perturbation of a few parts in $10^5$ is enough to move an argmax over
roughly $27$ near-tied rungs, and leaves both the ranking and the emitted answer intact.

The audit provides an initial live-decoding check, but does not establish equivalence with replay.
No claim about \emph{which} checkpoint a policy selects should be read
out of these experiments. We also do not measure memory growth over long streams.

Two properties bound what the cost figures mean. Because every substrate here poses a fixed answer
set whose options sit in the prompt, the final prompt token's logits are the answer distribution and
that same token's residual is the probe's input: one forward pass yields both, so running the policy
online adds no decoding and no retained key-value state. An open-ended answer would instead require
generating at each checkpoint or deferring generation to the deadline, and the overhead figures
would not carry over. Separately, our windows are fixed-span, so context never exceeds the model's
budget; whether the readiness direction survives dynamic key-value eviction on longer streams is
untested.

\section{Extended related work}
\label{app:related}

This appendix expands \autoref{sec:related}, including the representational-similarity methods
used in Appendix~\ref{app:generality} and the calibration questions raised by
Appendix~\ref{app:transfer}.

\paragraph{Streaming video and learned response triggers.}
Streaming benchmarks measure the gap between models that consume a complete clip and systems that must
act as frames arrive \citep{lin2024streamingbench,li2025ovobench}, and online VideoLLMs close it by
training for it: interleaved video--text training, memory, proactive-response modules, or a learned
readiness mechanism
\citep{chen2024videollm,wang2024mmduet,qian2025dispider,wang2025streambridge,xie2026streamov,azad2026streamready}.
This is the closest neighbourhood. Those systems \emph{train} a head to answer ``respond now?''; we ask
whether a frozen, unmodified model already carries the state such a head would need. The comparison in
\autoref{sec:rivals} is deliberately asymmetric: our probe is fitted on the substrate and scored out
of fold, whereas a released trigger is applied zero-shot. It therefore establishes that a foreign trigger does not
transfer to this task, not that our monitor is the better system on the task that trigger was built for.
Because each released trigger is itself a linear readout of a residual stream, the two can be compared as
the same kind of object (Appendix~\ref{app:rivals}).

\paragraph{Temporal grounding and evidence annotation.}
Moment retrieval localises a language-described interval in untrimmed video
\citep{hendricks2017localizing,gao2017tall,krishna2017dense}, and grounded or evidence-backed VideoQA
attaches temporal or spatio-temporal support to an answer
\citep{xiao2023trust,wang2026evidence,benami2025herbench}. We consume these annotations rather than
compete with them: a span intersected with the frames delivered so far defines the readiness label, which
is fixed before the model runs. Retrieval asks \emph{where} the moment is; we ask whether an internal
state marks that enough of it has arrived. A frame-level retrieval baseline is in the control battery
(Appendix~\ref{app:probes}); a trained span regressor is not, and that is a stated gap. The annotations
also bound us: an annotated span need not be the earliest sufficient evidence.

\paragraph{Uncertainty, abstention, and answerability.}
Calibration and self-evaluation estimate whether an output is likely correct
\citep{guo2017calibration,kadavath2022language}; consistency and semantic entropy measure stability
across generations or meanings \citep{wang2023selfconsistency,farquhar2024semantic}. SelectiveNet learns
prediction and rejection jointly on a fixed input \citep{geifman2019selectivenet}; our streaming policy
can instead wait for additional visual evidence. Recent abstention work separates whether an answer is
right from whether the question is answerable \citep{wagner2026two,gu2026bridging,xu2026look}.
Our readiness label is an \emph{input}-side quantity, defined without consulting the model's answer.
The distinction is tested rather
than asserted: a battery of such estimators, including a semantic-entropy implementation with
entailment-based clustering, is scored against the same labels (\autoref{sec:notconf}), and several read
the model's own correctness well while reading readiness worse.

\paragraph{Probing and activation intervention.}
Linear probes expose information carried in a representation even when generation does not express it
\citep{belinkov2021probing,li2023emergent,marks2024geometry,orgad2025llms}, and work on visual
information flow asks where in a VideoLLM such information becomes available
\citep{kim2026mapflow,salazar2026pathways}. Control-task work motivates interpreting probe performance
alongside controls for what the probe itself can learn \citep{hewitt2019designing}.
The contribution is not that a probe decodes something,
but the design around it: a question-swapped, byte-identical-pixel construction with a
DiD and a label that never touches model behaviour. Activation steering and causal
tracing then ask whether a decoded direction matters behaviourally
\citep{li2023inference,zou2023representation,mueller2024quest}, where conclusions depend on the
intervention and its control \citep{zhang2023towards,tan2024analysing}; we use per-example magnitude
matching, treat sufficiency and necessity as separate hypotheses, and report the necessity side as it
came out (\autoref{sec:causal}). Critiques showing that video-LLM accuracy can be attained without
genuine visual grounding \citep{lee2026accuracy,li2023evaluating} are one reason the label is defined on
the delivered frames rather than on the model's success.

\paragraph{Representational similarity between models.}
Methods for comparing or relating representations include orthogonal Procrustes alignment
\citep{schonemann1966procrustes}, SVCCA and CKA
\citep{raghu2017svcca,kornblith2019similarity}, and relative representations
\citep{moschella2023relative}. The Platonic Representation Hypothesis proposes that increasingly
capable models converge toward shared representational structure \citep{huh2024platonic}.
Appendix~\ref{app:generality} asks a narrower functional question: we fit an orthogonal map between two frozen VideoLLMs' residual
streams \emph{without labels}, on shared inputs, and then read one model's \emph{task probe} in the
other's basis. This functional test asks whether a specific readout survives the change of basis,
rather than treating global similarity as evidence of readiness-specific structure. Orthogonal alignment in high dimension fits almost anything, so the
controls carry the result: a random-rotation null at the same rank, a correspondence-broken null, an
un-aligned baseline, and a sweep over rank. The un-aligned baseline is at chance, which is what makes the
aligned reading meaningful and also what bounds it: the shared structure is recoverable only up to a
rotation, and recovering that rotation needs paired activations on shared inputs, so this is not a probe
that ships between models.

\paragraph{Label shift, post-hoc calibration, and source-free adaptation.}
When a probe moves to a new distribution the standard repairs are label-shift estimators such as
black-box shift estimation and its regularised variants \citep{lipton2018bbse,azizzadenesheli2019rlls},
and post-hoc calibration such as temperature scaling \citep{guo2017calibration}.
Entropy minimisation also supports source-free test-time adaptation, as in Tent \citep{wang2021tent},
which updates model parameters rather than only calibrating a frozen score. Appendix~\ref{app:transfer}
instead measures threshold repair with a small labelled target budget while keeping the VideoLLM and
probe direction frozen. These approaches address different failure modes. A label-shift
estimator's input is the source classifier's confusion \emph{at its own operating threshold}, so its
estimate of the target prior is a function of the predicted positive rate at exactly the threshold that
has broken; when that threshold calls almost every row ready, the statistic it consumes is already
saturated. Temperature scaling is worse placed still: it is a strictly monotone rescoring, so it cannot
change the ranking. At the binary decision threshold of $0.5$, positive-temperature scaling changes
the reported probabilities but not the predicted class. The same monotonicity argument governs a quantile-thresholded stopping policy,
where any monotone rescoring moves zero stops. What survives is the distinction between a ranking result
and an operating-point result, and only the second needs target labels to repair.

\paragraph{Answer timing, anytime prediction, and cascades.}
Anytime prediction \citep{huang2018multiscale}, early exiting \citep{teerapittayanon2016branchynet}
and cascaded inference \citep{chen2024frugalgpt} allocate computation before emitting a prediction.
Our policy instead chooses when to answer as additional video evidence arrives, or which cached answer
to return at a fixed deadline. Our stopping and
selection policies are deliberately standard, so that what varies between arms is the monitor consulted
and not the policy consuming it; the contribution on this axis is the monitor and the measured relation
between what it gains and how much an evidence-timing policy could recover at all (\autoref{sec:utility}).

\end{document}